\documentclass[mnsc,sglanonrev]{informs4}

\makeatletter
\def\theARTICLETOPLEFT{}
\def\theARTICLETOPRIGHT{}
\def\theJOURNAL{}%
\def\theJOURNALshort{}%
\RRHFirstLine{}
\LRHFirstLine{}
\RRHSecondLine{\bf\theRUNAUTHOR:\enskip {\it\theRUNTITLE}}
\LRHSecondLine{\bf\theRUNAUTHOR:\enskip {\it\theRUNTITLE}}
\makeatother

\RequirePackage{tgtermes}
\RequirePackage{newtxtext}
\RequirePackage{newtxmath}
\RequirePackage{bm}
\RequirePackage{endnotes}
\newcommand{\qedsymbol}{\hfill \ensuremath{\square}}

\OneAndAHalfSpacedXII 

\usepackage{algorithm}
\usepackage[noend]{algpseudocode}
\usepackage{multirow}
\usepackage{booktabs}
\usepackage{makecell}
\usepackage{lscape}

\usepackage[most]{tcolorbox}

\tcbset{
  vasstyle/.style={
    colback=red!5!white,
    colframe=red!75!black,
    boxrule=0.8pt,
    arc=4pt,
    left=6pt,
    right=6pt,
    top=4pt,
    bottom=4pt,
    enhanced,
  },
  agnistyle/.style={
    colback=blue!5!white,
    colframe=blue!75!black,
    boxrule=0.8pt,
    arc=4pt,
    left=6pt,
    right=6pt,
    top=4pt,
    bottom=4pt,
    enhanced,
  },
  gonstyle/.style={
    colback=yellow!5!white,
    colframe=yellow!75!black,
    boxrule=0.8pt,
    arc=4pt,
    left=6pt,
    right=6pt,
    top=4pt,
    bottom=4pt,
    enhanced,
  }
}

\usepackage{tikz}
\usetikzlibrary{arrows.meta, positioning, fit, calc, shapes.geometric}

\usepackage{natbib}
 \bibpunct[, ]{(}{)}{,}{a}{}{,}%
 \def\bibfont{\small}%
\EquationsNumberedThrough    

\TheoremsNumberedThrough     
\ECRepeatTheorems  %

\MANUSCRIPTNO{MNSC-0001-2024.00}

\begin{document}


\RUNAUTHOR{Digalakis and Krishnan and Martín Fernández and Orfanoudaki}

\RUNTITLE{ML Compass}

\TITLE{ML Compass: Navigating Capability, Cost, and Compliance Trade-offs in AI Model Deployment}


\ARTICLEAUTHORS{%
\AUTHOR{Vassilis Digalakis Jr}
\AFF{Questrom School of Business,
Boston University}

\AUTHOR{Ramayya Krishnan}
\AFF{H. John Heinz III School of Infomraton Systems and Public Policy,
Carnegie Mellon University}

\AUTHOR{Gonzalo Mart\'{i}n Fern\'{a}ndez}
\AFF{Centre de Formació
Interdisciplinària Superior and Universitat Politècnica de Catalunya} 

\AUTHOR{Agni Orfanoudaki}
\AFF{Sa\"{i}d Business School,
Oxford University} 

} 

\ABSTRACT{
We study how organizations should select among competing AI models when user utility, deployment costs, and compliance requirements jointly matter.
Widely used capability leaderboards do not translate directly into deployment decisions, creating a capability--deployment gap; to bridge it, we adopt a systems-level view of model selection and deployment, in which model choice is tied to application outcomes, operating constraints, and a feasible capability--cost frontier.
We develop ML Compass, a unifying framework that treats model selection as a constrained optimization problem over this frontier. 
On the theory side, we characterize optimal model configurations under a parametric frontier and show that optimal internal measures exhibit a three-regime structure: some dimensions are pinned at compliance minima, some saturate at their maximum feasible levels, and the remainder take interior values governed by the curvature of the frontier. 
We derive comparative statics that quantify how budget changes, regulatory tightening, and technological progress propagate across capability dimensions and costs. 
On the implementation side, we propose a practical pipeline that (i) extracts low-dimensional internal measures from heterogeneous model descriptors, (ii) estimates an empirical frontier from capability and cost data, (iii) learns a user- or task-specific utility function from interaction-level outcome data, and (iv) uses these components to target capability--cost profiles and recommend models. 
We validate ML Compass with two case studies: a general-purpose conversational setting using the PRISM Alignment dataset and a healthcare setting using a custom dataset we build using HealthBench. 
In both environments, our framework produces recommendations---and corresponding \emph{deployment-aware leaderboards} based on predicted deployment value under explicit constraints---that can differ materially from capability-only rankings, and clarifies how trade-offs between capability, cost, and safety shape optimal model choice.
}%


\maketitle


\section{Introduction}\label{sec:Intro}

Organizations deploying artificial intelligence (AI) systems face unprecedented complexity in model selection decisions. A financial services firm choosing a conversational chatbot must balance response quality against inference costs and latency constraints. A healthcare provider selecting an AI assistant for medical advice must weigh diagnostic accuracy against interpretability requirements for clinical staff and regulatory compliance for patient safety. In practice, AI deployers must simultaneously satisfy users demanding performance, business leaders controlling costs, and regulators (or internal governance teams) enforcing compliance. The challenge is selecting models that satisfy multiple stakeholder requirements within a rapidly changing technological landscape.

The proliferation of large language models (LLMs) has intensified this challenge. 
Unlike traditional supervised-learning systems optimized for a specific task, LLMs are general-purpose and must be evaluated across many dimensions. 
Organizations now navigate an expanding landscape of options, with over 160 foundation models released within 2023--2024 alone \citep{hai2025index}. 
Even when attention is restricted to major cloud providers, deployers still face a nontrivial and fast-moving choice set: providers curate evolving catalogs that mix first-party and third-party models, and availability and pricing change frequently.
Each model presents distinct trade-offs across accuracy, inference latency, computational cost, safety alignment, and domain expertise \citep{liang2022holistic}. 
Current evaluation practices rely heavily on leaderboards that rank models on individual benchmarks, yet top positions can change rapidly as new releases arrive \citep{srivastava2023beyond}. 
A model leading in mathematical reasoning may lag in code generation or multilingual support, and single-metric rules that are often adequate in traditional ML do not resolve such multidimensional trade-offs \citep{folk2024google}. 
More fundamentally, capability leaderboards are typically disconnected from the full set of deployment considerations: benchmark scores often have unclear operational meaning and can be weak proxies for application outcomes and organizational constraints \citep{fodor2025line}. 
Deployers therefore need a decision framework that captures persistent \emph{trade-off structure} rather than transient rankings.


Our work addresses the central question AI deployers face: \textit{given a landscape of AI models with heterogeneous capabilities and costs, how should deployers select models that maximize user value while satisfying cost and compliance constraints?} 
We adopt the perspective of the deployer who must translate technical capabilities into deployment value under real operating constraints. 
Users derive value from task performance and interaction quality; business leaders evaluate models through lifecycle costs spanning inference, licensing, integration, and ongoing operational overhead; regulators (or internal governance teams) impose minimum standards for, e.g., safety and transparency.

Despite extensive research on AI evaluation and responsible AI deployment, existing approaches provide limited actionable guidance for these deployment decisions. 
Technical evaluation frameworks emphasize benchmark reporting for accuracy, robustness, and fairness without formalizing the joint trade-offs with resources and constraints \citep{liang2022holistic, srivastava2023beyond}. 
Business research studies adoption barriers and success factors but rarely models selection as an optimization problem with explicit capability--cost structure \citep{allen2022algorithm, dietvorst2018overcoming}. 
Regulatory guidance emphasizes compliance and risk management but typically abstracts from spillovers, where tightening one requirement can degrade performance on other dimensions \citep{euaiact2024, nist2023airmf}. 
Our work bridges these perspectives by introducing an integrated framework that explicitly models the interaction between model capabilities, deployment resources, and stakeholder constraints.

\subsection{Contributions \& Outline} \label{ssec:contributions}

Combining tools from optimization, structural analysis, and machine learning (ML), we develop ML Compass (MLC), a general framework for deployment-aware model selection. Our main contributions are:
\begin{itemize}
    \item \textbf{A unified deployment decision model.}
    We formalize AI model selection as a constrained optimization problem that jointly incorporates (i) user/task utility, (ii) deployment resources and operational constraints, and (iii) compliance requirements (imposed externally by regulators or internally by organizational governance).
    A central ingredient is a \emph{technological frontier} that bounds feasible capability profiles at a given resource level, making capability--cost trade-offs explicit and turning benchmark measurements into inputs to a deployment decision.

    \item \textbf{Structural characterization and comparative statics.}
    Under a parametric frontier and linear utility, we characterize the optimal capability target in a \emph{three-regime} structure: some dimensions bind at compliance minima, some saturate at feasible maxima, and the remainder take interior values pinned down by frontier curvature and cost sensitivity.
    We derive comparative statics showing how (i) marginal budget expansions scale all interior dimensions proportionally, (ii) tightening a binding compliance threshold induces quantifiable spillovers onto other interior dimensions with magnitudes governed by substitutability, and (iii) technological progress (frontier shifts) systematically relaxes the capability--cost trade-off, expanding the feasible set and improving the optimal objective in ways that depend on the regime configuration.

    \item \textbf{A scalable implementation pipeline and a deployment-aware leaderboard.}
    We operationalize the framework as a pipeline that (i) extracts low-dimensional internal measures from high-dimensional capability descriptors, (ii) estimates an empirical capability--cost frontier, (iii) learns context-dependent utility from interaction-level outcomes, and (iv) optimizes under explicit cost and compliance inputs to produce targets and recommended models.
    The same scenario objective induces a \emph{deployment-aware leaderboard}: fixing a deployment scenario (cost sensitivity and constraints), each model receives a score, and ranking feasible models by this score yields a benchmark that is tied to deployment requirements rather than to a single capability metric.

    \item \textbf{Empirical validation in two deployment settings.}
    We validate MLC in (i) a general conversational deployment using PRISM \citep{kirk2024prism}, where outcomes reflect heterogeneous human preferences, and (ii) a healthcare deployment using a custom evaluation dataset we built with HealthBench rubrics \citep{arora2025healthbench}, where outcomes reflect physician-designed clinical quality and safety criteria.
    Across both settings, the resulting recommendations can diverge materially from capability-only rankings once costs and constraints are made explicit, and the framework clarifies \emph{why} a given model is selected (value--cost trade-off versus binding constraints).
\end{itemize}

The remainder of the paper is organized as follows.
Section~\ref{sec:problem-formulation} formalizes the MLC optimization problem.
Section~\ref{sec:stylized-model} presents the stylized model and derives structural results and comparative statics.
Section~\ref{sec:compasspipeline} translates the framework into an implementable pipeline, including algorithms for internal-measure extraction, empirical frontier estimation, and utility estimation from interaction-level outcomes.
Section~\ref{sec:casestudies} applies the pipeline end-to-end in the PRISM and HealthBench case studies and evaluates deployment-aware recommendations under multiple deployment scenarios.
Section~\ref{sec:discussion} concludes with managerial takeaways, limitations, and future directions.
Proofs and additional robustness and implementation details appear in the Electronic Companion (EC).



\subsection{Related Literature}\label{sec:literature}
Our work relates to three streams: technology selection and adoption, AI/LLM evaluation, and multi-criteria decision making. Across these areas, a recurring challenge is that deployers must choose among alternatives that trade off quality, resources, and constraints; in the LLM setting this is amplified by high-dimensional capability profiles, volatile leaderboards, and the fact that deployers typically select from a \emph{discrete} and rapidly changing set of third-party models. We contribute a deployment-focused \emph{selection} framework that combines an empirical capability--cost frontier with context-dependent utility and explicit cost/compliance constraints, rather than studying adoption dynamics or proposing new benchmarks in isolation. We briefly highlight the key contributions from each related stream and discuss their relevance to our study.

\paragraph{Technology selection and adoption.}
The operations management literature has long examined how firms select and adopt new technologies under conditions of uncertainty. \citet{fine1990optimal} developed models for optimal investment in flexible manufacturing technology, demonstrating how capacity and capability trade-offs shape technology choices. \citet{goyal2007strategic} extended this work to consider strategic technology selection with network effects and competitive dynamics. Recent work has begun examining AI and ML adoption specifically. \citet{bastani2021mostly} study how firms navigate the exploration-exploitation trade-off when selecting ML algorithms, while \citet{bertsimas2020predictive} show that the value of predictive models depends critically on their integration with decision-making processes. This stream of literature provides foundational insights into technology adoption under uncertainty, yet lacks a unified framework that captures the binding technological constraints and multi-dimensional trade-offs inherent in AI model selection. We make these trade-offs explicit via a technological frontier and a constrained objective for model selection, rather than modeling dynamic capability accumulation or adoption under uncertainty.


A parallel literature has examined behavioral and organizational barriers to AI adoption. \citet{dietvorst2015algorithm} identify ``algorithm aversion,'' demonstrating that decision-makers often prefer human judgment over algorithmic recommendations even when algorithms demonstrably outperform humans. \citet{dietvorst2018overcoming} extend this work by showing that giving users modest control over algorithmic outputs increases adoption without significantly degrading performance. \citet{ibrahim2021eliciting} develop methods for incorporating human expertise into prediction algorithms, revealing that successful AI deployment requires careful calibration between human judgment and machine recommendations. \citet{choudhury2020machine} examine how ML affects organizational decision-making, finding that AI adoption success depends critically on aligning algorithmic capabilities with organizational processes. \citet{grand2024best} show that the value of algorithmic advice depends critically on user perception and degree of adherence, highlighting the importance of considering user characteristics in model selection.  Together, these studies underscore that successful AI deployment requires alignment with user preferences, organizational processes, and stakeholder expectations. Our approach does not aim to explain these behavioral mechanisms; instead, we treat heterogeneity as a deployment input (context $z$), estimate context-dependent utility from outcomes, and optimize model choices under explicit constraints.

\paragraph{AI evaluation.}
The emergence of LLMs has further complicated model selection decisions and deployment performance evaluation. Unlike traditional ML systems optimized for specific tasks, LLMs must be assessed across diverse capabilities. A large technical literature develops evaluation suites and leaderboards for LLMs, spanning reasoning \citep{wei2022emergent}, factual accuracy \citep{min2023factscore}, safety alignment \citep{bai2022training}, and domain expertise \citep{singhal2023large}. The degree to which these models generate factually grounded responses without hallucination has become a critical evaluation dimension \citep{ji2023survey}, alongside concerns about bias amplification \citep{weidinger2021ethical} and robustness to adversarial inputs \citep{wang2021adversarial, li2025firm}. 

To holistically consider these complementary aspects of LLM performance, numerous evaluation frameworks have been proposed that assess models across multiple dimensions \citep{srivastava2023beyond}. Technical benchmarking initiatives provide comprehensive assessment across accuracy, calibration, robustness, fairness, and efficiency dimensions for a diverse set of tasks \citep{liang2022holistic}. Complementary approaches focus on computational efficiency, measuring the speed with which trained models can reach a target quality metric, although without integrating them with task performance or stakeholder requirements \citep{reddi2020mlperf}. However, these frameworks primarily measure and report performance without providing guidance on how to select models when no option dominates across all metrics. Our contribution is to turn these measurements into a deployment-aware objective and feasibility region for model selection (and, by extension, a deployment-aware leaderboard). In parallel, governance frameworks such as the NIST AI Risk Management Framework \citep{nist2023airmf} and the EU AI Act \citep{euaiact2024} establish compliance requirements across these dimensions, but do not provide optimization methods for meeting multiple constraints simultaneously.

\paragraph{Multi-criteria decision making.}
To address these limitations, recent work has turned to formal optimization methods from operations research. Multi-objective optimization allows ML modelers to tackle the often conflicting objectives by including multiple criteria in the objective function scaled by different weights of importance. Being agnostic to the specific criteria, \citet{liu2021stochastic} proposed a stochastic multi-gradient algorithm for multi-objective optimization where different types of target criteria (e.g., robustness and safety) can be used as input. In the context of ML, \citet{zafar2017fairness} proposed a bi-criteria framework to balance the trade-offs between accuracy and fairness in classification. Similarly, \citet{bertsimas2019price}, and \citet{bertsimas2023improving} and \citet{bertsimas2024towards} investigated, respectively, the ``price'' of interpretability and stability with respect to accuracy when decision makers are keen to ensure the transparency and consistency of the derived models. More generally, \cite{cortes2020agnostic,sukenik2022generalization} (and the references therein) study generalization in multi-objective ML from a statistical perspective. However, these methods lack the structural analysis needed to understand how cost and compliance constraints shape optimal model choice. Moreover, relative to this literature, we focus on deployment-time selection among existing AI models by (i) estimating an empirical capability--cost frontier from model data, (ii) learning context-dependent deployment value from outcomes, and (iii) optimizing under explicit cost and compliance constraints, rather than proposing new training-time multi-objective algorithms or treating the frontier as an end in itself.

In all, our work synthesizes these streams to address the absence of a unified framework focused on model selection decisions that combines rigorous optimization techniques with practical AI evaluation needs while explicitly modeling technological constraints. To this end, our proposed Pareto frontier approach builds on established methods from production economics and efficiency analysis. Data Envelopment Analysis, pioneered by \citet{charnes1978measuring} and extended by \citet{banker1984some}, provides non-parametric approaches to estimating efficiency frontiers when multiple inputs produce multiple outputs. The production frontier literature \citep{farrell1957measurement, aigner1977formulation} offers theoretical foundations for understanding the boundary of achievable performance given technological constraints. We adapt these concepts to the AI model selection context, where the frontier represents achievable combinations of model capabilities (e.g., accuracy, speed, interpretability), extending traditional efficiency analysis to multi-stakeholder optimization problems. 

\section{Problem and Model Formulation} \label{sec:problem-formulation}
This section formalizes AI model selection as a constrained optimization problem that captures the multi-stakeholder nature of deployment decisions. We take the perspective of the \emph{AI deployer}, who must decide which model to select given the perspectives of three stakeholders: users, business leaders, and regulators. Each stakeholder evaluates a model through distinct criteria that link internal measures of model performance to deployment value, resource usage, or compliance outcomes.

Let $\mathcal{M}$ denote the set of candidate models. Each model $m \in \mathcal{M}$ is characterized by a vector of internal performance measures $x(m) \in \mathbb{R}^I_+$, capturing capabilities such as accuracy on standardized benchmarks, inference latency, interpretability, robustness to adversarial inputs, and safety alignment. For example, when evaluating LLMs, these measures may include MMLU scores for knowledge assessment \citep{hendrycks2020measuring}, TruthfulQA metrics for factual accuracy \citep{lin2022truthfulqa}, and HumanEval pass rates for code generation \citep{chen2021evaluating}. Each model $m$ is also associated with a lifecycle cost $c(m) \in \mathbb{R}_{+}$ capturing deployment resources such as inference compute, licensing fees, and maintenance overhead. \footnote{In LLM deployments, a practical scalar notion of cost is often the \emph{expected cost per interaction}, which depends on both per-token pricing and \emph{token efficiency} (expected tokens under the deployment workload). Hosting choices (API vs.\ self-hosting) and engineering decisions (caching, batching) change how raw cost components map into the scalar $c(m)$.}
Together, $x(m)$ and $c(m)$ define the technical and economic profile that determines whether a model is viable for deployment. \footnote{Throughout the paper, internal measures are normalized so that $x(m)\in[0,1]^I$, with larger values indicating better capability, while costs are normalized so that $c(m)\in[0,1]$, with smaller values indicating lower deployment cost.}




In some ML settings, a deployer can effectively optimize over model design or training choices (e.g., architecture, hyperparameters, training compute), so the mapping from design decisions to $(x,c)$ is part of the deployer’s control. For foundation models such as LLMs, however, deployers typically cannot tune model internals at will: they select from a \emph{discrete} set of available pre-trained models with fixed capability--cost profiles. This motivates a two-stage view of deployment decisions: first, characterize an ideal target profile $(x^*,c^*)$ given stakeholder objectives and constraints; second, map that target to an available model in $\mathcal{M}$. In our empirical implementation (Sections~\ref{sec:compasspipeline}--\ref{sec:casestudies}), we operationalize this mapping using dominance/Pareto structure to identify efficient candidates and to estimate the technological frontier from observed model profiles.

\subsection{Stakeholders and their Evaluation Functions}\label{ssec:stakeholders} 
The three stakeholder groups evaluate models according to distinct criteria that often conflict. 

\textit{Users} derive utility from model capabilities that enhance task performance. For a customer service chatbot, users may value response accuracy and conversational fluency; for a medical assistant, they may prioritize clinical accuracy and explanation quality. We express user utility as $U(x(m); z)$, where $z \in \mathcal{Z}$ captures user/task context that mediates how technical capabilities translate into perceived value.

\textit{Business leaders} focus on managing the total resources required for deployment. We assume their primary concern is the lifecycle cost \(c(m)\). Accordingly, we represent business evaluation functions as \(B_j(c(m); w)\) for \(j \in \{0,1,\dots,J\}\), where \(w \in \mathcal{B}\) encodes environment parameters such as compute prices, service-level requirements, and budget limits. The function \(B_0(c(m);w)\) captures the baseline cost of adopting model \(m\), while constraints \(B_j(c(m);w) \le 0\) for \(j \in \{1, \dots J\} := [J]\) capture binding resource and operational requirements. In practice, organizations may also impose deployment policies---e.g., maximum acceptable latency for real-time applications or minimum throughput for high-volume services---which we treat in the next paragraph as \emph{internal} compliance requirements alongside externally imposed regulatory constraints.

\textit{Regulators} (and, in many deployments, internal governance functions) impose minimum acceptable standards (compliance thresholds) to ensure safe and ethical deployment. For example, in healthcare, models may be required to meet minimum performance and safety thresholds and avoid disparate impact. We express such requirements as $R_k(x(m); r)$ for $k \in [K]$, where $r \in \mathcal{R}$ specifies the applicable regime; constraints take the form $R_k(x(m); r) \leq 0$.

Our framework does not restrict the functional form of these evaluation functions: e.g., Section~\ref{sec:stylized-model} analyzes a linear specification for the utility function for tractability, but Section~\ref{sec:casestudies} estimates flexible (potentially nonlinear) utility mappings within the empirical pipeline.





\subsection{The Technological Frontier}\label{ssec:frontier}
A key constraint in model selection is that not all combinations of internal measures are technologically feasible. Current AI technology imposes a frontier that bounds achievable capability profiles at different resource levels. For LLMs, this often appears as trade-offs between model scale and latency, between safety alignment and capability preservation, and between general knowledge and specialized expertise.

We formalize this constraint by introducing a technological frontier $F(x(m), c(m))$ that couples capability and deployment resources. Feasible models satisfy the ``frontier constraint''
\(
F(x(m), c(m)) \leq 0 \ := \  F_X(x(m)) \leq F_C(c(m)), 
\)
where $F_X(x(m))$ aggregates performance measures into a scalar index of technological difficulty and $F_C(c(m))$ maps resources into an attainable capability level.
This frontier constraint ensures that models with better performance across multiple dimensions require proportionally greater resources. 

To make this relationship concrete, we adopt parametric forms on each side. On the capability side, we use a constant-elasticity-of-substitution (CES) aggregator
\(
F_X(x) \;=\; \Bigg(\sum_{i=1}^I a_i x_i^{\,b}\Bigg)^{\!1/b},
\)
where weights $a_i>0$ with $\sum_{i=1}^I a_i=1$ capture the relative contribution of each dimension and $b$ governs substitutability across dimensions. 
In the structural analysis we focus on the empirically most relevant regime $b \geq 1$ (weak substitutability), though the framework allows broader values (empirically, we also observe values with $0<b<1$, albeit less frequently). 
CES functions are widely used in economics to model multi-dimensional trade-offs because they impose a constant elasticity of substitution across inputs while allowing rich curvature and relative weighting \citep{solow1956contribution,arrow1961capital}.
On the cost side, we use a power function
\(
F_C(c) \;=\; c_0 c^{\,d},
\)
which maps total resource investment into an attainable capability level and, in the empirically relevant case $0<d\leq1$, captures diminishing returns to aggregate spend.

In practice, the capability--cost relationship is not static: new model releases and pricing changes shift the technological frontier over time. In our framework, this corresponds to time-varying frontiers $F_X^t,F_C^t$, with the feasible set at date $t+1$ expanding or reshaping relative to date $t$. The empirical frontier $(\widehat{F}_X,\widehat{F}_C)$ and the resulting recommendations should therefore be interpreted as \emph{snapshots} conditional on the set of available models and costs at a given point in time, to be re-estimated as technology and pricing evolve. 
A natural extension is to embed MLC in a multi-period setting where $F_X^t,F_C^t$ evolve and platform or switching costs (e.g., between different provider ecosystems) enter via $B_0$ or $B_j$, allowing deployers to trade off current performance against the option value of future improvements and the cost of moving across platforms. We view the explicit modeling of dynamic frontier evolution and ecosystem lock-in as an important direction for future work, complementary to the static, cross-sectional analysis developed here.


\subsection{Optimization Formulation}\label{ssec:optimization_formulation}
Putting these components together, the deployer selects a model that maximizes utility net of a cost penalty while satisfying cost and compliance requirements within the technological frontier:
\begin{equation} \label{eq:general_problem}
\begin{aligned}
    \max_{m \in \mathcal{M}} \quad & U(x(m); z) - \lambda B_0(c(m);w) \\ 
    \text{subject to} \quad & B_j(c(m); b) \leq 0, \quad \forall j \in [J], \\
                   & R_k(x(m); r) \leq 0, \quad \forall k \in [K], \\
                   & F(x(m), c(m)) \leq 0.
\end{aligned}
\end{equation}
The parameter $\lambda > 0$ represents the organization’s cost sensitivity, balancing performance gains against resource expenditure: cost-insensitive deployments (e.g., safety-critical settings) correspond to smaller $\lambda$, while cost-sensitive deployments correspond to larger $\lambda$. The frontier constraint enforces feasibility by coupling attainable internal measures $x(m)$ to resource use $c(m)$. This formulation forms the analytical core of MLC.



\section{Structural Analysis} \label{sec:stylized-model}
This section analyzes a tractable version of the general optimization problem to derive structural insights about optimal model selection. By specializing to linear utility functions and a parametric technological frontier, we characterize closed-form solutions that reveal how cost and compliance requirements, and the technological environment jointly determine optimal configurations. These analytical results provide intuition and guide the empirical implementation in subsequent sections. We defer all proofs to Section~\ref{ec:proofs} of the EC.

\subsection{Assumptions and Stylized Model}\label{ssec:assumptions}
To obtain analytical tractability while preserving the essential trade-offs in model selection, we impose the following structural assumptions:
\begin{assumption}[Utility]\label{assump:utility}
Utility is linear in the internal measures $x\in[0,1]^I$:
\(
U(x)\;=\;\sum_{i=1}^{I}\beta_i\,x_i,\text{ with } \beta_i>0,\ \sum_{i=1}^{I}\beta_i=1.
\)
\end{assumption} 
\begin{assumption}[Cost and budget]\label{assump:cost}
Lifecycle cost $c\in[0,1]$ is penalized linearly in the objective and a single exogenous cap $B \in (0,1]$ applies:
\(
B_0(c) = c,\ B_1(c) = c - B.
\)
\end{assumption}
\begin{assumption}[Compliance thresholds]\label{assump:reg}
Each internal measure must exceed an exogenous compliance minimum $R_i\in[0,1)$:
\(
R_i(x) = R_i - x_i.
\)
\end{assumption}
\begin{assumption}[Frontier attainability]\label{assump:frontier}
Feasible internal measures satisfy the frontier constraint
\(
F_X(x) \leq F_C(c), \ 
\text{where} \ 
F_X(x) = \Big(\sum_{i=1}^{I} a_i\,x_i^{\,b}\Big)^{\!1/b} \ 
\text{and} \ 
F_C(c) = c_0 c^{\,d},
\)
with weights $a_i>0$, $\sum_{i=1}^{I}a_i=1$, and $c_0>0$, and exponents
$b\geq1$ and $0<d\leq1$.
\end{assumption}
\begin{assumption}[Nondegeneracy]\label{assump:nondeg}
The frontier strictly dominates the compliance minimums at the budget cap
\(
\sum_{i=1}^I a_i R_i^{\,b} < c_0^{\,b} B^{\,bd},
\)
and at least one $R_i>0$.
\end{assumption}
Taken together, Assumptions~\ref{assump:utility}--\ref{assump:nondeg} provide a well-structured optimization problem linking preferences, resource allocation (costs and compliance), and technological limitations. 
Assumption~\ref{assump:utility} specifies a representative user whose linear utility captures adoption preferences.
Assumption~\ref{assump:frontier} defines a smooth technological frontier relating attainable internal measures $x$ to available spend $c$: the parameter $b\ge1$ governs substitutability across dimensions, while $0<d\le1$ ensures
diminishing returns to investment. 
Finally, Assumption~\ref{assump:nondeg} requires that the frontier strictly dominates the compliance minimums at the budget cap, guaranteeing the existence of a strictly feasible point and ruling out degenerate boundary
cases.
Under Assumptions~\ref{assump:utility}--\ref{assump:nondeg}, Problem~\ref{eq:general_problem} reduces to:
\begin{equation}\label{eq:stylized-problem}
\begin{aligned}
    \max_{x,\, c} & \quad \sum_{i=1}^I \beta_i x_i \;-\;\lambda c \quad
    \text{s.t.} \quad R_i \;\le\; x_i \;\le\; 1, \ \forall i \in [I], 
    \quad 0 \;\le\; c \;\le\; B, 
    \quad \Big(\sum_{i=1}^I a_i x_i^{\,b}\Big)^{1/b} \;\le\; c_0 c^{d}.
\end{aligned}
\end{equation}

\subsection{Optimal Solution Characterization} \label{ssec:structuralresults} 
Before characterizing the solution, we note two structural properties of Problem~\ref{eq:stylized-problem}. First, under Assumptions~\ref{assump:utility}--\ref{assump:nondeg}, Problem \ref{eq:stylized-problem} is a convex optimization problem satisfying Slater’s condition and therefore admits a well-defined global optimum. Second, any optimum necessarily lies exactly on the technological frontier: relaxing the frontier would either increase utility or reduce cost, so the constraint must be binding. We formalize both statements in Section~\ref{ec:proofs} of the EC. These properties allow us to focus directly on the optimality conditions on the frontier, yielding the following closed-form characterization:
\begin{theorem}[Optimal Solution Characterization]\label{thm:optimal-solution}
Under Assumptions~\ref{assump:utility}--\ref{assump:nondeg}, any optimal solution $(x^*, c^*)$ to Problem~\eqref{eq:stylized-problem} satisfies, for each $i\in[I]$,
\begin{equation*}
x_i^* \;=\;
\begin{cases}
    R_i, \quad \text{if } \beta_i \le \mu_0 a_i b R_i^{\,b-1}, \\[6pt]
    1,   \quad \text{if } \beta_i \ge \mu_0 a_i b, \\[6pt]
    \left(\dfrac{\beta_i}{\mu_0 a_i b}\right)^{\tfrac{1}{b-1}}, \quad \text{otherwise},
\end{cases}
\quad \text{and} \qquad
c^* \;=\;
\begin{cases}
    B, \quad \text{if } \mu_0 c_0^{\,b} b d\, B^{\,bd-1} \;\ge\; \lambda, \\[6pt]
    \left(\dfrac{\lambda}{\mu_0 c_0^{\,b} b d}\right)^{\tfrac{1}{bd-1}}, \quad \text{otherwise},
\end{cases}
\end{equation*}
where $\mu_0 > 0$ is the Lagrange multiplier associated with the frontier constraint.
If in addition $b>1$, then the optimal solution $(x^*,c^*)$ is unique.
\end{theorem}
Theorem~\ref{thm:optimal-solution} reveals that optimal internal measures $x^*$ partition into three distinct regimes based on the ratio of user utility to technological difficulty. Internal measure dimensions with low utility relative to their technological cost ($\beta_i \le \mu_0 a_i b R_i^{b-1}$) remain at compliance minimums, reflecting compliance-driven rather than value-driven allocation. Conversely, dimensions with high utility-to-difficulty ratios ($\beta_i \ge \mu_0 a_i b$) saturate at their maximum feasible levels. Between these extremes, dimensions take interior values that balance marginal utility against marginal technological cost. This creates a natural taxonomy of measures: those driven by user preferences versus those maintained only for compliance. For regulators, this suggests that requirements on low-utility measures impose the highest shadow costs on organizations.

The optimal resource level $c^*$ follows analogous logic: organizations either exhaust their budget when the marginal return to investment exceeds the cost sensitivity $\lambda$, or choose an interior spending level that equates marginal benefit to marginal cost. When $\lambda$ is low (indicating cost-insensitive environments like safety-critical applications), organizations use their full budget. When $\lambda$ is high (cost-sensitive environments), they may optimally underspend. 
Finally, $\mu_0$ is the shadow value of the technological frontier: it measures how ``expensive'' it is, in terms of utility, to push performance further, and thus determines which dimensions are worth improving and which stay at their bounds.

\subsection{Comparative Statics} \label{ssec:comp-stat}
Having characterized its structure, we next study how the optimal solution $(x^*,c^*)$ varies with changes in economic, compliance, and technological parameters. The resulting comparative statics provide actionable insights for stakeholders.
We consider local perturbations under a standard fixed-active-set assumption (FAS), i.e., the set of binding and nonbinding constraints remains unchanged in a small neighborhood of the optimum. We denote by
$S := \{\,i : R_i < x_i^* < 1\,\}$ the interior dimensions and assume that $S \not= \emptyset$,
by $W^* := \sum_{j=1}^I a_j (x_j^*)^b,$ the total frontier mass, and
by $Y^* := \sum_{j\in S} a_j (x_j^*)^b,$ its interior component. 
\begin{proposition}[Budget Sensitivity Analysis]\label{thm:budget-sensitivity}
Fix all parameters except $B$ and let FAS hold as $B$ varies.
Assume the budget constraint binds, $c^* = B>0$. 
Then, for each internal measure,
\[
\frac{\partial x_i^*}{\partial B}
=
\begin{cases}
\varepsilon_B\,\dfrac{x_i^*}{B}, & i\in S,\\[6pt]
0, & i\notin S,
\end{cases}
\qquad \varepsilon_B =  d\,\dfrac{W^*}{Y^*},
\]
where ${\partial \ln x_i^*}/{\partial \ln B} = \varepsilon_B$ is a common elasticity shared by all $i\in S$.
In particular, all interior measures respond proportionally equally to
marginal increases in $B$. 
If $S=[I]$, then $\varepsilon_B = d$.
\end{proposition}
This proposition establishes a symmetry result: under a fixed active set, all interior dimensions respond to budget in a proportional way, regardless of their utility weights $\beta_i$ or technological weights $a_i$. Every flexible (interior) measure shares the same elasticity $\varepsilon_B$, while bound measures do not respond at all. The magnitude of $\varepsilon_B$ depends on how much of the frontier ``mass'' lies in interior coordinates ($W^*/Y^*$): when few dimensions remain flexible, each absorbs a larger share of the marginal budget increase. Thus, small budget changes produce balanced growth across all interior measures, but the strength of this response depends on how many measures are already constrained. Once the active set changes, the elasticity must be recomputed. This proportional scaling property has important implications for investment decisions: organizations cannot use budget increases to selectively improve specific measures as the technology frontier enforces balanced growth. 

\begin{proposition}[Regulatory stringency]\label{prop:regulatory-impact}
For any $k \in [I]$, fix all parameters except $R_k$ and let FAS hold as $R_k$ varies. 
Let
\(
S' := S \setminus\{k\}.
\)
Assume $b>1$, $bd\not=1$.\footnote{These conditions rule out degenerate cases, ensuring cleaner expressions; the qualitative insights remain unchanged without them.}
Then:
\begin{enumerate}
\item \textbf{Direct effect.}
For the directly affected internal measure dimension $k$,
\[
\frac{\partial x_k^*}{\partial R_k}
=
\begin{cases}
1, & x_k^* = R_k,\\[3pt]
0, & x_k^* > R_k.
\end{cases}
\]

\item \textbf{Spillovers and cost when the budget binds.}
Suppose $x_k^* = R_k$ and $c^* = B$. 
Then,
\[
\frac{\partial x_i^*}{\partial R_k} = -\,\frac{a_k R_k^{\,b-1} x_i^*}{Y^*} \quad \text{for each $i\in S'$,}
\qquad 
\frac{\partial x_i^*}{\partial R_k}=0  \quad \text{for all $i\notin S'$,}
\qquad 
\frac{\partial c^*}{\partial R_k} = 0.
\]
If instead $x_k^*>R_k$ while $c^*=B$, then
$\frac{\partial x_i^*}{\partial R_k} = 0$ for all $i$ and
$\frac{\partial c^*}{\partial R_k} = 0$.

\item \textbf{Spillovers and cost when the budget is slack.}\footnote{Detailed expressions for case (iii) are provided in the EC; here we report only the resulting sign patterns.}
Suppose $x_k^* = R_k$ and $c^*<B$. 
Then, all interior measures $i\in S'$ adjust in the same direction: $\partial x_i^*/\partial R_k>0$ when $bd>1$ and $\partial x_i^*/\partial R_k<0$ when $bd<1$. 
For all binding measures $i\notin S'$, $\partial x_i^*/\partial R_k=0$.
For the cost $c$, $\frac{\partial c^*}{\partial R_k} \geq 0$, 
with $\frac{\partial c^*}{\partial R_k} = 0 \Leftrightarrow d=1$.
If instead $x_k^*>R_k$, then $\frac{\partial x_i^*}{\partial R_k} = 0$ for all $i$ and
$\frac{\partial c^*}{\partial R_k} = 0$.
\end{enumerate}
\end{proposition}
This result highlights a spillover effect in regulated ML systems: tightening standards for one measure forces organizations to reduce performance on other measures when operating at the technology frontier. This spillover occurs because improving any measure consumes scarce resources, whether computational, data, or human expertise.
In particular, the directly constrained measure moves one-for-one with the new compliance threshold. The adjustment propagates to other measures through the technological frontier: when the budget binds, all interior measures are pushed downward in proportion to their contribution to the frontier, while under a slack budget they all move in the same direction, with the sign determined by the frontier curvature ($bd>1$ or $bd<1$). Bound dimensions never respond. Finally, when the budget is slack, the organization must spend more to remain feasible, so the optimal cost weakly increases with regulatory tightness. 
For regulators, these effects show that single-metric compliance requirements have system-wide consequences that must be anticipated, and their direction and magnitude are dictated by the geometry of the frontier. For organizations, these results quantify the hidden costs of compliance and suggests that regulatory changes require reoptimization of the entire model portfolio, not just the directly affected measure.

\begin{proposition}[Technological Progress]\label{thm:technology-progress}
Suppose $c^* = B>0$ and $b>1$.
\begin{enumerate}
\item[\textup{(i)}] \textbf{Impact of frontier level $c_0$.}
Fix all parameters except $c_0$ and let FAS hold. Then, 
\[
\frac{\partial x_i^*}{\partial c_0}
=
\begin{cases}
\displaystyle \varepsilon_c\,\dfrac{x_i^*}{c_0}, & i\in S,\\[6pt]
0, & i\notin S,
\end{cases}
\qquad
\varepsilon_c = \dfrac{W^*}{Y^*},
\]
so all interior coordinates share the same elasticity
$\partial \ln x_i^*/\partial \ln c_0 = \varepsilon_c$.

\item[\textup{(ii)}] \textbf{Impact of returns to scale $d$.}
Fix all parameters except $d$ and let FAS hold. Then,
\[
\frac{\partial x_i^*}{\partial d}
=
\begin{cases}
\varepsilon_d\,x_i^*, & i\in S,\\[6pt]
0, & i\notin S,
\end{cases}
\qquad
\varepsilon_d = \dfrac{W^*}{Y^*}\,\ln B,
\]
so all interior coordinates share the same semi--elasticity
$\partial \ln x_i^*/\partial d = \varepsilon_d$.

\item[\textup{(iii)}] \textbf{Impact of substitutability $b$.}
Fix all parameters except $b$ and let FAS hold. 
Then, for any measures $i,j\in S$,
\(
x_i^* > x_j^*
\quad\Longrightarrow\quad
\frac{\partial \ln x_i^*}{\partial b}
<
\frac{\partial \ln x_j^*}{\partial b},
\)
i.e., changes in $b$ tend to reallocate performance from larger interior measures toward smaller ones.
For all binding measures $i\notin S$, $\partial x_i^*/\partial b = 0$. 
The cost $c$ satisfies $\partial c^*/\partial b = 0$.
\end{enumerate}
\end{proposition}
In this proposition, we distinguish between three types of technological progress with different implications on the optimal solution.
First, raising the frontier level $c_0$ uniformly scales all interior dimensions, since a higher attainable frontier relaxes technological constraints symmetrically across coordinates. This represents breakthrough innovations, such as moving from GPT-3 to GPT-4. 
Second, increasing the returns exponent $d$ amplifies the value of budget, generating proportional gains for all interior measures, with magnitude governed by $\ln B$. This corresponds to a scenario where investments become more productive, particularly benefiting organizations with larger budgets.
Third, increasing substitutability $b$ redistributes performance within the interior set: higher-performing dimensions contract while lower-performing ones expand, producing a systematic rebalancing rather than uniform growth. As ML techniques make different capabilities more interchangeable, resources flow from over-performing measures to under-performing ones. This rebalancing effect suggests that technological advances that increase flexibility (higher $b$) could paradoxically reduce peak performance on any single metric while improving overall system utility. 

\subsection{Managerial Takeaways}\label{ssec:managerial-takeaways}
The structural results imply certain patterns for how capability dimensions and cost respond to changes in preferences, cost and compliance requirements, and technology. At the optimum, each dimension is either held at a compliance floor, pushed to a ceiling, or used as a trade-off variable; marginal budget changes scale all interior dimensions proportionally when the budget binds; tightening a binding compliance threshold creates spillovers on other interior dimensions and, when budgets are slack, typically requires additional spend; and different forms of technological progress shift the optimal profile in systematic ways. Table~\ref{tab:ms_takeaways_quickref} summarizes these comparative statics as levers, their local effects at the optimum, and how they should be read in deployment decisions.
\begin{table}[t]
\centering
\caption{Summary of levers, their effects on the optimal solution, and their practical implications.}
\label{tab:ms_takeaways_quickref}
\small 
\setlength{\tabcolsep}{4pt} 
\renewcommand{\arraystretch}{1.1} 
\resizebox{\linewidth}{!}{%
\begin{tabular}{p{3.2cm} p{6.2cm} p{5.2cm}}
\toprule
\textbf{Lever / situation} & \textbf{What moves at the optimum} & \textbf{Practical implication} \\
\midrule
Low vs.\ high $\beta_i/a_i$  “value-to-difficulty’’ &
$x_i$ goes to {floor} ($R_i$), {interior}, or {ceiling} ($1$) &
Identify and manage three regimes: compliance, trade-off, and saturation dimensions \\

Resource cap $B\uparrow$ (when $c^*=B$) &
All interior $x_i$ scale up by the \emph{same \%}; bounds unchanged &
Additional budget yields broad upgrades; targeted gains require altering constraints or technology \\

Compliance $R_k\uparrow$ \  ($x_k^*=R_k$, $c^*=B$) &
Interior $x_i\downarrow$ (spillover); $c$ unchanged &
Compliance changes affect the whole profile; expect trade-offs across capabilities \\

Compliance $R_k\uparrow$ \  ($x_k^*=R_k$, $c^*<B$) &
Interior $x_i$ move together; $c^*$ weakly $\uparrow$ (often strictly if $d<1$) &
Tighter standards typically require more spend or relaxing other requirements \\

Frontier level $c_0\uparrow$ (when $c^*=B$)&
Interior $x_i\uparrow$ proportionally; $c^*$ unchanged &
Technology shifts can raise multiple capability dimensions simultaneously \\

Returns to spend $d\uparrow$ (when $c^*=B$) &
Interior $x_i$ increase; effect scales with $\ln B$ &
More efficient spend amplifies budget impact; high-budget deployments benefit the most \\

Substitutability $b\uparrow$ (when $c^*=B$)&
Interior profile rebalances &
Greater substitutability reduces specialization; single-metric rankings become less informative \\
\bottomrule
\end{tabular}}
\end{table}

\section{From Model to Practice: the MLC Pipeline}\label{sec:compasspipeline}
In this section, we operationalize MLC as a practical \emph{decision-support pipeline}.
Specifically, in Sections~\ref{sec:problem-formulation}--\ref{sec:stylized-model}, we formulated MLC as an optimization framework for AI deployment decisions: the deployer trades off deployment net value against cost and compliance constraints while respecting a technological frontier that links capabilities to deployment cost. 
Turning that framework into a deployable decision-support tool raises two practical questions.
First, the key objects that define the optimization problem---the frontier and the utility function---are not known a priori and must be estimated from data in the deployment environment.
Second, even when the optimization yields an ideal target capability--cost profile $(x^*,c^*)$, a deployer still needs to map that target to an actual model choice from the available model set. 

This section answers these questions by presenting a practical pipeline that an organization can run with its own data to produce target profiles and model recommendations. The same workflow can also be used to build a \emph{deployment-aware leaderboard}: when cost and compliance constraints are interpreted as sector-level requirements and inputs come from public benchmarks and pricing, the pipeline ranks models by expected deployment value under explicit constraints.
In what follows, we first give a system-level overview of the pipeline and then provide details on each component. In Section \ref{sec:casestudies}, we present a concrete instantiation of the pipeline in two distinct deployment contexts.




\subsection{Pipeline Overview}\label{ssec:pipeline-overview}
The MLC pipeline takes four classes of \textbf{inputs}:
(i) \emph{Capability Data} $x_{\mathrm{raw}}$ (raw model descriptors and benchmark results),
(ii) \emph{Cost Data} $c_{\mathrm{raw}}$ (deployment resource use or pricing),
(iii) \emph{Outcome Data} $(z,y)$ (application-level outcomes from users or experts, indexed by user/task context $z$),
and (iv) \emph{Cost and Compliance Requirements} (constraints imposed by the organization or by regulators).
Given these inputs, MLC consists of four components:
\begin{itemize}
    \item \textbf{Measure extraction ($x_{\mathrm{raw}}\rightarrow x$).}
    Converts heterogeneous, high-dimensional capability signals into a small set of internal measures $x$ that can be compared consistently across models.

    \item \textbf{Frontier estimation ($(x,c)\rightarrow \widehat{F}$).}
    Estimates an empirical capability--cost frontier $\widehat{F}$ describing which capability profiles are feasible at different deployment costs in the current model landscape.

    \item \textbf{Utility estimation ($(x,z,y)\rightarrow \widehat{U}$).}
    Learns how internal measures translate into application outcomes, allowing the mapping $\widehat{U}(x;z)$ to vary across user/task contexts $z$.

    \item \textbf{Optimization and model recommendation ($\widehat{U},\widehat{F}$ + constraints $\rightarrow (x^*,c^*)$ and $m^*$).}
    Solves the deployer’s constrained problem to obtain an ideal target profile $(x^*,c^*)$ and then selects the ``closest'' feasible model(s) from the available set.
\end{itemize}
Figure~\ref{fig:compass-flowchart} summarizes the MLC pipeline. 
For the deployment-aware leaderboard, the same steps are run using public capability/cost inputs together with a specified set of cost and compliance constraints. The resulting objective values and recommended models induce a ranking that is explicitly tied to deployment requirements rather than to raw capability scores alone.

\newcommand{\boxwidth}{3.2cm}
\begin{figure}[t]
\centering
\resizebox{\linewidth}{!}{%
\begin{tikzpicture}[
    x=1.5cm,y=1.5cm,
    >=Latex,
    font=\small,
    input/.style={
        trapezium, trapezium left angle=70, trapezium right angle=110,
        draw, align=center,
        minimum width=\boxwidth,
        text width=\boxwidth,
        minimum height=0.9cm
    },
    process/.style={
        ellipse, draw, thick, align=center,
        minimum width=\boxwidth,
        text width=\boxwidth,
        minimum height=0.9cm
    },
    decision/.style={
        diamond, draw, thick, aspect=2, align=center,
        inner sep=1pt,
        minimum height=1.5cm
    },
    output/.style={
        trapezium, trapezium left angle=70, trapezium right angle=110,
        draw, align=center,
        minimum width=\boxwidth,
        text width=\boxwidth,
        minimum height=0.9cm
    },
    groupbox/.style={dashed, rounded corners, draw, inner sep=0.35cm}
]

\node[input] (cap)   at (0,0)   {Capability Data\\$x_{\text{raw}}$};
\node[input] (cost)    at (3,0)   {Cost Data\\$c_{\text{raw}}$};
\node[input] (out)   at (6,0)   {Outcome Data\\$y$};
\node[input] (con)   at (9,0)   {Cost/Compliance\\$B_j,\,R_k$};

\node[groupbox, fit=(cap) (cost) (out) (con),
      label=above:{\normalsize Inputs}] (inputsbox) {};

\node[process] (meas) at (0,-1.5)
    {Measure Extraction: $x$};

\node[process] (frontier) at (3,-2)
    {Frontier Est.: $\widehat{F}$};

\node[process] (utility)  at (6,-2.5)
    {Utility Estimation: $\widehat{U}$};

\node[decision] (opt) at (9,-3)
    {Optimization:\\$x^*$, $c^*$, $m^*$};



\draw[->, thick]
    (cap.south) -- (meas.north);

\draw[->, thick]
    (cost.south) -- (frontier.north);

\draw[->, thick]
    (out.south) -- (utility.north);

\draw[->, thick]
    (con.south) -- (opt.north);


\draw[->, thick]
    (meas.south) -- ++(0,-0.2) -- (frontier.west);

\draw[->, thick]
    (meas.south) -- ++(0,-0.7) -- (utility.west);


\draw[->, thick]
    (frontier.south) -- ++(0,-0.7) -- (opt.west);

\draw[->, thick]
    (utility.south) -- ++(0,-0.165) -- (opt.west);

\end{tikzpicture}
}
\caption{MLC pipeline.} 
\smallskip \footnotesize \parbox{0.95\linewidth}{\textit{Note.} 
Capability data $x_{\mathrm{raw}}$ are mapped into internal measures $x$. The frontier estimator uses $(x,c)$ to estimate the capability--cost frontier $\widehat{F}$. The utility estimator uses $(x,z,y)$ to estimate the context-dependent utility function $\widehat{U}(x;z)$. Cost and compliance constraints feed into the optimization, which outputs target profiles $(x^*,c^*)$ and model recommendations $m^*$.
}
\label{fig:compass-flowchart}
\end{figure}
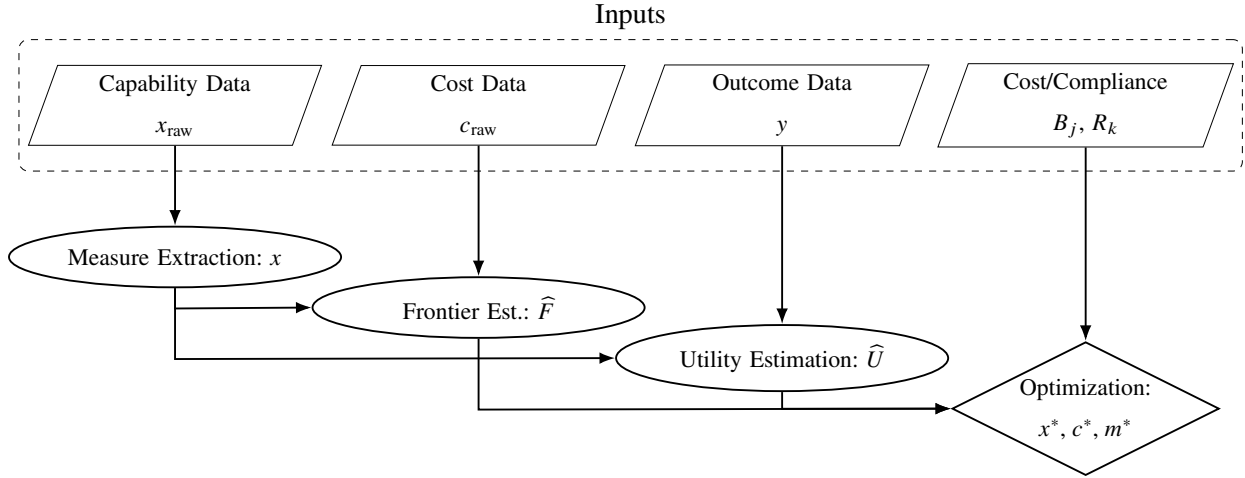


\subsection{Inputs and Data Collection}\label{ssec:inputs}
Implementing MLC in a given deployment context requires four families of inputs. We keep the notation from Section~\ref{sec:problem-formulation} and define each input in a way that can be collected in practice.
\begin{itemize}
    \item \textbf{Capability Data $x_{\mathrm{raw}}(m)$.}
    For each candidate model $m\in\mathcal{M}$, the deployer collects a vector of model-level descriptors
    $x_{\mathrm{raw}}(m)\in\mathbb{R}^{d_{\mathrm{raw}}}$ that summarize how the model behaves \emph{independently of a specific user interaction}.
    These may include generic benchmark scores (e.g., reasoning, coding, factuality), and technical metrics (e.g., throughput or latency statistics) when they are available and comparable across models \citep{LLMLeaderboard}.

    \item \textbf{Cost Data $c_{\mathrm{raw}}(m)$.}
    MLC uses a scalar cost $c(m)\in\mathbb{R}_+$ inside the optimization, but in practice costs are recorded as raw components
    $c_{\mathrm{raw}}(m)$ (e.g., price per input/output token, measured latency, hardware requirements, licensing fees).
    A deployer then maps these components into the scalar $c(m)$ that reflects the deployment setting.
    A practical choice is \emph{expected cost per interaction} under a workload distribution,
    \(
        c(m)\;=\; p_{\text{tok}}(m)\cdot \mathbb{E}\big[\text{tokens}(m)\mid \text{workload}\big]\;+\;c_{\text{fixed}}(m),
    \)
    where $p_{\text{tok}}(m)$ is the per-token price and $\mathbb{E}[\text{tokens}(m)\mid \text{workload}]$ captures \emph{token efficiency} (how many tokens the model tends to use to complete the organization’s typical tasks). 

    \item \textbf{Outcome Data $(i,m_i,z_i,y_i)$.}
    To learn how internal measures translate into value in the target application, the deployer assembles interaction-level data indexed by $i\in\mathcal{I}$.
    Each interaction records the model used $m_i$, contextual features $z_i$ (user type, task type, or risk tier), and an external outcome $y_i$ (e.g., a user choice or rating, a success indicator, or an expert rubric score).

    \item \textbf{Cost and Compliance Data $(B_j,R_k)$.}
    The deployer specifies two families of constraints in functional form.    
    \emph{Cost constraints} $B_j$ are imposed by the organization on deployment resources, and are stated in terms of internal requirements (e.g., maximum spend). 
    These apply to $c(m)$ via constraints of the form $B_j\big(c(m);w\big)\le 0$.
    \emph{Compliance constraints} $R_k$ encode minimum acceptable standards (e.g., safety or quality thresholds) that must be satisfied and can be imposed either internally by the organization or externally by regulators. 
    When imposed internally by the organization, compliance requirements are expressed on the extracted internal measures $x(m)$ (e.g., a policy that the {safety factor} must exceed a threshold), i.e., $R_k\big(x(m);r\big)\le 0 .$  
    When imposed by regulators, they are stated on {observable, auditable} raw measures $x_{\mathrm{raw}}(m)$ (e.g., in healthcare, requiring the deployed model to exceed a minimum score on a clinical safety benchmark), i.e., $R_k\big(x_{\mathrm{raw}}(m);r\big)\le 0 .$
\end{itemize}
In summary, $x_{\mathrm{raw}}(m)$ describes \emph{what a model can do}, $c_{\mathrm{raw}}(m)$ describes \emph{what it costs to run under a workload}, $(i,m_i,z_i,y_i)$ describes \emph{how model capabilities translate into outcomes in the target setting}, and $(B_j,R_k)$ specifies \emph{the organization’s feasible operating region}. The next subsections describe how MLC turns these inputs into internal measures, an empirical frontier, an estimated utility function, and model recommendations.


\subsection{Measure Extraction}\label{ssec:measure-extractor}
The goal of the Measure Extraction step is to convert heterogeneous raw capability descriptors
$x_{\mathrm{raw}}(m)\in\mathbb{R}^{d_{\mathrm{raw}}}$ into a small set of internal measures
$x(m)\in\mathbb{R}^I$ that are (i) stable to estimate with the available number of models,
(ii) interpretable for deployment decisions, and (iii) usable in the frontier and utility
modules that follow.
This step is important in practice because $d_{\mathrm{raw}}$ can be large and highly
collinear (many benchmarks measure overlapping skills), and frontier and utility estimation in a
high-dimensional space is statistically and computationally unstable.

Let $X_{\mathrm{raw}}\in\mathbb{R}^{|\mathcal{M}|\times d_{\mathrm{raw}}}$ be the matrix obtained by stacking
$x_{\mathrm{raw}}(m)$ across models $m\in\mathcal{M}$, and let $\widetilde{X}_{\mathrm{raw}}$ denote the
column-standardized version (zero mean, unit variance).
We fit a factor model of the form
\(
\widetilde{X}_{\mathrm{raw}} \;=\; F L^\top + E,
\)
where $F\in\mathbb{R}^{|\mathcal{M}|\times I}$ contains the latent factor scores (one row per model),
$L\in\mathbb{R}^{d_{\mathrm{raw}}\times I}$ is the loading matrix, and $E$ is the residual.
We choose the number of factors $I$ using standard criteria (explained variance and scree
plots), with the practical objective of keeping $I$ small enough to support interpretability robust frontier and utility
estimation. %

To make the factors interpretable, we apply a rotation and a sparsification step to $L$
so that each raw measure loads primarily on one factor. We then define the internal measures
for each model $m$ as its factor-score vector,
\(
x(m) \;:=\; \big(x_1(m),\dots,x_I(m)\big),
\)
where $x_i(m)$ is the score of model $m$ on factor $i$.
Finally, for comparability across dimensions and to support downstream thresholds,
we rescale each coordinate across models to lie in $[0,1]$ (min--max scaling within $\mathcal{M}$).\footnote{We use $[0,1]$ scaling
to keep the internal-measure geometry stable and to make constraint thresholds easy to interpret.}

If the raw capability vector $x_{\mathrm{raw}}$ is already low-dimensional, well-measured, and
approximately non-redundant, the deployer can skip this step and set $x(m)=x_{\mathrm{raw}}(m)$.
In most LLM settings, however, measure extraction improves stability and
interpretability for the frontier and utility estimation stages, especially given the large number and  overlapping nature of available benchmarks.

\subsection{Frontier Estimation}\label{ssec:frontier-estimation}
The goal of the Frontier Estimation step is to construct an empirical analogue of the technological constraint
introduced in Section~\ref{ssec:frontier}: not all capability profiles are feasible at a
given deployment resource level. Using the observed model pairs
$\{(x(m),c(m))\}_{m\in\mathcal{M}}$, we estimate functions $(\widehat{F}_X,\widehat{F}_C)$ such that
\(
\widehat{F}_X\!\big(x(m)\big)\ \le\ \widehat{F}_C\!\big(c(m)\big)
\)
for technologically feasible (efficient) models, 
where $\widehat{F}_X$ aggregates multi-dimensional internal measures into a scalar capability
index and $\widehat{F}_C$ maps resources into an attainable capability level. 

\paragraph{Inputs.}
The Frontier Estimation step takes as input:
(i) internal measures $x(m)\in\mathbb{R}^I_+$
and (ii) a scalar resource index $c(m)\in\mathbb{R}_+$ for each model.
If multiple resource metrics are available (e.g., price and latency), one can either (i) form a single
index $c(m)$ for frontier estimation and keep the remaining metrics as separate cost
constraints in the optimization layer, or (ii) estimate separate frontiers for different
deployment regimes (e.g., latency-constrained vs.\ throughput-constrained).

\paragraph{Frontier specification.}
To maintain interpretability and following the discussion in
Section~\ref{ssec:frontier}, we adopt a parametric form on each side:
\(
\widehat{F}_X(x)
=\Big(\sum_{i=1}^I \hat{a}_i\,x_i^{\hat{b}}\Big)^{1/\hat{b}}
\)
and
\(
\widehat{F}_C(c)
=\hat{c}_0\,c^{\hat{d}},
\)
with $\hat{a}_i>0$, $\sum_i \hat{a}_i=1$, and $\hat{b},\hat{c}_0,\hat{d}>0$.
Estimation proceeds in three main steps: we first construct ordered capability tiers via Pareto peeling in the internal-measure space, then recover the shape and tier levels of $\widehat{F}_X$ by fitting the CES aggregator on Pareto-efficient models via alternating optimization, and finally link these tier-specific capability levels to costs via a parametric fit of $\widehat{F}_C$.

\paragraph{Why Pareto filtering and tiering?}
Real model sets contain both efficient and inefficient points. A model $m$ is \emph{Pareto-dominated}
(in the capability space) if there exists another model $m'$ with $x(m')\ge x(m)$ componentwise
and $x(m')\neq x(m)$, i.e., $m'$ is at least as good on every internal measure and strictly better
on at least one. Dominated models are not informative about the technological boundary and can
bias frontier estimates inward. Moreover, with a finite and noisy set of models, especially when
$|\mathcal{M}|$ is moderate, estimating a smooth frontier benefits from first organizing models into
coarse capability \emph{tiers} and focusing on models that are approximately non-dominated within each tier.
This is the empirical motivation for Pareto ``peeling'': we iteratively extract non-dominated layers,
which is analogous to non-dominated sorting in multi-objective optimization.

\paragraph{Step 1: Pareto peeling and tier construction.}
We construct ordered tiers $t=1,\dots,T$ (weakest to strongest) using repeated Pareto peeling
in the $I$-dimensional internal-measure space:
\begin{enumerate}
\item Compute the non-dominated set (Pareto front) among remaining models.
\item Remove it (``peel'' it) and repeat to obtain successive layers.
\item Merge adjacent layers to obtain a desired number of tiers $T$.
\end{enumerate}
Within each tier $t$, we retain only the Pareto-efficient models (relative to $x$) as the
tier's empirical boundary points. We denote the retained set by $\mathcal{M}_{\text{eff}}$ and
the tier assignment by $t(m)\in[T]$.

\paragraph{Step 2: Capability-side shape estimation ($\widehat{F}_X$).}
Let $s(m;a,b) := \big(\sum_{i=1}^I a_i x_i(m)^b\big)^{1/b}$ denote the CES capability score for parameters
$(a,b)$. We associate each tier $t$ with a latent frontier level $\lambda_t$ (a scalar capability
level shared by efficient models in that tier), with $\lambda_1\le\cdots\le\lambda_T$.
We estimate $(\hat{a},\hat{b},\hat{\lambda})$ from the efficient set by solving:
\begin{equation} \label{eq:frontier-estimation}
    \min_{a,b,\lambda}\ \sum_{m\in\mathcal{M}_{\text{eff}}}
\ell\!\big(s(m;a,b)-\lambda_{t(m)}\big)
\quad\text{s.t.}\quad
a_i>0,\ \sum_i a_i=1,\ \lambda_1\le\cdots\le\lambda_T,
\end{equation}
where $\ell(\cdot)$ is the robust soft-$\ell_1$ loss to reduce sensitivity to measurement noise.
We solve Problem~\ref{eq:frontier-estimation} by alternating optimization:
(i) for fixed $(a,b)$, update $\lambda$ via isotonic regression over tiers (enforcing monotonicity);
(ii) for fixed $(b,\lambda)$, update $a$ on the simplex;
(iii) for fixed $(a,\lambda)$, update $b$ by a one-dimensional bounded search.
The output is a capability aggregator $\widehat{F}_X$ and tier levels
$\hat{\lambda}_1,\dots,\hat{\lambda}_T$.

\paragraph{Step 3: Linking tier levels to resources ($\widehat{F}_C$).}
To map tier capability levels into resources, we summarize model resources within each tier.
Let $c_t$ be a representative resource level for tier $t$ (e.g., median of
$\{c(m): t(m)=t\}$), monotone-adjusted via isotonic regression so that $c_1\le\cdots\le c_T$.
We then estimate the slope $\hat{d}$ from the log-linear relation
\(
\log \hat{\lambda}_t \approx b_0 + d \log c_t,
\)
using ordinary least squares. Given $\hat{d}$, we set the intercept as an upper envelope
\(
\hat{c}_0 \ :=\ \max_{t\in[T]}\ \frac{\hat{\lambda}_t}{c_t^{\,\hat{d}}},
\)
so that $\widehat{F}_C(c_t)=\hat{c}_0 c_t^{\hat{d}} \ge \hat{\lambda}_t$ for all tiers, i.e., the fitted cost frontier upper-envelopes the tier points.

\paragraph{Outputs.} Together, $\widehat{F}=(\widehat{F}_X,\widehat{F}_C)$ define the empirical feasibility constraint. In all, the Frontier Estimation step outputs:
(i) fitted parameters $(\hat{a},\hat{b},\hat{c}_0,\hat{d})$,
(ii) tier levels $\hat{\lambda}_1,\dots,\hat{\lambda}_T$,
(iii) tier assignments $t(m)$ and the efficient set $\mathcal{M}_{\text{eff}}$.



\subsection{Utility Estimation}\label{ssec:utility-estimation}
The goal of Utility Estimation step is to learn how the internal measures $x(m)$ translate into deployment value.
It maps the extracted capability profile $x$ into an
\emph{external} outcome $y$ that reflects performance in the target setting, allowing preferences to vary
by user or task context $z$. The output is an empirical utility function
$\widehat{U}(x;z)$ that we use as the objective term in the optimization layer.

\paragraph{Inputs.}
The Utility Estimation step takes as input interaction-level Outcome Data of the form
\(
(i, m_i, z_i, y_i) \quad \text{for } i \in \mathcal{I},
\)
where $m_i\in\mathcal{M}$ is the model used in interaction $i$, $z_i$ encodes the relevant context (user profile or task descriptors or both), and $y_i$ is an observed external outcome.
We link outcomes to internal measures via $x_i := x(m_i)$.
The outcome $y_i$ may be a binary (e.g., success vs. failure, or model chosen vs. not chosen), a continuous (ratings or rubric scores), or a rank-based signal; the only requirement is that it reflects value in the intended deployment setting.

    

\paragraph{Personalized utility via user/task context $z$.}
Utility can vary across users and tasks. We use $z_i$ to define \emph{personalized} utility estimators in
a way that remains stable with finite data.
We map each context $z$ to a group label $g(z)\in\mathcal{G}$ and estimate one utility model
per group:
\(
\mathcal{I}_g := \{i\in\mathcal{I} : g(z_i)=g\}.
\)
This produces a collection of group-specific utilities $\{\widehat{U}_g\}_{g\in\mathcal{G}}$,
which we use as $\widehat{U}(x;z) := \widehat{U}_{g(z)}(x)$.

\paragraph{Estimators for $\widehat{U}(x;z)$.}
For each group $g$, we fit a predictive model that maps internal measures to external outcomes by solving
\(
\widehat{U}_g \in \arg\min_{u \in \mathcal{U}} \ \sum_{i\in\mathcal{I}_g} \ell\!\big(y_i, u(x_i)\big),
\)
where $\ell(\cdot,\cdot)$ is a loss function measuring the prediction error of $u(x_i)$ relative to the realized outcome $y_i$ and $\mathcal{U}$ is a chosen function class. The choice of estimator depends on (i) whether we restrict $\mathcal{U}$ to be linear or allow nonlinear functions, and (ii) whether the outcome $y_i$ is continuous or binary. Specifically:
\begin{itemize}
    \item \textbf{Linear function class, continuous outcomes.}
    If $y_i\in\mathbb{R}$ and we restrict $\mathcal{U}$ to linear functions, we estimate a linear regression
    \(
    \widehat{U}_g(x)=\hat\alpha_g+\hat\beta_g^\top x.
    \)
    In this case, $\hat\beta_g$ has a direct objective interpretation in Problem \ref{eq:general_problem}: up to an additive constant, the utility term is $\hat\beta_g^\top x$, so $\hat\beta_{g,i}$ is the marginal value of improving internal measure $x_i$ for group $g$.

    \item \textbf{Linear function class, binary outcomes.}
    If $y_i\in\{0,1\}$ and we restrict $\mathcal{U}$ to linear functions, we estimate a logistic regression 
    \(
    \widehat{U}_g(x) \;=\; \sigma\!\big(\hat\alpha_g + \hat\beta_g^\top x\big)
    \text{ with } \sigma(t)=\frac{1}{1+e^{-t}}.
    \)
    The coefficients $\hat\beta_g$ are identified in log-odds units: $\hat\beta_{g,i}$ is the marginal effect of $x_i$ on the log-odds of success. In the optimization layer, we can either use $\widehat{U}_g(x)$ directly as the utility term (interpreting it as a predicted success probability), or use the linear index $\hat\alpha_g+\hat\beta_g^\top x$ as a linear surrogate objective. The surrogate preserves the ordering induced by $\widehat{U}_g$ because $\sigma(\cdot)$ is strictly increasing, and it yields an explicit weight vector $\hat\beta_g$ that makes trade-offs across internal measures transparent.

    \item \textbf{Nonlinear function class.} If we allow $\mathcal{U}$ to be nonlinear, we estimate a nonlinear predictor $\widehat{U}_g$ (in our implementations, LightGBM \citep{ke2017lightgbm}) mapping internal measures $x$ to outcomes $y$. In the optimization layer, we use $\widehat{U}_g(x)$ directly as the objective term in Problem \ref{eq:empirical_problem} (i.e., we treat $\widehat{U}_g$ as a black-box utility function), which is computationally feasible because $x$ is low-dimensional. To obtain an interpretable analogue of the ``weights'' in the linear cases, we report a normalized feature-importance vector $\hat\beta_g\in\mathbb{R}_+^I$ computed from the fitted model (in our implementations, split-gain importances) and scaled so that $\sum_{i=1}^I \hat\beta_{g,i}=1$; $\hat\beta_{g,i}$ summarizes the relative contribution of internal measure $x_i$ to the predictive utility within group $g$.

\end{itemize}

\paragraph{Outputs.} In all, the Utility Estimation step outputs the set of group-specific utilities $\{\widehat{U}_g\}_{g\in\mathcal{G}}$ and the corresponding utility weights $\{\hat{\beta}_g\}_{g\in\mathcal{G}}$.

\subsection{Optimization and Outputs}\label{ssec:optimization}
The Optimization step combines the empirical frontier
$\widehat{F}$, the estimated utility function $\widehat{U}$, and the deployer’s \emph{cost} and \emph{compliance} constraints to
produce (i) an \emph{ideal} target capability--cost profile $(x^*,c^*)$ and (ii) a \emph{concrete}
recommended model (or model portfolio) from the available set $\mathcal{M}$.
Moreover, when desired, the Optimization step may produce
a \emph{deployment-aware leaderboard} that ranks models under a specified deployment scenario.

\paragraph{Continuous target selection in $(x,c)$-space.}
Given a context $z$, we first solve for an \emph{ideal} target $(x^*,c^*)$ by optimizing
over the continuous capability--cost space:
\begin{equation} \label{eq:empirical_problem}
\begin{aligned}
    \max_{x,c} \quad & \widehat{U}(x;z) - \lambda B_0(c;w) \\ 
    \text{s.t.} \quad & B_j(c; b) \leq 0,\ \forall j \in [J], \quad
                               R_k(x; r) \leq 0,\ \forall k \in [K], \quad
                               \widehat{F}_X(x) \leq \widehat{F}_C(c).
\end{aligned}
\end{equation}
We make a couple of remarks. First, here, $\lambda$ governs how aggressively the deployer trades capability for cost. Concretely, if $\widehat{U}(x;z)$ is interpreted as expected value per interaction (e.g., expected success probability per prompt) and $B_0(c;w)$ as expected deployment cost per interaction (e.g., dollars per prompt), then $\lambda$ is the exchange rate between the two: increasing cost by $\Delta$ is worthwhile only if it raises expected utility by at least $\lambda\,\Delta$.
Second, to simplify exposition, we impose compliance constraints $R_k$ directly on the decision variables $x$, representing internal organizational requirements.\footnote{Regulatory compliance requirements are specified on raw descriptors $x_{\mathrm{raw}}$ rather than on the internal measures $x$. There are several ways to incorporate such constraints within MLC. (i) We may include the regulated raw descriptors directly into the internal measure vector $x$, bypassing the Measure Extraction step for such descriptors. (ii) We may enforce regulatory constraints as a pre-filter on the candidate model set and incorporate them in the continuous optimization through a conservative translation into $x$-space. Using the fitted measure-extraction model, we map a candidate internal profile $x$ to an implied raw profile $\widehat{x}_{\mathrm{raw}}(x)$, subtract a safety margin calibrated from prediction errors, and require this conservative estimate to satisfy the regulatory threshold. We leave this extension for future work.}

\paragraph{Single-policy vs.\ multi-policy deployment across contexts.}
If the organization can deploy different models across contexts (e.g., user or task types), it solves \eqref{eq:empirical_problem} separately for each $z$. If it must select \emph{one} policy across a set of contexts $\mathcal{Z}$, the aggregation strategy depends on the deployment setting.
In some cases, it may seek to maximize \emph{average} performance across a set of user or task types $\mathcal{Z}$ with weights $w_z \geq 0$:
\begin{equation}\label{eq:empirical_problem_average}
\begin{aligned}
    \max_{x,c} \quad & \sum_{z\in\mathcal{Z}} w_z\,\widehat{U}(x;z) \;-\; \lambda B_0(c;w) \quad
    \text{s.t.   (same constaints as in Problem \ref{eq:empirical_problem})}.
\end{aligned}
\end{equation}
In settings that prioritize worst-case guarantees, a \emph{robust} formulation is more appropriate, in which the objective maximizes the minimum utility across types:
\begin{equation}\label{eq:empirical_problem_robust}
\begin{aligned}
    \max_{x,c} \quad & \min_{z\in\mathcal{Z}} \big\{ \widehat{U}(x;z) \big\} \;-\; \lambda B_0(c;w)  \quad
    \text{s.t.   (same constaints as in Problem \ref{eq:empirical_problem})}.
\end{aligned}
\end{equation}

\paragraph{Discrete model recommendation.}
Because deployers ultimately choose among a finite set of available models, we map the target $(x^*,c^*)$ to a concrete recommendation. Define the discrete feasible set
\(
\mathcal{M}_{\mathrm{feas}}
:=
\Big\{
m\in\mathcal{M}:\ 
B_j(c(m);w)\le 0\ \forall j\in[J],\
R_k(x(m);r)\le 0\ \forall k\in[K]
\Big\}.
\)
\footnote{
We do not additionally enforce the empirical frontier at this stage, because real models may lie slightly above or below the estimated frontier due to sampling and specification error, and excluding them would be inconsistent with the observed set.}
If $\mathcal{M}_{\mathrm{feas}}=\emptyset$, then no currently available model satisfies the stated cost and compliance constraints; MLC flags infeasibility and the deployer must relax constraints or expand the candidate set.
If $\mathcal{M}_{\mathrm{feas}}\not=\emptyset$, then we recommend the model that maximizes the deployer's objective within this feasible set
\footnote{
Alternatively, one may select the model whose capability--cost profile is closest to the continuous target,
\(
m^*_{\mathrm{proj}} \in \arg\min_{m\in\mathcal{M}_{\mathrm{feas}}}
\Big\| (x(m),c(m))-(x^*,c^*)\Big\|_2.
\)
If $f(x,c):=\widehat{U}(x;z)-\lambda B_0(c;w)$ satisfies
$|f(p)-f(q)|\le L\|p-q\|_2$ for all feasible profiles $p,q$ (after normalizing coordinates), then the objective loss is bounded by
$f(x^*,c^*)-f(x(m^*_{\mathrm{proj}}),c(m^*_{\mathrm{proj}})) \le L\|(x(m^*_{\mathrm{proj}}),c(m^*_{\mathrm{proj}}))-(x^*,c^*)\|_2$.
This projection heuristic is justified when the estimated objective varies smoothly in the capability--cost profile and the available models densely cover the neighborhood of the continuous target: in such cases, the objective-maximizing discrete recommendation and the projection-to-target recommendation coincide. When they differ, we prioritize the objective-based discrete recommendation because it directly maximizes the deployer’s estimated payoff over feasible models.
}:
\begin{equation} \label{eq:projection}
m^* \in \arg\max_{m\in\mathcal{M}_{\mathrm{feas}}}\ \widehat{U}(x(m);z)-\lambda B_0(c(m);w).
\end{equation} 

\paragraph{Why continuous--then--projection.}
A purely discrete approach could enumerate all feasible $m\in\mathcal{M}_{\mathrm{feas}}$ and select the
best by objective value. We deliberately solve the continuous problem first because it produces an
interpretable \emph{target} capability--cost profile $(x^\star,c^\star)$, clarifies which constraints
bind and how sensitive the solution is to them, and provides a stable reference point for evaluating new
model releases by checking how close their $(x(m),c(m))$ lie to $(x^\star,c^\star)$.

\paragraph{Deployment-aware leaderboard.}
In addition to producing a single recommendation, MLC can generate a \emph{deployment-aware}
leaderboard tailored to a specific organization or sector. Fix a deployment scenario
$s := (\lambda,\{B_j\}_{j\in[J]},\{R_k\}_{k\in[K]},\{w_z\}_{z\in\mathcal{Z}})$,
i.e., a cost sensitivity, cost constraints on $c$, compliance constraints on $x$, and a distribution of contexts
(weights $w_z$). For each candidate model $m\in\mathcal{M}$, define the scenario score
\(
\mathrm{Score}_s(m)
=
\sum_{z\in\mathcal{Z}} w_z\,\widehat{U}(x(m);z)\;-\;\lambda\,B_0(c(m);w),
\)
and retain only models that satisfy the scenario constraints.
Ranking feasible models by $\mathrm{Score}_s(m)$ yields a \emph{deployment-aware leaderboard}.
Because the ranking is scenario-specific, MLC naturally produces \emph{multiple leaderboards} for
different deployment inputs (e.g., cost-focused vs.\ capability-focused $\lambda$, tighter vs.\ looser
compliance thresholds, average vs.\ worst-case aggregation), and these leaderboards can be refreshed whenever $\mathcal{M}$ or costs change. 
In Section~\ref{sec:casestudies}, we apply this leaderboard-inspired evaluation methodology in our two case studies---general-purpose conversational deployment (PRISM) and healthcare deployment (HealthBench).





\section{Empirical Validation: PRISM and HealthBench Case Studies}\label{sec:casestudies}
We now apply MLC end-to-end in two deployment settings that differ in both the source of outcome data and the practical meaning of deployment constraints. 
The first case study uses the PRISM Alignment dataset to represent a general-purpose conversational deployment in which deployment value is inferred from heterogeneous human preferences. 
The second uses HealthBench to represent a healthcare deployment in which outcomes are assessed using physician-designed rubrics and operational considerations (e.g., latency or scale) are central. 
Together, these case studies illustrate how the same pipeline can be instantiated with different inputs and how its recommendations can diverge from capability-only rankings once costs and compliance requirements are made explicit.

\subsection{Dataset Creation}\label{ssec:datasets}
Our empirical analysis draws on two complementary resources that correspond to distinct model-selection environments faced by deployers. To keep the main text focused on the MLC instantiation and results, we provide details on our dataset creation process in the EC (Sections~\ref{ec-ssec:prism} and \ref{ec-ssec:hb}), including summary statistics on the model set (e.g., correlations among internal measures) and analyses of user and task types across our case studies.

\paragraph{PRISM Alignment Dataset.}
The PRISM Alignment dataset is a public resource designed to capture diverse human preferences in LLM interactions.
It links sociodemographic characteristics and stated preferences of 1{,}500 participants from 75 countries to feedback on 8{,}011 multi-turn conversations with 21 LLMs, yielding over 68{,}000 human--AI interaction records. 
Conversations span value-laden topics where interpersonal and cross-cultural disagreement is expected, making PRISM well suited to general-purpose conversational deployments. 
For each interaction, the dataset records user-level preference profiles together with user-provided scores along multiple evaluation dimensions. \footnote{PRISM data accessed from the project’s public repository in May 2025.}

\paragraph{HealthBench.}
HealthBench is an open-source \emph{evaluation framework} developed by OpenAI to assess LLM performance and safety in realistic healthcare scenarios.
It provides a library of approximately 5{,}000 multi-turn healthcare conversations together with conversation-specific rubrics created and validated by physicians, which are used to score open-ended model responses across task families (e.g., emergent referrals, non-emergent referrals, global health). 
Because HealthBench specifies prompts and scoring rubrics (rather than distributing pre-computed outputs for a fixed set of models), we build a custom evaluation dataset by (i) selecting a set of candidate models (drawn from a public LLM leaderboard), (ii) generating model responses on the HealthBench conversations, and (iii) scoring those responses using the HealthBench rubrics. 
This yields interaction-level outcomes suitable for MLC, together with aligned model descriptors and operational proxies used to estimate the frontier and utility functions.
\footnote{HealthBench materials (e.g., rubrics) accessed from the benchmark repository in September 2025.}




\subsection{Inputs}\label{ssec:case-inputs}
We instantiate the inputs defined in Section~\ref{ssec:inputs} using the data available in PRISM and in our HealthBench-based evaluation dataset. 
For each environment we specify: (i) model-level \emph{Capability Data} $x_{\mathrm{raw}}(m)$, (ii) model-level \emph{Cost Data} $c_{\mathrm{raw}}(m)$ and its mapping to the scalar resource index $c(m)$ used in the optimization, (iii) interaction-level \emph{Outcome Data} $(i,m_i,z_i,y_i)$, and (iv) a set of cost and compliance requirements.\footnote{We defer the discussion of input (iv) to Section~\ref{ssec:case-optimization}, where, using different input sets of cost and compliance requirements, we define different evaluation scenarios.} 
Our input choices balance coverage of deployment-relevant capability dimensions and parsimony given the number of available models. Table~\ref{tab:inputs_case_studies} summarizes the design.\footnote{Model capabilities, technical metrics, and costs accessed from the LLM Leaderboard \citep{LLMLeaderboard} in October 2025. For PRISM, we used leaderboard data of February 2024 (close to the time the study took place); for HealthBench we used leaderboard data of October 2025.}

\begin{table}[t]
\centering
\caption{Inputs used to implement MLC for the PRISM and HealthBench case studies.}
\label{tab:inputs_case_studies}
\resizebox{1\textwidth}{!}{
\begin{tabular}{lll}
\toprule
Dataset & Input type & Variables used \\ 
\midrule
\multirow{3}{*}{PRISM} 
  & Capability Data $x_{\mathrm{raw}}(m)$
  & Aggregated model-level scores on creativity, diversity, factuality, fluency, helpfulness, safety, and values; \\
  & & generic benchmark scores (MMLU, MTEB). \\[4pt]

  & Outcome Data $(i,m_i,z_i,y_i)$ 
  & Interaction-level user scores (overall user satisfaction), with user/context descriptors in $z_i$. \\[4pt]

  & Cost Data $c_{\mathrm{raw}}(m)$ 
  & Blended per-token API price for each model $m$ (used as the main cost proxy in this case study). \\[6pt]
\midrule
\multirow{3}{*}{HealthBench} 
  & Capability Data $x_{\mathrm{raw}}(m)$
  & Throughput metrics (median and selected percentiles of tokens-per-second) and generic benchmarks \\
  & & (AAII, Terminal-Bench Hard, AA-LCR, Humanity Last Exam, IFBench, MMLU-Pro, LiveCodeBench, \\
  & & GPQA Diamond, SciCode, $\tau^2$-Bench Telecom, AIME 2025). \\[4pt]

  & Outcome Data $(i,m_i,z_i,y_i)$
  & Rubric-based HealthBench scores for clinical quality/safety, indexed by task descriptors in $z_i$. \\[4pt]

  & Cost Data $c_{\mathrm{raw}}(m)$ 
  & Parameter count (resource proxy) and first-answer token time (latency proxy) for each model $m$. \\
\bottomrule
\end{tabular}}
\end{table}

\paragraph{PRISM.}
For PRISM, the goal is to capture how conversational models trade off across user-facing quality dimensions and how those capabilities map to user-evaluated performance. 
\emph{(i) Capability Data} $x_{\mathrm{raw}}(m)$ consist of model-level aggregates of PRISM’s evaluation dimensions (creativity, diversity, factuality, fluency, helpfulness, safety, and values), which summarize how each model performs on average across many conversations. We augment these with a small set of generic benchmarks (MMLU and MTEB) to anchor $x_{\mathrm{raw}}$ in standard capability metrics without relying only on PRISM-specific scores. We do not include additional technical descriptors in PRISM because consistent, comparable throughput/latency measurements were not available across the full model set.\footnote{The PRISM case study was implemented in late 2023 using a set of models for which consistent technical metrics were not yet available or comparable across providers. We therefore restrict $x_{\mathrm{raw}}$ in this case to PRISM-derived aggregates and generic benchmarks, and defer richer technical descriptors to settings where reliable measurements exist for all models under consideration.}
\emph{(ii) Outcome Data} are interaction-level user evaluations: each interaction $i$ records the model $m_i$, user features $z_i$ from PRISM (used to build user-specific utility estimators), and user-provided satisfaction scores $y_i$. The exact construction of the outcome label used for utility estimation is described in Section~\ref{ssec:utility-estimation}. 
\emph{(iii) Cost Data} $c_{\mathrm{raw}}(m)$ are taken as blended per-token API prices. Because PRISM does not provide a consistent measure of model-specific token usage under a fixed workload, we treat token price as the primary driver of relative deployment cost in this setting (i.e., differences in expected tokens per interaction are not separately modeled here).

\paragraph{HealthBench.}
For HealthBench, the goal is to study how general-purpose capabilities and operational characteristics translate into clinical performance in a domain-specific setting.
\emph{(i) Capability Data} $x_{\mathrm{raw}}(m)$ combine (a) generic capability benchmarks that cover reasoning, scientific knowledge, and coding (AAII, Terminal-Bench Hard, AA-LCR, Humanity Last Exam, IFBench, MMLU-Pro, LiveCodeBench, GPQA Diamond, SciCode, $\tau^2$-Bench Telecom, AIME 2025) and (b) technical throughput metrics (median and selected percentiles of tokens-per-second) extracted from the LLM leaderboard. This design intentionally avoids healthcare-specific capability signals on the input side, so that clinical performance is captured through outcomes rather than “baked into” $x_{\mathrm{raw}}(m)$.
\emph{(ii) Outcome Data} are built by running each candidate model on HealthBench conversations and scoring responses using the physician-validated rubrics. Each interaction $i$ corresponds to a (conversation, model) pair, with task descriptors in $z_i$ (e.g., task family such as emergent vs.\ non-emergent referral) and rubric-based outcomes $y_i$ (used to construct the task-conditional utility estimators described in Section~\ref{ssec:utility-estimation}). For 
\emph{(iii) Cost Data}, comparable all-in API pricing is often unavailable for the evaluated healthcare workflows, so we use operational proxies. In the core analysis we use parameter count as the scalar resource proxy $c(m)$ (capturing scale-related deployment burden), and we use first-answer token time as a latency proxy that can be treated either as an additional resource requirement or in robustness variants. Together, these capture the main operational costs an organization would face when deploying these models in a clinical setting.

\subsection{Measure Extraction}
\begin{figure}[!htb]
    \begin{minipage}{0.47\textwidth}
     \centering
     \includegraphics[width=\linewidth]{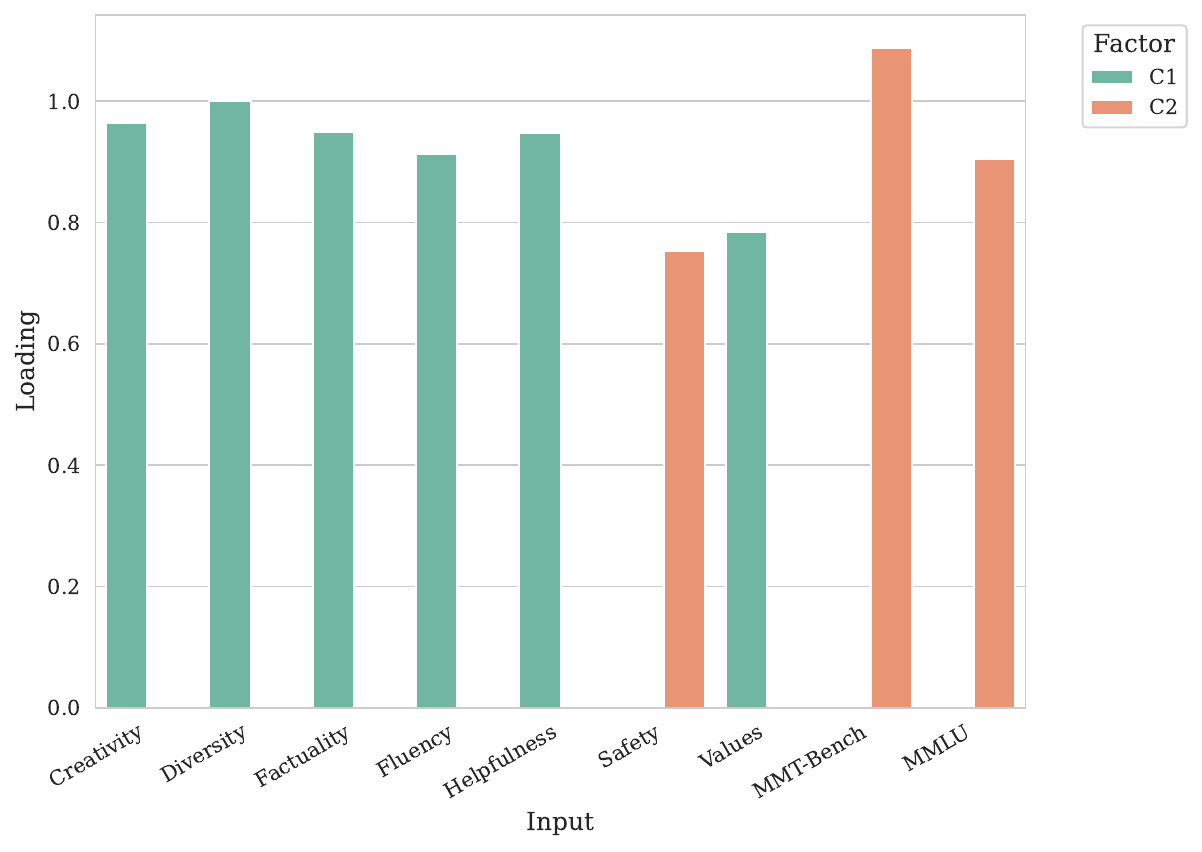}
   \end{minipage}
   \begin{minipage}{0.51\textwidth}
     \centering
     \includegraphics[width=\linewidth]{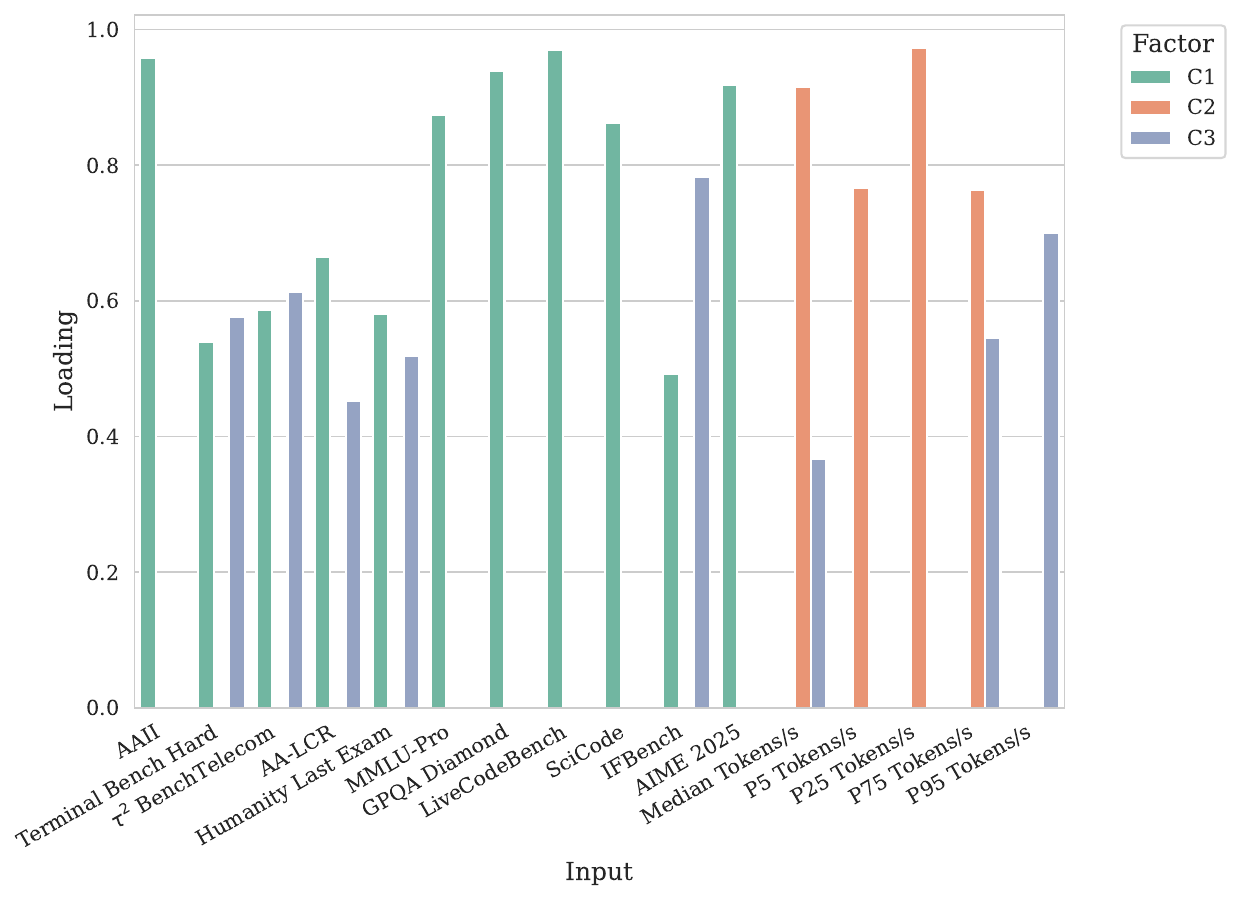}
   \end{minipage}\hfill
   \caption{Measure extraction: factor loadings for capability data.}
   \smallskip \footnotesize \parbox{0.95\linewidth}{\textit{Note.}
   This figure instantiates the Measure extraction step ($x_{\mathrm{raw}}\rightarrow x$) described in Section~\ref{ssec:measure-extractor}. 
   Each panel shows the factor-loading matrix estimated on the model-level Capability Data $x_{\mathrm{raw}}(m)$ (Table~\ref{tab:inputs_case_studies}), after rotation and sparsification to improve interpretability. 
   The PRISM panel retains two factors ($C_1,C_2$) and the HealthBench panel retains three factors ($C_1,C_2,C_3$). 
   Larger loadings indicate that the corresponding raw capability metric contributes more to the associated latent factor. 
   We interpret the resulting factors as internal measures $x(m)$ used throughout MLC: conversational experience and risk-aware competence for PRISM, and knowledge/reasoning, efficiency, and human-aligned agentic capability for HealthBench. }
   \label{Fig:FAloadings}
\end{figure}

Following Section~\ref{ssec:measure-extractor}, we apply the Measure Extraction step to each of the two case studies. 
Using the model-level Capability Data $x_{\mathrm{raw}}(m)$ listed in Table~\ref{tab:inputs_case_studies}, we apply factor analysis (with rotation and sparsification for interpretability) to construct a low-dimensional vector of internal measures $x(m)$. 
We retain $I=2$ factors for PRISM and $I=3$ factors for HealthBench. 
Figure~\ref{Fig:FAloadings} shows the factor-loadings estimated in each case study.
Under our specification, the retained factors explain 86.1\% of the common variance in PRISM and 75.0\% in HealthBench.
Factor scores are min--max scaled to $[0,1]$ across models before frontier and utility estimation. 

\paragraph{PRISM.}
Figure~\ref{Fig:FAloadings} (left) shows that the first PRISM factor loads primarily on user-facing experience dimensions (creativity, diversity, fluency, helpfulness, and values), while the second factor is dominated by safety together with the generic capability benchmarks. We interpret these internal measures as: (i) \emph{conversational experience} (Factor~1) and (ii) \emph{risk-aware competence} (Factor~2). This separation is useful in downstream steps because it distinguishes “how users experience the interaction” from “capability and safety as captured by auditable benchmarks.”

\paragraph{HealthBench.}
For HealthBench, the loading structure (Figure~\ref{Fig:FAloadings}, right) separates three distinct sources of variation. The first factor is driven mainly by generic capability benchmarks (reasoning, scientific knowledge, coding). The second factor loads primarily on throughput variables (tokens-per-second statistics), capturing operational \emph{efficiency}. The third factor is associated with more agentic or interaction-intensive benchmarks (e.g., IFBench, Humanity Last Exam, Terminal-Bench Hard, $\tau^2$-Bench Telecom). We interpret these as: (i) \emph{knowledge and reasoning} (Factor~1), (ii) \emph{efficiency} (Factor~2), and (iii) \emph{human-aligned/agentic capability} (Factor~3).

\paragraph{Robustness.} 
In the EC (Section~\ref{ec:measure-extractor}), we assess the robustness of the Measure Extraction step to standard factor-analysis design choices---specifically the number of factors, rotation, and sparsity-inducing post-processing. Across both PRISM and HealthBench, the leading factors remain interpretable and stable. Adding rotation and/or sparsification improves interpretability but can reduce variance explained, reflecting (i) the deliberate information loss induced by sparsification (zeroing small loadings) and (ii) numerical sensitivity of the rotation implementation. Accordingly, we prioritize low-dimensional representations that are interpretable, preserve most variance, and yield stable factor structure. 


\subsection{Frontier Estimation}\label{ssec:case-frontier}
\begin{figure}[!htb]
    \begin{minipage}{0.5\textwidth}
     \centering
     \includegraphics[width=\linewidth]{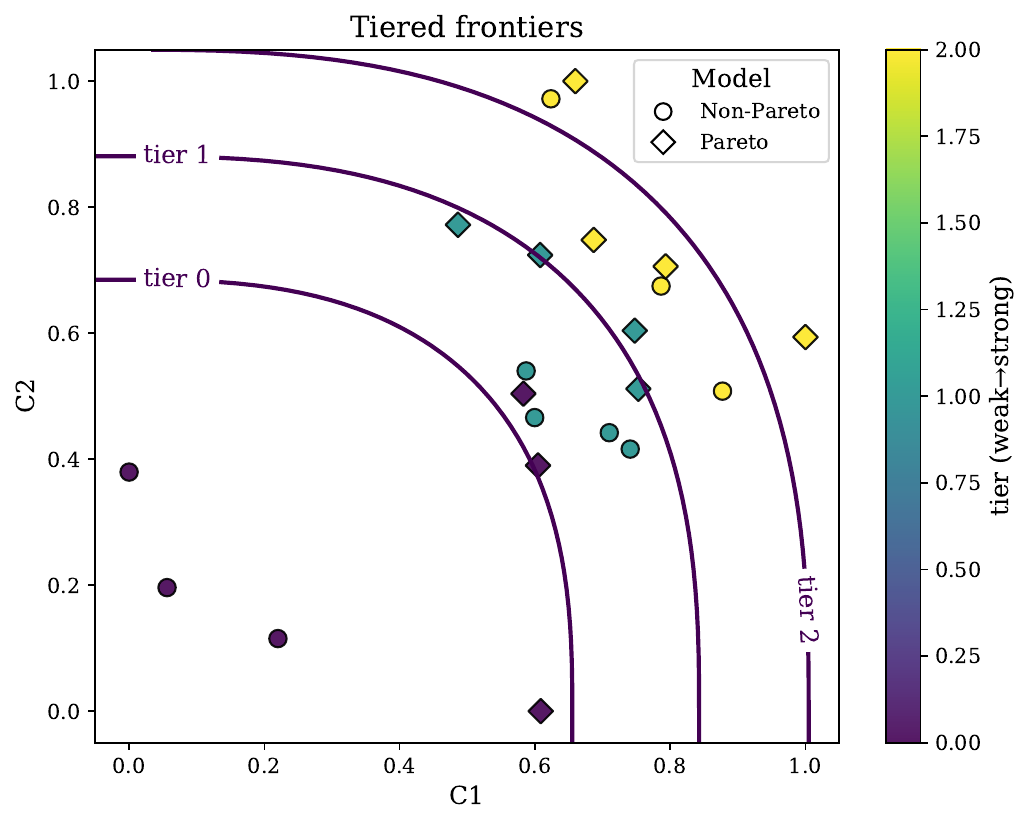}
   \end{minipage}
   \begin{minipage}{0.5\textwidth}
     \centering
     \includegraphics[width=\linewidth]{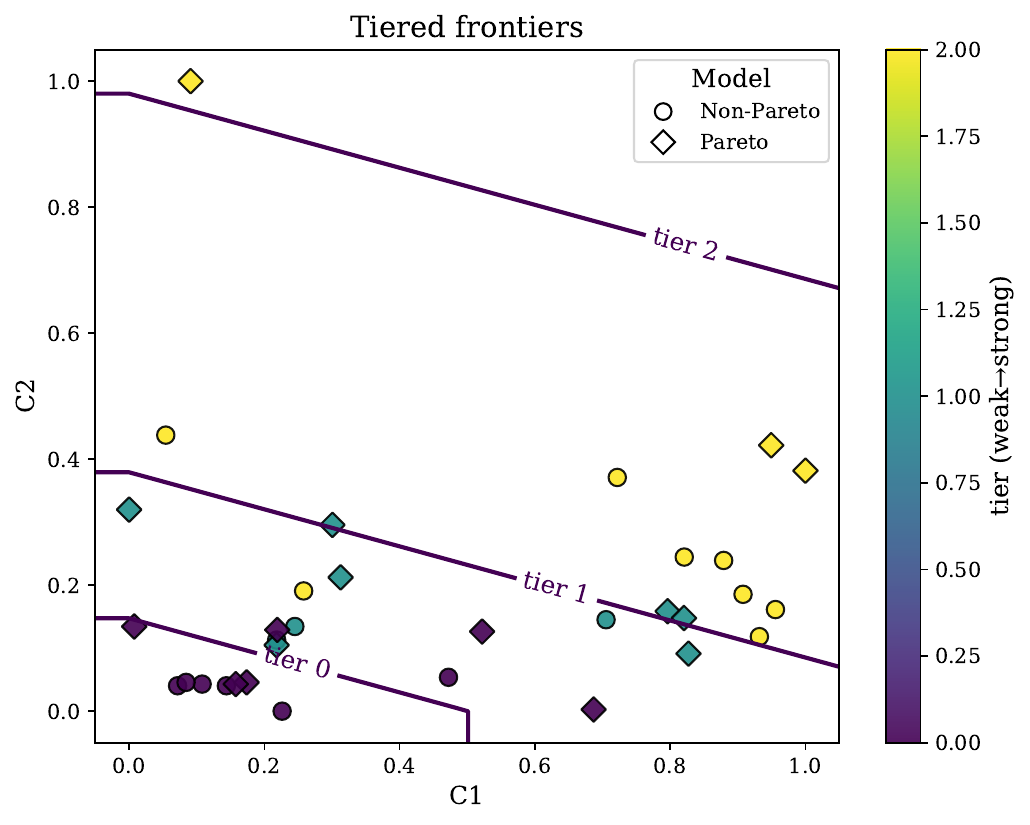}
   \end{minipage}\hfill
   \caption{Frontier estimation.}
   \smallskip \footnotesize \parbox{0.95\linewidth}{\textit{Note.}
   Each panel shows the fitted frontier tiers. 
   The PRISM panel is plotted in the full extracted internal measure space $(C_1,C_2)$. 
   The HealthBench panel shows a \emph{projection} onto $(C_1,C_2)$ for readability (the HealthBench frontier is estimated in $(C_1,C_2,C_3)$). 
   In both cases, we estimate 3 tiers: tiers are ordered from weakest to strongest, and color-coded accordingly (purple$\rightarrow$blue$\rightarrow$yellow).
   The displayed curves summarize the implied trade-off surface between internal measures and deployment resources. 
   Within each tier, squares correspond to Pareto-dominating models and circles correspond to Pareto-dominated models.  
   }
   \label{fig:frontier-visualization}
\end{figure}

Following Section~\ref{ssec:frontier-estimation}, we apply the Frontier Estimation step to the internal measures extracted above and the corresponding Cost Data (Table~\ref{tab:inputs_case_studies}). 
For each dataset, we (i) tier models using approximate Pareto peeling in capability space, (ii) treat Pareto-efficient models within each tier as empirical frontier points, and (iii) fit a tiered CES frontier with a power-law cost mapping. 
We use $T=3$ tiers for interpretability. Figure~\ref{fig:frontier-visualization} visualizes the fitted tiered frontier for PRISM in $(C_1,C_2)$ and the 2D projection onto $(C_1,C_2)$ for HealthBench (the HealthBench frontier is estimated in three dimensions, using $(C_1,C_2,C_3)$; additional projections appear in the EC, Section~\ref{ec:frontier}). 

\begin{table}[h]
\centering
\caption{Tier assignment and Pareto status of evaluated models.}
\smallskip \footnotesize \parbox{0.95\linewidth}{\textit{Note.}
Models are first assigned to tiers via approximate Pareto peeling in capability space (weakest to strongest). 
Pareto peeling produces only Pareto-dominating models at each peeled layer, but may yield more tiers than desired. 
To obtain a fixed number of tiers, we merge adjacent peeled layers; after merging, some models become Pareto-dominated within their merged tier. 
Within each final tier, Pareto-dominating models lie on the tier’s empirical Pareto frontier in internal-measure space, while Pareto-dominated models are strictly worse in at least one internal measure without being better in any other. 
This table reports the resulting tier assignments and identifies the models that anchor each tier in Figure~\ref{fig:frontier-visualization}.
}
\smallskip 
\label{tab:frontiermodels}
\resizebox{1\textwidth}{!}{
\begin{tabular}{lcll}
\toprule
Dataset & Tier & Pareto Dominated models & Pareto Dominating Models \\ 
\midrule
\multirow{3}{*}{PRISM} 
  & 0 & \texttt{flan-t5-xxl}, \texttt{luminous-extended-control}, \texttt{luminous-supreme-control} & \texttt{llama-2-7b-chat} and \texttt{oasst-pythia-12b} \\[4pt]
\cline{2-4}
  & 1 & & \texttt{mistral-7b-instruct-v0.1}, \texttt{llama-2-13b-chat}, \texttt{guanaco-33b}, \\ 
  & & & \texttt{falcon-7b-instruct} and \texttt{chat-bison} \\[4pt]
\cline{2-4}
  & 2  & \texttt{zephyr-7b-beta}, \texttt{cohere-command-light}, \texttt{gpt-3.5-turbo}, & \texttt{gpt-4}, \texttt{claude-2}, \texttt{cohere-command-nightly}, \\
  & & \texttt{llama-2-70b-chat}, \texttt{claude-2.1} and \texttt{gpt-4-1106-preview} & \texttt{claude-instant-1} and \texttt{cohere-command} \\
\midrule
\multirow{3}{*}{HealthBench} 
  & 0 & \texttt{Gemma 3 4B}, \texttt{Gemma 3n E2B}, \texttt{Gemma 3n E4B}, & \texttt{Hermes 4 70B (non reasoning)}, \texttt{DeepSeek R1 0528 Qwen3 8B}, \texttt{Gemma 3 12B},\\
  & & \texttt{Phi-4}, \texttt{Reka Flash 3} and \texttt{Aya Expanse 32B} & \texttt{Gemma 3 27B}, \texttt{EXAONE 4.0 32B (non reasoning)} and \texttt{Aya Expanse 8B} \\[4pt]
\cline{2-4}
  & 1 & \texttt{Command A}, \texttt{Qwen3 30B A3B 2507 (non reasoning)}, & \texttt{Mistral Small 3.2}, \texttt{Devstral Small}, \texttt{Qwen3 30B A3B 2507 (reasoning)}, \texttt{Jamba 1.7 Mini},\\ 
  & & and \texttt{Llama Nemotron Super 49B v1.5 (non reasoning)} & \texttt{Llama 3.3 70B}, \texttt{Hermes 4 70B (reasoning)}, and \texttt{Qwen3 Next 80B A3B (non reasoning)} \\[4pt]
\cline{2-4}
  & 2 & \texttt{EXAONE 4.0 32B (reasoning)}, \texttt{GLM-4.5-Air}, \texttt{Llama 4 Scout} & \texttt{Qwen3 Next 80B A3B (reasoning)} \\
  & & \texttt{gpt-oss-20B}, \texttt{NVIDIA Nemotron Nano 9B V2 (non reasoning)} & and \texttt{gpt-oss-120B} and \texttt{Granite 3.3 8B}\\ 
  & & \texttt{Ministral 8B}, \texttt{NVIDIA Nemotron Nano 9B V2 (reasoning)} & \\
  & & and \texttt{Llama Nemotron Super 49B v1.5 (reasoning)} & \\
\bottomrule
\end{tabular}}
\end{table}

\paragraph{PRISM.}
For PRISM, the estimated CES weights are $\hat{a}\approx(0.53,\,0.47)$, so the frontier loads slightly more on conversational experience ($C_1$) than on risk-aware competence ($C_2$). 
The curvature parameter is $\hat{b}\approx2.67$, implying a visibly curved frontier in $(C_1,C_2)$: improvements can be achieved by trading off across dimensions rather than moving only “diagonally.” 
The tier levels are $\hat{\lambda}\approx(0.52,\,0.66,\,0.79)$, indicating relatively compressed capability gaps across tiers. 
On the cost side, $\widehat{F}_C(c)=\hat{c}_0 c^{\hat{d}}$ with $\hat{c}_0\approx0.49$ and $\hat{d}\approx0.21$, so frontier capability increases slowly with deployment spend in this environment.
Table~\ref{tab:frontiermodels} reports tier membership and (within-tier) Pareto status. 
In PRISM, the strongest-tier Pareto-efficient set contains frontier models such as \texttt{gpt-4}, \texttt{claude-2}, and \texttt{cohere-command(-nightly)}, while weaker tiers contain smaller open-weight chat models that trade off capability dimensions differently. 
Comparing model-level blended API prices to the implied tier-level costs yields Spearman and Pearson correlations of 0.37 and 0.34, respectively, suggesting that PRISM-era prices are only weakly aligned with frontier capability.\footnote{This is consistent with (i) substantial within-tier heterogeneity in a small model set and (ii) pricing that reflects market strategy and ecosystem positioning rather than capability alone. PRISM pricing relies on the 2023 market, which was comparatively ad hoc.}

\paragraph{HealthBench.}
For HealthBench, the CES weights are $\hat{a}\approx(0.17,\,0.60,\,0.26)$, so the estimated frontier loads most on efficiency ($C_2$) relative to knowledge/reasoning ($C_1$) and human-aligned/agentic capability ($C_3$). 
The curvature parameter $\hat{b}\approx1.00$ is close to linear aggregation, which produces near-linear frontier level sets (and sharper “rate-of-substitution” trade-offs) in the projected $(C_1,C_2)$ view. 
Tier levels are $\hat{\lambda}\approx(0.12,\,0.26,\,0.60)$, showing much wider tier separation than PRISM (roughly a five-fold difference between the strongest and weakest tier). 
The cost mapping $\widehat{F}_C(c)=\hat{c}_0 c^{\hat{d}}$ with $\hat{c}_0\approx1.02$ and $\hat{d}\approx0.43$ is steeper than in PRISM, consistent with capability increasing more directly with resource proxies in this setting.
As a concrete illustration of the capability--efficiency trade-off, the highest-tier Pareto-efficient set includes models such as \texttt{gpt-oss-120B} (strong benchmark capability), \texttt{Granite 3.3 8B} (high efficiency), and \texttt{Qwen3 Next 80B A3B (reasoning)} (strong on both in our internal-measure space). 
In this case, the Spearman and Pearson correlations between tier-implied costs and model-level costs/prices are 0.33 and 0.44, respectively: the linear association is stronger than PRISM, but rank alignment remains imperfect, reflecting meaningful outliers and cross-model pricing heterogeneity. 
We attribute this behavior to the larger model sample size, which leads to a more homogeneous Pareto tiering, and to the nature of the benchmark itself. In particular, reasoning-oriented benchmarks—which we intuitively expect to reflect clinical knowledge—are more closely associated with model scale and technical specifications, such as parameter count. Nevertheless, notable outliers remain, as evidenced by the first tier containing models ranging from 8B to 120B parameters, which weakens the consistency of rank-based correlations.

\paragraph{Robustness.}
We evaluate robustness by varying: (i) the Capability/Cost input set, (ii) the number of input internal measures, (iii) the number of tiers, (iv) the Pareto tolerance used for dominance comparisons, and (v) the within-tier cost aggregation method (mean/geometric mean/median). 
Across these variations, the qualitative patterns above remain stable: PRISM continues to exhibit a more curved, compressed-tier frontier, while HealthBench continues to exhibit a nearer-linear frontier with wider tier separation. 
Quantitatively, increasing the number of factors or tiers tends to increase the estimated substitutability parameter $b$ and reduce the returns parameter $d$ (consistent with more available substitution directions and finer grouping of similar models), while fitting error and cost-alignment metrics (Pearson/Spearman) remain comparatively stable across most configurations. 
Finally, the frontier-parameter assumptions used for the stylized analysis are broadly supported in PRISM and for the main HealthBench specifications, but configurations that omit technical variables in HealthBench can yield lower estimated $b$ values, consistent with reduced substitutability when efficiency-related dimensions are excluded. 
Section~\ref{ec:frontier} reports the full sensitivity tables and parameter histograms.




\subsection{Utility Estimation}
Following Section~\ref{ssec:utility-estimation}, we apply the Utility Estimation step. We estimate context-dependent utility models $\widehat{U}(x;z)$ that map a model’s internal measures $x(m)$ to the probability of a successful interaction outcome. Context may correspond to user characteristics, task characteristics, or both; to illustrate this flexibility, we focus on user-based contexts in PRISM (where rich user features are available) and on task-based contexts in HealthBench (where interactions are organized into diverse clinical categories but user-level information is not available).
Consistent with the recipe in Section~\ref{ssec:utility-estimation}, we map contexts to a finite set of groups and estimate one classifier per group predicting success vs. failure.
For diversity in model classes, we use a linear function class (logistic regression) with linear utility (by plugging the estimated regression coefficients into the optimization objectioni) for PRISM, and a nonlinear function class (LightGBM) with nonlinear utility (by using the estimated model as a black-box function in the optimization objective) for HealthBench.
Table~\ref{tab:utility_auc-SHORT} summarizes the predictive performance of the estimated utility models across contexts. Importantly, our goal is not predictive modeling per se: we estimate an empirical proxy for the deployer’s utility function and assess whether it is stable and informative enough to support model selection. 

\begin{table}[t]
\centering
\caption{Performance of estimated utility models.}
\smallskip \footnotesize \parbox{0.95\linewidth}{\textit{Note.}
For each context group $z$, we fit a classifier that predicts a binary success outcome $y_i$ from the internal measures $x(m_i)$ of the model used in interaction $i$.
We report training/test sample sizes and AUC; values in parentheses are standard deviations across repeated random train/test splits.
The last columns summarize how the fitted utility depends on the internal measures: for PRISM, $(\hat\beta_1,\hat\beta_2)$ are the normalized absolute logistic-regression coefficients on $(C_1,C_2)$; for HealthBench, $(\hat\beta_1,\hat\beta_2,\hat\beta_3)$ are normalized LightGBM feature-importance gains on $(C_1,C_2,C_3)$.
For PRISM, $C_1$ corresponds to conversational experience and $C_2$ to risk-aware competence; for HealthBench, $C_1$ corresponds to knowledge/reasoning, $C_2$ to efficiency, and $C_3$ to human-aligned/agentic capability.}
\smallskip 
\label{tab:utility_auc-SHORT}
\resizebox{0.88\textwidth}{!}{
\begin{tabular}{llccccccc}
\toprule
Dataset & Context group $z$ & $|\mathcal{I}_{\text{train}}|$ & $|\mathcal{I}_{\text{test}}|$ & Train AUC & Test AUC & $\hat\beta_1$ & $\hat\beta_2$ & $\hat\beta_3$\\
\midrule
\multirow{3}{*}{PRISM} 
  & Ethics-focused users & 11,688 & 2,922 & 0.668 (0.003) & 0.668 (0.014) & 1.000 & 0.000 & \textemdash\\
  & Safety-focused users & 1,764  & 441   & 0.687 (0.006) & 0.687 (0.027) & 0.017 & 0.983 & \textemdash\\
  & General users        & 9,889  & 2,472 & 0.649 (0.003) & 0.649 (0.013) & 0.579 & 0.421 & \textemdash\\
  & MEAN        & 7,780&	1,945&	0.668 (0.016)	&	0.668 (0.016)	& \textemdash & \textemdash & \textemdash\\
\midrule
\multirow{3}{*}{HealthBench} 
  & Health professional        & 8,500  & 2,125 & 0.805 (0.006) & 0.801 (0.022) & 0.848 & 0.091 & 0.061\\
  & Non-health professional & 11,193 & 2,798 & 0.687 (0.002) & 0.682 (0.010) & 0.295 & 0.380 & 0.325\\
  & Simple answer                              & 4,303  & 1,075 & 0.648 (0.003) & 0.640 (0.014) & 0.155 & 0.677 & 0.168\\
  & MEAN &	5,700&	1,424&	0.720 (0.049)&	0.711 (0.050)& \textemdash & \textemdash & \textemdash\\
\bottomrule
\end{tabular}}
\end{table}

\paragraph{PRISM.} We construct user types as preference orientations over the internal measures: $C_1$-oriented (``Ethics-focused''), $C_2$-oriented (``Safety-focused''), or balanced (``General''); Section \ref{ec:datasets} of the EC reports the user type creation process and descriptive profiles based on observed covariates (e.g., demographics and stated preferences).
For each interaction $i$ from user $u$, we define a within-user success indicator by normalizing scores relative to the user’s own scale: letting $\bar{s}_u$ denote user $u$'s mean score across their interactions, we set $y_i=1$ if the score in interaction $i$ exceeds $\bar{s}_u$, and $y_i=0$ otherwise.
We estimate an type-specific logistic utility model $\widehat{U}(x;z)$ that predicts within-user success $y_i$ from the internal measures $x(m_i)$.
Table~\ref{tab:utility_auc-SHORT} shows that these models achieve test AUCs around 0.65--0.69, indicating meaningful out-of-sample signal.
The estimated weights clarify which capabilities matter for different user types.
For ethics-focused users, predicted success loads almost entirely on conversational experience ($C_1$), while for safety-focused users it loads almost entirely on risk-aware competence ($C_2$). 
General users place substantial weight on both dimensions.

\paragraph{HealthBench.}
For HealthBench, we define context groups $z$ using the benchmark's metadata, corresponding to different task types and resulting in a total of 17 groups. To simplify presentation, we report three representative task types that we use consistently in later figures: \emph{health professional (expert-level)}, \emph{non-health professional (everyday-language)}, and \emph{simple answer} (the full results are given in Section~\ref{ec:utility} of the EC). 
We define $y_i=1$ if the response achieves full marks on the HealthBench rubric for that interaction and $y_i=0$ otherwise, and we fit a LightGBM classifier predicting $y_i$ from $x(m_i)$.
Table~\ref{tab:utility_auc-SHORT} shows that, across all 17 task types, these models achieve a mean test AUC of 0.71.
The three task types exhibit distinct capability--value mappings.
For health professional (expert-level) prompts, utility is driven primarily by knowledge/reasoning ($C_1$). 
For non-health professional (everyday-language) prompts, success depends more evenly on all three internal measures, consistent with a larger role for interaction style and human-aligned behavior in addition to raw knowledge. 
For simple-answer prompts, efficiency ($C_2$) receives the largest weight, reflecting the value of producing correct, concise outputs.

\paragraph{Robustness.}
Section~\ref{ec:utility} of the EC reports additional utility-estimation specifications. We vary (i) how context $z$ is used (grouping only versus also including context variables directly as features), (ii) the grouping context (users or tasks or both), and (iii) the estimator class (logistic regression versus boosted trees). These checks help understand how sensitive the inferred context dependence of utility is to modeling choices. In the remainder of the case studies, we use the resulting $\widehat{U}(x;z)$ as an empirical proxy for context-specific deployment value when computing recommendations and scenario scores.



\subsection{Optimization and Outputs} \label{ssec:case-optimization}
In the final step of the pipeline, for any fixed deployment scenario (cost sensitivity and any cost/compliance requirements) and context (user type in PRISM and task type in HealthBench), MLC selects a model by solving for a target capability--cost profile, mapping it to a discrete model, and finally estimating its predicted deployment value net of a resource penalty.
Moreover, the same deployment value used for recommendation assigns each candidate model a scenario score; ranking models by this score yields a deployment-aware leaderboard tied to deployment requirements rather than to any benchmark.
Thus, in the final step of our empirical validation, we evaluate MLC's target capability-cost profiles and recommended models in terms of their deployment value across different scenarios in each of the two case studies (as per the methodology of Section~\ref{ssec:optimization}). 

\paragraph{Deployment scenarios and compared policies.}
We evaluate five deployment settings that mirror common organizational decision regimes:
(i) \emph{Pure capability} (no resource penalty),
(ii) \emph{Cost-aware (low)} (mild resource penalty),
(iii) \emph{Cost-aware (high)} (strong resource penalty),
(iv) \emph{Constrained (single)} (mild resource penalty plus a binding compliance screen on one key capability), and
(v) \emph{Constrained (all)} (mild resource penalty plus multi-dimensional compliance screens).
\footnote{We set the cost-sensitivity levels to reflect the scale of costs in each dataset. For PRISM we use $\lambda\in\{0.05,\,0.5\}$ for the low and high cost-aware regimes, respectively. For HealthBench, costs are roughly an order of magnitude smaller while internal measures are on a comparable scale, so we use larger sensitivities $\lambda\in\{0.5,\,5\}$ to obtain comparable cost pressure. In the constrained regimes we use the low-$\lambda$ value and impose additional compliance screens: \emph{Constrained (all)} requires each internal measure to be at least $0.33$ of its observed range, while \emph{Constrained (single)} requires the second internal measure (C2) to be at least $0.5$ of its range.}

\paragraph{MLC variants.}
Within each setting, we implement three variants of the MLC framework:
\emph{(i) MLC (per-type)} selects a potentially different model for each context $z$, maximizing the context-specific objective subject to feasibility (as per Problem \ref{eq:empirical_problem}).
\emph{(ii) MLC (mean)} selects a single model to deploy across all contexts, maximizing the objective averaged (uniformly) over contexts (as per Problem \ref{eq:empirical_problem_average}).
\emph{(iii) MLC (robust)} selects a single model to deploy across all contexts, maximizing the worst-case objective across contexts (as per Problem \ref{eq:empirical_problem_robust}).

\paragraph{Comparison benchmarks.}
We compare the above approaches against two types of benchmarks:
\textit{(i)} As baseline, we report \emph{single-metric heuristics} that choose the best feasible model according to one raw capability measure (or a cost proxy) at a time, ignoring preference heterogeneity and the frontier; we aggregate performance across the full set of such heuristics by reporting their mean and standard deviation.
\textit{(ii)} As external check, we report an \emph{observed winner} (``Actual''): in PRISM, the most commonly selected feasible model within each user type; in HealthBench, the feasible model with the highest average rubric score within each task type.
The observed winners are not designed to optimize our cost-aware deployment objective; we use them as a sanity check on whether MLC recommendations remain close to the chosen models under the data-generating process.

\paragraph{Evaluation metric.} 
All approaches are evaluated using the same estimated deployment value built from $\widehat{U}$ and the observed cost inputs. 
To stay as close as possible to deployment value given the available data, we score each policy as follows: (i) we plug the selected model into $\widehat{U}(\cdot;z)$ for each context $z$ (thus using the most granular context-dependent utility estimates), (ii) we average across contexts, and (iii) we subtract any cost penalty using the selected model's actual price (instead of the tier's price that we use in the optimization) to obtain the final deployment value.

We now explain how to interpret this deployment value.  Internal measures are min--max scaled to $[0,1]$. In PRISM, the utility term is a linear score $\widehat{U}(x;z)=\hat\beta_z^\top x$ with nonnegative coefficients normalized to sum to one, hence $\widehat{U}(x;z)\in[0,1]$; under \emph{Pure capability} ($\lambda=0$) the deployment value equals this score, while under cost-aware regimes we report $\hat\beta_z^\top x(m)-\lambda c(m)$ using the (unnormalized) blended API price $c(m)$ (max $c(m)=37.5$ in our PRISM candidate set, which is the blended price of \texttt{gpt-4}). Therefore, the deployment value lies in $[-\lambda\cdot 37.5,\;1]$.
In HealthBench, $\widehat{U}(x;z)\in[0,1]$ is a predicted success probability; under cost-aware regimes we penalize it with a normalized cost proxy in $[0,1]$, so the objective lies in $[-\lambda,\;1]$.

\begin{figure}[!htb]
\centering
\begin{minipage}{0.49\textwidth}
  \centering
  \includegraphics[width=\linewidth]{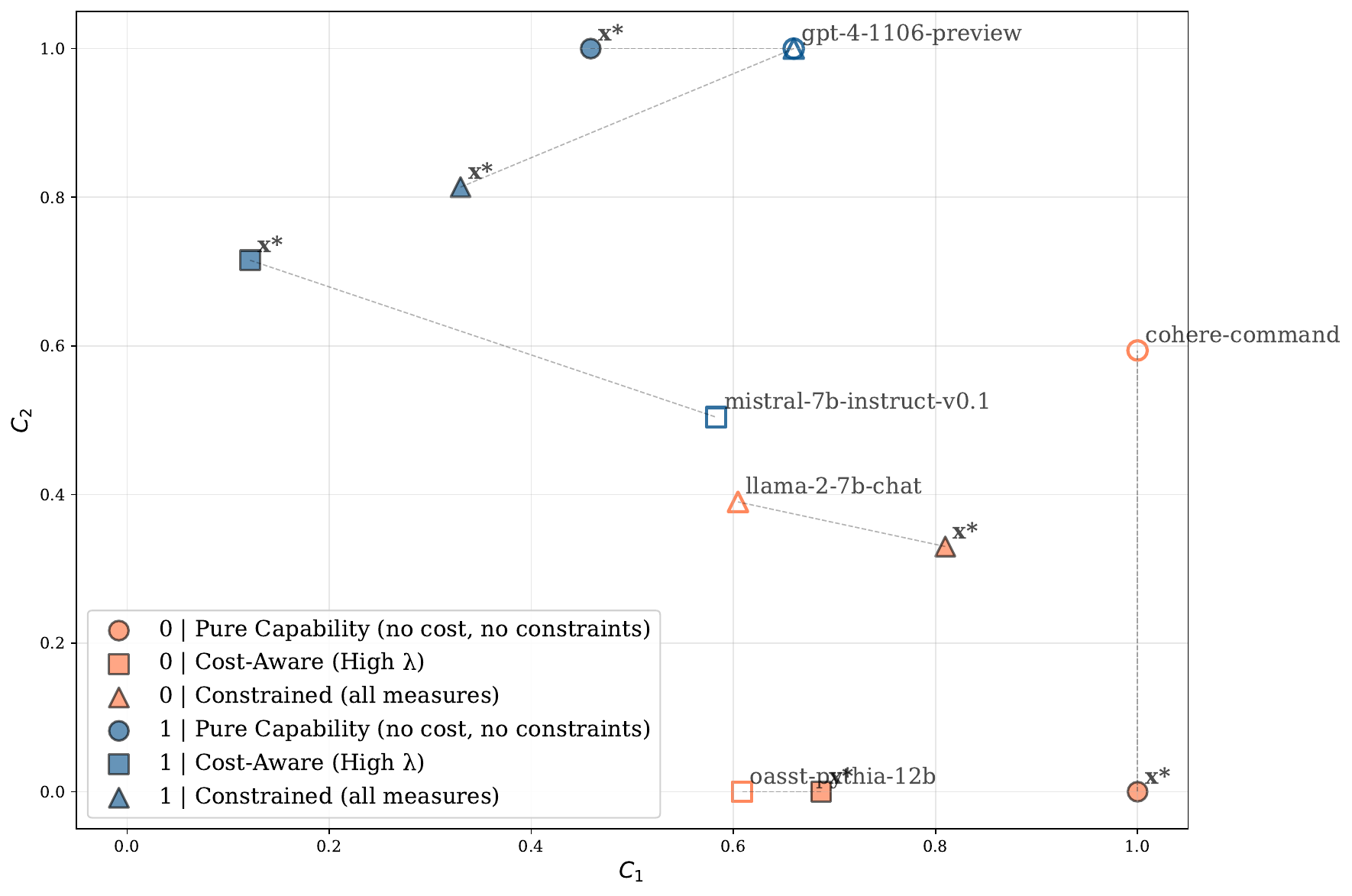}
\end{minipage}\hfill
\begin{minipage}{0.49\textwidth}
  \centering
  \includegraphics[width=\linewidth]{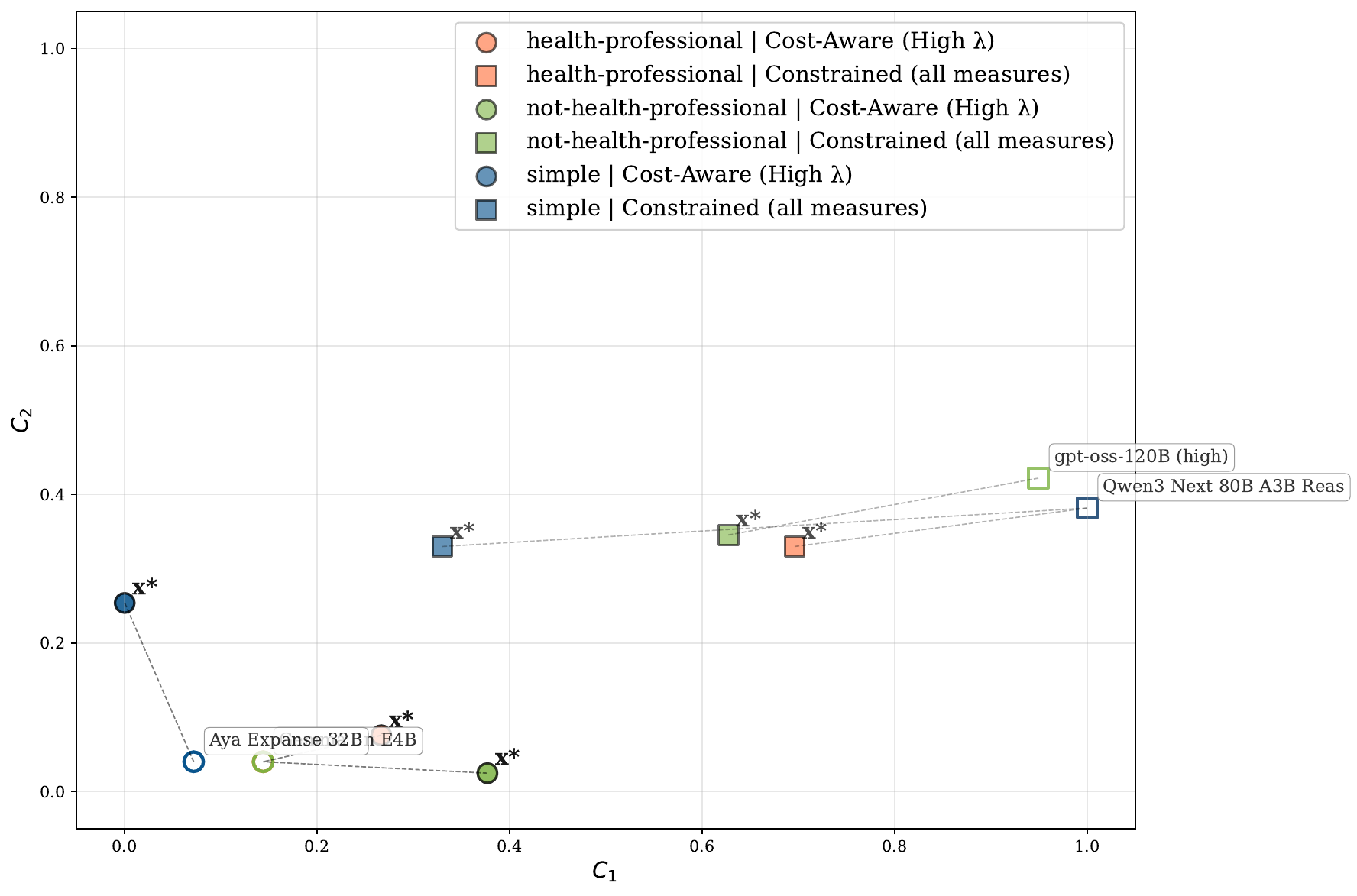}
\end{minipage}
\caption{Optimal targets and recommended models in internal-measure space.}
\smallskip \footnotesize \parbox{0.95\linewidth}{\textit{Note.}
Panels show the continuous optimum (target) produced by the optimization step for each context and deployment setting, plotted in internal-measure coordinates.
For PRISM, we show the first two user types (Ethics-Focused is shown in orange, Safety-Focused is shown in blue) across three selected deployment scenarios. 
For HealthBench, we show all three selected task types (Health Professional, Non-Health Professional, Simple) across two deployment scenarios.
Dashed guides link each continuous target to the selected discrete model from the candidate set.
Apparent ``large'' projections can occur because cost enters the optimization but is not shown in this two-dimensional plot, because we use observed per-model cost inputs (pricing/proxies), and because the candidate model set is discrete and unevenly distributed in capability--cost space.
In case of ties in predicted utility, the lowest-cost feasible model is selected.
}
\label{fig:case-solutions}
\end{figure}

\paragraph{Results.}
Results are summarized in Table~\ref{tab:summary_objective} (deployment values) and Table~\ref{tab:model_selection} (selected models), with the corresponding target capability profiles $(x^*)$ and projections to models $(m^*)$ visualized in Figure~\ref{fig:case-solutions}. 
Throughout, we report {deployment value} (predicted utility minus a resource penalty). As discussed above, the optimization layer uses estimated quantities ($\widehat{U}(\cdot;z)$ and tier-aggregated costs), whereas the evaluation metric still relies on $\widehat{U}(\cdot;z)$, but uses the selected model's {actual} deployment cost when computing the reported deployment value. Thus, when cost enters the objective ($\lambda>0$), the optimization criterion and the reported evaluation metric need not coincide.
In the pure-capability setting ($\lambda=0$), the deployment value reduces to predicted utility, so the optimization objective and the evaluation metric reported in the tables coincide. Therefore, selecting the best model separately for each type maximizes utility pointwise and also maximizes any average across types. As a result, MLC (per-type) will always achieve the highest deployment value.
Differences arise in cost-aware settings ($\lambda>0$): a per-type policy can pull some types toward relatively expensive models under actual pricing, whereas a single-model policy can score higher by choosing a cheaper model that sacrifices little utility across all types in exchange for a larger reduction in cost.
We report detailed results in Section~\ref{ec:optimization} of the EC.\footnote{For each dataset, deployment scenario, and context type, we list (i) the continuous targets $(x^*,c^*)$ and their implied utility and objective values, and (ii) the corresponding discrete outcomes after model selection (selected tier, selected model, selected $(x_{\mathrm{sel}},c_{\mathrm{sel}})$, and achieved utility and deployment value.}

\begin{table}[t]
\centering
\caption{Deployment values under alternative deployment policies.}
\smallskip \footnotesize \parbox{0.95\linewidth}{\textit{Note.}
Each row is a deployment setting.
Columns report the achieved deployment value for: (i) an observed-winner policy chosen separately by type (Actual per type), (ii) MLC chosen separately by type (MLC per type), (iii) a single-model policy chosen by the observed winner criterion (Actual mean), (iv) a single-model MLC policy optimizing the mean objective (MLC mean), (v) a single-model MLC policy optimizing the worst-case objective (MLC robust), and (vi) single-metric baselines that select models directly on raw inputs.
All objective values are computed using the estimated utility $\widehat{U}$ and the same resource penalty/cost inputs, so policies are evaluated on a common deployment-value proxy; higher is better.
Parentheses report standard deviations across types (for per-type columns) or across baseline rules (for the baseline column).
}
\smallskip 
\label{tab:summary_objective}
\setlength{\tabcolsep}{3pt}
\renewcommand{\arraystretch}{1.15}
\resizebox{\linewidth}{!}{%
\begin{tabular}{llcccccc}
Dataset & Setting 
& Actual (per type) & MLC (per type) 
& Actual (mean) & MLC (mean) & MLC (robust) & Baselines \\ \hline
\multirow{5}{*}{PRISM}       
& Pure capability      
& 0.932 (0.091)   & 0.941 (0.097) & 0.932 & 0.819 & 0.752 & 0.788 (0.056)  \\
& Cost-aware (low)     
& 0.252 (1.010)   & 0.460 (0.189) & 0.252 & 0.538 & 0.538 & 0.460 (0.554)  \\
& Cost-aware (high)    
& -5.860 (10.327) & 0.450 (0.023) & -5.860 & 0.465 & 0.465 & -2.491 (5.409) \\
& Constrained (single) 
& 0.252 (1.010)   & 0.454 (0.182) & 0.252 & 0.538 & 0.538 & 0.460 (0.554)  \\
& Constrained (all)    
& 0.252 (1.010)   & 0.462 (0.191) & 0.252 & 0.538 & 0.538 & 0.460 (0.554)  \\ \hline
\multirow{5}{*}{HealthBench} 
& Pure capability      
& 0.905 (0.073)   & 0.905 (0.073) & 0.905 & 0.774 & 0.774 & 0.530 (0.196)  \\
& Cost-aware (low)     
& 0.619 (0.221)   & 0.643 (0.221) & 0.619 & 0.274 & 0.274 & 0.301 (0.097)  \\
& Cost-aware (high)    
& -1.956 (2.099)  & 0.182 (0.491) & -1.956 & 0.399 & 0.399 & -1.756 (2.109) \\
& Constrained (single) 
& 0.530 (0.170)   & 0.532 (0.158) & 0.530 & 0.274 & 0.274 & 0.361 (0.096)  \\
& Constrained (all)    
& 0.378 (0.148)   & 0.378 (0.148) & 0.378 & 0.274 & 0.274 & 0.306 (0.069)  \\ \hline
\end{tabular}
}
\end{table}

\begin{table}[t]
\centering
\caption{Recommended models under alternative deployment policies.}
\smallskip \footnotesize \parbox{0.95\linewidth}{\textit{Note.}
Each row is a deployment setting.
``MLC'' is the MLC recommendation; ``Actual'' is the observed-winner benchmark.
The ``Mean'' columns report the single-model policies (one model deployed for all types).
We report three headline types in each case study: PRISM (Ethics-focused, Safety-focused, General) and HealthBench (Health Professional, Non-Health Professional, Simple Answer). When multiple feasible models are tied under the objective, we select the lower-cost model.
}
\smallskip 
\label{tab:model_selection}
\setlength{\tabcolsep}{3pt}
\renewcommand{\arraystretch}{1.15}
\resizebox{\linewidth}{!}{%
\begin{tabular}{llcccccccc}
\toprule
Dataset & Setting 
& \multicolumn{2}{c}{Ethics-focused / Health Professional} 
& \multicolumn{2}{c}{Safety-focused / Non-Health Professional} 
& \multicolumn{2}{c}{General / Simple Answer} 
& \multicolumn{2}{c}{Mean} \\
\cmidrule(lr){3-4}\cmidrule(lr){5-6}\cmidrule(lr){7-8}\cmidrule(lr){9-10}
&  & MLC & Actual & MLC & Actual & MLC & Actual & MLC & Actual \\
\midrule

\multirow{5}{*}{PRISM}
& Pure capability 
& cohere-com & cohere-com 
& gpt-4-1106 & gpt-4 
& cohere-com & cohere-com 
& gpt-4-1106 & cohere-com \\
& Cost-aware (low) 
& oasst-pythia-12b & cohere-com 
& gpt-4-1106 & gpt-4 
& mistral-7b-instr & cohere-com 
& mistral-7b-instr & cohere-com \\
& Cost-aware (high) 
& oasst-pythia-12b & cohere-com 
& mistral-7b-instr & gpt-4 
& mistral-7b-instr & cohere-com 
& mistral-7b-instr & cohere-com \\
& Constrained (single) 
& mistral-7b-instr & cohere-com 
& gpt-4-1106 & gpt-4 
& mistral-7b-instr & cohere-com 
& mistral-7b-instr & cohere-com \\
& Constrained (all) 
& llama-2-7b-chat & cohere-com 
& gpt-4-1106 & gpt-4 
& mistral-7b-instr & cohere-com 
& mistral-7b-instr & cohere-com \\

\midrule

\multirow{5}{*}{HealthBench}
& Pure capability 
& qwen3-80b-a3b-nr & qwen3-80b-a3b-r 
& gpt-oss-120B & gpt-oss-120B 
& phi-4 & phi-4
& gpt-oss-120B & gpt-oss-120B \\
& Cost-aware (low) 
& qwen3-80b-a3b-nr & qwen3-80b-a3b-r 
& gpt-oss-120B & gpt-oss-120B 
& phi-4 & phi-4
& gpt-oss-120B & gpt-oss-120B \\
& Cost-aware (high) 
& gemma-3-27b & qwen3-80b-a3b-r 
& gemma-3-12b & gpt-oss-120B 
& phi-4 & phi-4
& Gemma 3 4B & gpt-oss-120B \\
& Constrained (single) 
& qwen3-80b-a3b-nr & qwen3-80b-a3b-r 
& gpt-oss-120B & gpt-oss-120B 
& llama-nem-sup-49b & llama-nem-sup-49b
& gpt-oss-120B & gpt-oss-120B \\
& Constrained (all) 
& qwen3-80b-a3b-r & qwen3-80b-a3b-r 
& gpt-oss-120B & gpt-oss-120B 
& gpt-oss-120B & gpt-oss-120B
& gpt-oss-120B & gpt-oss-120B \\

\bottomrule
\end{tabular}%
}
\end{table}

\paragraph{PRISM.}
Table~\ref{tab:summary_objective} shows that, in the capability-first setting, MLC closely tracks the observed-winner benchmark.
Once deployment becomes cost-aware, the gap widens: observed-winner selections (which do not incorporate costs) can perform poorly under cost penalties, whereas MLC explicitly trades off predicted value against cost.
Table~\ref{tab:model_selection} shows how this trade-off manifests in discrete recommendations across types and settings, while single-metric baselines remain weaker because they ignore preference heterogeneity and reduce deployment to a single raw metric.

\paragraph{HealthBench.}
HealthBench exhibits the same pattern but with a different interpretation of the external benchmark: the observed-winner policy tracks rubric-maximizing models and is therefore naturally aligned with the capability-first objective.
Once we introduce resource penalties, MLC shifts toward models that preserve predicted clinical success while reducing deployment cost.
This effect is most pronounced in settings where many models achieve similarly high predicted utility (e.g., simpler tasks), in which case the optimization (and our tie-breaking) favors the lowest-cost feasible option, as reflected in Table~\ref{tab:model_selection}.

\section{Conclusion}\label{sec:discussion}
This paper develops MLC, a decision-support framework for AI deployment that closes the capability--deployment gap by turning model evaluation inputs into \emph{deployment-value} recommendations. MLC formulates model selection as a constrained optimization problem in which a deployer trades off context-dependent utility against deployment resources while satisfying cost and compliance requirements. A central ingredient is an empirical technological frontier linking feasible capability profiles to resource levels. On the theory side, we characterize optimal targets under a parametric frontier and show a three-regime structure in internal measures: some dimensions bind at compliance minima, some saturate at their feasible maxima, and the rest adjust in the interior according to frontier curvature and cost sensitivity; comparative statics quantify spillovers from tighter requirements and how budget or technology shifts translate into capability profiles. On the implementation side, we operationalize MLC through a pipeline that extracts low-dimensional internal measures from heterogeneous descriptors, estimates the frontier from observed model profiles, learns context-specific utility from outcome data, and produces recommended models and target capability--cost profiles. In two end-to-end case studies (PRISM and HealthBench), MLC yields recommendations and scenario-specific leaderboards that can differ materially from capability-only rankings once costs and requirements are made explicit, and it makes clear \emph{why} a given recommendation is chosen and which constraints are driving it.

\paragraph{Managerial takeaways.}
MLC supports deployment decisions with a small set of actionable outputs: \emph{(i)} replace “benchmark chasing” with \emph{scenario scoring} (predicted utility net of a resource penalty, subject to requirements), so model comparisons are tied to the organization’s operating constraints rather than to a single leaderboard; \emph{(ii)} use the frontier and the three-regime decomposition to diagnose what is binding (compliance floors vs.\ interior trade-offs vs.\ saturation), and to communicate the opportunity cost of tighter standards or lower budgets; \emph{(iii)} decide when to deploy a portfolio: when estimated utility differs across user/task contexts, per-context recommendations can raise value, while in cost-sensitive regimes a single cheaper model can dominate in \emph{deployment value} even if it sacrifices some utility; and \emph{(iv)} strengthen governance by making recommendations traceable to explicit cost and compliance inputs, enabling clearer internal review and more auditable justification when constraints are externally imposed.

\paragraph{Future work.}
Extensions that would bring MLC closer to large-scale practice include: \emph{(i)} dynamic deployment, where prices, model sets, and frontiers evolve over time and switching/platform costs matter; \emph{(ii)} tighter links from the available external outcomes (ratings or rubrics) to downstream organizational KPIs, so the estimated utility can be validated and calibrated as a proxy for true deployment value; and, perhaps most importantly, \emph{(iii)} development of complete deployment-aware leaderboards across application domains.

\begin{APPENDICES}

\end{APPENDICES}

\ACKNOWLEDGMENT{
This research was funded by the National Institute of Standards and Technology (ror.org/05xpvk416) under Federal Award ID Number 60NANB24D231 and Carnegie Mellon University (https://ror.org/05x2bcf33) AI Measurement Science and Engineering Center (AIMSEC).
Parts of this research were conducted using ORCHARD, a high-performance cloud computing cluster. The authors would like to acknowledge Carnegie Mellon University for making this resource available to its community.
}



\bibliographystyle{informs2014} 
\bibliography{references} 

@online{folk2024google,
  author = {Zachary Folk},
  title = {Google Stops {G}emini {AI} From Making Images Of People After Musk Calls Service ``Woke''},
  year = {2024},
  url = {www.forbes.com/sites/zacharyfolk/2024/02/22/google-stops-gemini-ai-from\allowbreak-making-images-of-people-after-musk-calls-service-woke/},
  urldate = {2024-03-15},
  note = {Accessed: 2024/10/01}
}

@article{bertsimas2024towards,
  title={Towards Stable Machine Learning Model Retraining via Slowly Varying Sequences},
  author={Bertsimas, Dimitris and Digalakis Jr, Vassilis and Ma, Yu and Paschalidis, Phevos},
  journal={arXiv preprint arXiv:2403.19871},
  year={2024}
}

@article{arrow1961capital,
  title={Capital-labor substitution and economic efficiency},
  author={Arrow, Kenneth J and Chenery, Hollis B and Minhas, Bagicha S and Solow, Robert M},
  journal={The Review of Economics and Statistics},
  volume={43},
  number={3},
  pages={225--250},
  year={1961},
  publisher={JSTOR}
}

@article{solow1956contribution,
  title={A contribution to the theory of economic growth},
  author={Solow, Robert M},
  journal={The quarterly journal of economics},
  volume={70},
  number={1},
  pages={65--94},
  year={1956},
  publisher={MIT press}
}

@article{bertsimas2023improving,
  title={Improving Stability in Decision Tree Models},
  author={Bertsimas, Dimitris and Digalakis Jr, Vassilis},
  journal={arXiv preprint arXiv:2305.17299},
  year={2023}
}

@article{liu2021stochastic,
  title={The stochastic multi-gradient algorithm for multi-objective optimization and its application to supervised machine learning},
  author={Liu, Suyun and Vicente, Luis Nunes},
  journal={Annals of Operations Research},
  pages={1--30},
  year={2021},
  publisher={Springer}
}

@article{bertsimas2019price,
  title={The price of interpretability},
  author={Bertsimas, Dimitris and Delarue, Arthur and Jaillet, Patrick and Martin, Sebastien},
  journal={arXiv preprint arXiv:1907.03419},
  year={2019}
}

@inproceedings{zafar2017fairness,
  title={Fairness beyond disparate treatment \& disparate impact: Learning classification without disparate mistreatment},
  author={Zafar, Muhammad Bilal and Valera, Isabel and Gomez Rodriguez, Manuel and Gummadi, Krishna P},
  booktitle={Proceedings of the 26th international conference on world wide web},
  pages={1171--1180},
  year={2017}
}

@article{bai2022training,
  title={Training a helpful and harmless assistant with reinforcement learning from human feedback},
  author={Bai, Yuntao and Jones, Andy and Ndousse, Kamal and Askell, Amanda and Chen, Anna and DasSarma, Nova and Drain, Dawn and Fort, Stanislav and Ganguli, Deep and Henighan, Tom and others},
  journal={arXiv preprint arXiv:2204.05862},
  year={2022}
}

@article{hendrycks2020measuring,
  title={Measuring massive multitask language understanding},
  author={Hendrycks, Dan and Burns, Collin and Basart, Steven and Zou, Andy and Mazeika, Mantas and Song, Dawn and Steinhardt, Jacob},
  journal={arXiv preprint arXiv:2009.03300},
  year={2020}
}

@article{fodor2025line,
  title={Line goes up? inherent limitations of benchmarks for evaluating large language models},
  author={Fodor, James},
  journal={arXiv preprint arXiv:2502.14318},
  year={2025}
}

@article{chen2021evaluating,
  title={Evaluating large language models trained on code},
  author={Chen, Mark},
  journal={arXiv preprint arXiv:2107.03374},
  year={2021}
}

@inproceedings{lin2022truthfulqa,
  title={Truthfulqa: Measuring how models mimic human falsehoods},
  author={Lin, Stephanie and Hilton, Jacob and Evans, Owain},
  booktitle={Proceedings of the 60th annual meeting of the association for computational linguistics (volume 1: long papers)},
  pages={3214--3252},
  year={2022}
}

@article{aigner1977formulation,
  title={Formulation and estimation of stochastic frontier production function models},
  author={Aigner, Dennis and Lovell, CA Knox and Schmidt, Peter},
  journal={Journal of econometrics},
  volume={6},
  number={1},
  pages={21--37},
  year={1977},
  publisher={Elsevier}
}

@article{farrell1957measurement,
  title={The measurement of productive efficiency},
  author={Farrell, Michael James},
  journal={Journal of the royal statistical society series a: statistics in society},
  volume={120},
  number={3},
  pages={253--281},
  year={1957},
  publisher={Oxford University Press}
}

@article{banker1984some,
  title={Some models for estimating technical and scale inefficiencies in data envelopment analysis},
  author={Banker, Rajiv D and Charnes, Abraham and Cooper, William Wager},
  journal={Management science},
  volume={30},
  number={9},
  pages={1078--1092},
  year={1984},
  publisher={INFORMS}
}

@article{charnes1978measuring,
  title={Measuring the efficiency of decision making units},
  author={Charnes, Abraham and Cooper, William W and Rhodes, Edwardo},
  journal={European journal of operational research},
  volume={2},
  number={6},
  pages={429--444},
  year={1978},
  publisher={Elsevier}
}

@article{wang2021adversarial,
  title={Adversarial glue: A multi-task benchmark for robustness evaluation of language models},
  author={Wang, Boxin and Xu, Chejian and Wang, Shuohang and Gan, Zhe and Cheng, Yu and Gao, Jianfeng and Awadallah, Ahmed Hassan and Li, Bo},
  journal={arXiv preprint arXiv:2111.02840},
  year={2021}
}

@article{weidinger2021ethical,
  title={Ethical and social risks of harm from language models},
  author={Weidinger, Laura and Mellor, John and Rauh, Maribeth and Griffin, Conor and Uesato, Jonathan and Huang, Po-Sen and Cheng, Myra and Glaese, Mia and Balle, Borja and Kasirzadeh, Atoosa and others},
  journal={arXiv preprint arXiv:2112.04359},
  year={2021}
}

@article{ji2023survey,
  title={Survey of hallucination in natural language generation},
  author={Ji, Ziwei and Lee, Nayeon and Frieske, Rita and Yu, Tiezheng and Su, Dan and Xu, Yan and Ishii, Etsuko and Bang, Ye Jin and Madotto, Andrea and Fung, Pascale},
  journal={ACM computing surveys},
  volume={55},
  number={12},
  pages={1--38},
  year={2023},
  publisher={ACM New York, NY}
}

@article{singhal2023large,
  title={Large language models encode clinical knowledge},
  author={Singhal, Karan and Azizi, Shekoofeh and Tu, Tao and Mahdavi, S Sara and Wei, Jason and Chung, Hyung Won and Scales, Nathan and Tanwani, Ajay and Cole-Lewis, Heather and Pfohl, Stephen and others},
  journal={Nature},
  volume={620},
  number={7972},
  pages={172--180},
  year={2023},
  publisher={Nature Publishing Group}
}

@inproceedings{min2023factscore,
  title={Factscore: Fine-grained atomic evaluation of factual precision in long form text generation},
  author={Min, Sewon and Krishna, Kalpesh and Lyu, Xinxi and Lewis, Mike and Yih, Wen-tau and Koh, Pang and Iyyer, Mohit and Zettlemoyer, Luke and Hajishirzi, Hannaneh},
  booktitle={Proceedings of the 2023 Conference on Empirical Methods in Natural Language Processing},
  pages={12076--12100},
  year={2023}
}

@inproceedings{reddi2020mlperf,
  title={Mlperf inference benchmark},
  author={Reddi, Vijay Janapa and Cheng, Christine and Kanter, David and Mattson, Peter and Schmuelling, Guenther and Wu, Carole-Jean and Anderson, Brian and Breughe, Maximilien and Charlebois, Mark and Chou, William and others},
  booktitle={2020 ACM/IEEE 47th Annual International Symposium on Computer Architecture (ISCA)},
  pages={446--459},
  year={2020},
  organization={IEEE}
}

@article{allen2022algorithm,
  title={Algorithm-augmented work and domain experience: The countervailing forces of ability and aversion},
  author={Allen, Ryan and Choudhury, Prithwiraj},
  journal={Organization Science},
  volume={33},
  number={1},
  pages={149--169},
  year={2022},
  publisher={INFORMS}
}

@misc{euaiact2024,
  title = {Regulation (EU) 2024/1689 of the European Parliament and of the Council of 13 June 2024 laying down harmonised rules on artificial intelligence (Artificial Intelligence Act)},
  author = {{European Parliament and Council of the European Union}},
  year = {2024},
  month = {June},
  url = {https://eur-lex.europa.eu/legal-content/EN/TXT/?uri=CELEX:32024R1689},
  note = {OJ L 2024/1689}
}

@article{wei2022emergent,
  title={Emergent abilities of large language models},
  author={Wei, Jason and Tay, Yi and Bommasani, Rishi and Raffel, Colin and Zoph, Barret and Borgeaud, Sebastian and Yogatama, Dani and Bosma, Maarten and Zhou, Denny and Metzler, Donald and others},
  journal={arXiv preprint arXiv:2206.07682},
  year={2022}
}

@article{sukenik2022generalization,
  title={Generalization in multi-objective machine learning},
  author={S{\'u}ken{\'\i}k, Peter and Lampert, Christoph H},
  journal={arXiv preprint arXiv:2208.13499},
  year={2022}
}

@article{cortes2020agnostic,
  title={Agnostic learning with multiple objectives},
  author={Cortes, Corinna and Mohri, Mehryar and Gonzalvo, Javier and Storcheus, Dmitry},
  journal={Advances in Neural Information Processing Systems},
  volume={33},
  pages={20485--20495},
  year={2020}
}

@article{ibrahim2021eliciting,
  title={Eliciting human judgment for prediction algorithms},
  author={Ibrahim, Rouba and Kim, Song-Hee and Tong, Jordan},
  journal={Management Science},
  volume={67},
  number={4},
  pages={2314--2325},
  year={2021},
  publisher={INFORMS}
}

@article{dietvorst2015algorithm,
  title={Algorithm aversion: people erroneously avoid algorithms after seeing them err.},
  author={Dietvorst, Berkeley J and Simmons, Joseph P and Massey, Cade},
  journal={Journal of experimental psychology: General},
  volume={144},
  number={1},
  pages={114},
  year={2015},
  publisher={American Psychological Association}
}

@article{arora2025healthbench,
  title={Healthbench: Evaluating large language models towards improved human health},
  author={Arora, Rahul K and Wei, Jason and Hicks, Rebecca Soskin and Bowman, Preston and Qui{\~n}onero-Candela, Joaquin and Tsimpourlas, Foivos and Sharman, Michael and Shah, Meghan and Vallone, Andrea and Beutel, Alex and others},
  journal={arXiv preprint arXiv:2505.08775},
  year={2025}
}

@article{choudhury2020machine,
  title={Machine learning and human capital complementarities: Experimental evidence on bias mitigation},
  author={Choudhury, Prithwiraj and Starr, Evan and Agarwal, Rajshree},
  journal={Strategic Management Journal},
  volume={41},
  number={8},
  pages={1381--1411},
  year={2020},
  publisher={Wiley Online Library}
}

@article{bertsimas2020predictive,
  title={From predictive to prescriptive analytics},
  author={Bertsimas, Dimitris and Kallus, Nathan},
  journal={Management Science},
  volume={66},
  number={3},
  pages={1025--1044},
  year={2020},
  publisher={INFORMS}
}

@article{bastani2021mostly,
  title={Mostly exploration-free algorithms for contextual bandits},
  author={Bastani, Hamsa and Bayati, Mohsen and Khosravi, Khashayar},
  journal={Management Science},
  volume={67},
  number={3},
  pages={1329--1349},
  year={2021},
  publisher={INFORMS}
}

@article{goyal2007strategic,
  title={Strategic technology choice and capacity investment under demand uncertainty},
  author={Goyal, Manu and Netessine, Serguei},
  journal={Management science},
  volume={53},
  number={2},
  pages={192--207},
  year={2007},
  publisher={INFORMS}
}

@article{fine1990optimal,
  title={Optimal investment in product-flexible manufacturing capacity},
  author={Fine, Charles H and Freund, Robert M},
  journal={Management science},
  volume={36},
  number={4},
  pages={449--466},
  year={1990},
  publisher={INFORMS}
}

@article{kirk2024prism,
  title={The PRISM alignment dataset: What participatory, representative and individualised human feedback reveals about the subjective and multicultural alignment of large language models},
  author={Kirk, Hannah Rose and Whitefield, Alexander and Rottger, Paul and Bean, Andrew M and Margatina, Katerina and Mosquera-Gomez, Rafael and Ciro, Juan and Bartolo, Max and Williams, Adina and He, He and others},
  journal={Advances in Neural Information Processing Systems},
  volume={37},
  pages={105236--105344},
  year={2024}
}

@article{grand2024best,
  title={The best decisions are not the best advice: Making adherence-aware recommendations},
  author={Grand-Cl{\'e}ment, Julien and Pauphilet, Jean},
  journal={Management Science},
  year={2024},
  publisher={INFORMS}
}

@article{dietvorst2018overcoming,
  title={Overcoming algorithm aversion: People will use imperfect algorithms if they can (even slightly) modify them},
  author={Dietvorst, Berkeley J and Simmons, Joseph P and Massey, Cade},
  journal={Management science},
  volume={64},
  number={3},
  pages={1155--1170},
  year={2018},
  publisher={INFORMS}
}

@techreport{hai2025index,
  title     = {Artificial Intelligence Index Report 2025},
  author    = {{Stanford HAI}},
  institution = {Stanford Institute for Human-Centered Artificial Intelligence},
  year      = {2025},
  month     = {April},
  url       = {https://aiindex.stanford.edu/report/},
  note      = {Stanford University, Stanford, CA}
}

@article{srivastava2023beyond,
  title={Beyond the imitation game: Quantifying and extrapolating the capabilities of language models},
  author={Srivastava, Aarohi and Rastogi, Abhinav and Rao, Abhishek and Shoeb, Abu Awal Md and Abid, Abubakar and Fisch, Adam and Brown, Adam R and Santoro, Adam and Gupta, Aditya and Garriga-Alonso, Adri{\`a} and others},
  journal={Transactions on machine learning research},
  year={2023}
}

@techreport{nist2023airmf,
  title        = {Artificial Intelligence Risk Management Framework},
  author       = {{National Institute of Standards and Technology (NIST)}},
  year         = {2023},
  month        = {January},
  institution  = {National Institute of Standards and Technology},
  url          = {https://www.nist.gov/itl/ai-risk-management-framework}
}

@article{liang2022holistic,
  title={Holistic evaluation of language models},
  author={Liang, Percy and Bommasani, Rishi and Lee, Tony and Tsipras, Dimitris and Soylu, Dilara and Yasunaga, Michihiro and Zhang, Yian and Narayanan, Deepak and Wu, Yuhuai and Kumar, Ananya and others},
  journal={arXiv preprint arXiv:2211.09110},
  year={2022}
}

@online{LLMLeaderboard,
  author = {{Artificial Analysis}},
  title = {LLM Leaderboard},
  year = {2025},
  url = {https://artificialanalysis.ai/leaderboards/models},
  note = {Accessed: 2025/10/01}
}

@article{li2025firm,
  title={Firm or fickle? evaluating large language models consistency in sequential interactions},
  author={Li, Yubo and Miao, Yidi and Ding, Xueying and Krishnan, Ramayya and Padman, Rema},
  journal={arXiv preprint arXiv:2503.22353},
  year={2025}
}

@online{AAII,
  author = {{Artificial Analysis}},
  title = {Artificial Analysis Intelligence Benchmarking Methodology},
  year = {2025},
  url = {https://artificialanalysis.ai/methodology/intelligence-benchmarking},
  note = {Accessed: 2025/11/01}
}

@inproceedings{ke2017lightgbm,
  title={LightGBM: A Highly Efficient Gradient Boosting Decision Tree},
  author={Ke, Guolin and Meng, Qi and Finley, Thomas and Wang, Taifeng and Chen, Wei and Ma, Weidong and Ye, Qiwei and Liu, Tie-Yan},
  booktitle={Advances in Neural Information Processing Systems},
  volume={30},
  pages={3146--3154},
  year={2017}
}

@incollection{Joreskog1983,
  author    = {Karl G. J{\"o}reskog},
  title     = {Factor Analysis as an Errors-in-Variables Model},
  booktitle = {Principles of Modern Psychological Measurement},
  editor    = {Fred N. Kerlinger},
  publisher = {Erlbaum},
  address   = {Hillsdale},
  year      = {1983},
  pages     = {185--196},
  isbn      = {0-89859-277-1}
}

@article{HendricksonWhite1964,
  author  = {Hendrickson, A. E. and White, P. O.},
  title   = {Promax: A Quick Method for Rotation to Oblique Simple Structure},
  journal = {British Journal of Statistical Psychology},
  year    = {1964},
  volume  = {17},
  number  = {1},
  pages   = {65--70},
  doi     = {10.1111/j.2044-8317.1964.tb00244.x}
}

@article{Emmerich2018,
  author  = {Emmerich, Michael T. M. and Deutz, André H.},
  title   = {A tutorial on multiobjective optimization: fundamentals and evolutionary methods},
  journal = {Natural Computing},
  year    = {2018},
  volume  = {17},
  number  = {3},
  pages   = {585--609},
  doi     = {10.1007/s11047-018-9685-y},
  url     = {https://doi.org/10.1007/s11047-018-9685-y}
}

@misc{factoranalyzer,
  author       = {Jeremy Biggs and Nitin Madnani},
  title        = {factor\_analyzer},
  year         = {2024},
  version      = {0.5.1},
  url          = {https://github.com/EducationalTestingService/factor\_analyzer},
  note         = {Accessed: 2025‑09‑01}
}

@misc{brown2020languagemodelsfewshotlearners,
      title={Language Models are Few-Shot Learners}, 
      author={Tom B. Brown and Benjamin Mann and Nick Ryder and Melanie Subbiah and Jared Kaplan and Prafulla Dhariwal and Arvind Neelakantan and Pranav Shyam and Girish Sastry and Amanda Askell and Sandhini Agarwal and Ariel Herbert-Voss and Gretchen Krueger and Tom Henighan and Rewon Child and Aditya Ramesh and Daniel M. Ziegler and Jeffrey Wu and Clemens Winter and Christopher Hesse and Mark Chen and Eric Sigler and Mateusz Litwin and Scott Gray and Benjamin Chess and Jack Clark and Christopher Berner and Sam McCandlish and Alec Radford and Ilya Sutskever and Dario Amodei},
      year={2020},
      eprint={2005.14165},
      archivePrefix={arXiv},
      primaryClass={cs.CL},
      url={https://arxiv.org/abs/2005.14165}, 
}

\ECSwitch
\section{Technical Proofs} \label{ec:proofs}
This section provides technical proofs and auxiliary lemmas for the analytical results presented in Section~\ref{sec:stylized-model} of the main paper.

\medskip

\begin{lemma}[Convexity]\label{lem:convexity}
Under Assumptions~\ref{assump:utility}--\ref{assump:frontier}, Problem~\eqref{eq:stylized-problem} is a convex optimization problem in $(x,c)$.
\end{lemma}

\begin{proof}{Proof of Lemma~\ref{lem:convexity}.}
To establish convexity, we show that the objective is concave and the constraint set is convex: maximizing a concave function over a convex set is a convex optimization problem.

The objective $f(x,c) = \sum_{i=1}^I \beta_i x_i - \lambda c$ is linear, hence concave.

The box constraints $R_i \leq x_i \leq 1$ and $0 \leq c \leq B$ define convex sets. For the frontier constraint, we must show that the set
\[
S = \left\{(x,c) : \left(\sum_{i=1}^I a_i x_i^b\right)^{1/b} \leq c_0 c^d\right\}
\]
is convex. Equivalently, we show that $g(x,c) = \left(\sum_{i=1}^I a_i x_i^b\right)^{1/b} - c_0 c^d$ is convex.
For $b \geq 1$, the function $h(x) = \|a_i^{1/b} x_i\|_b$ is the weighted $\ell_b$-norm, which is convex for $b \geq 1$ (standard result). 
Since $0 < d \leq 1$, the function $k(c) = c_0 c^d$ is concave on $\mathbb{R}_+$. Therefore, $g(x,c) = h(x) - k(c)$ is convex as the sum of a convex function and the negative of a concave function.
The constraint set is the intersection of convex sets, hence convex. \qedsymbol
\end{proof}

\bigskip

\begin{lemma}[Slater’s condition]\label{lem:slater}
If
\(
\sum_{i=1}^I a_i R_i^{\,b} \;<\; c_0^{\,b} B^{\,bd},
\)
then Slater’s condition holds for Problem~\eqref{eq:stylized-problem}.
\end{lemma}

\begin{proof}{Proof of Lemma~\ref{lem:slater}.}
Denote
\(
F_X(x) := \Big(\sum_{i=1}^I a_i x_i^{\,b}\Big)^{1/b}.
\)
By continuity of $F_X$, and since
\(
F_X(R) = \Big(\sum_{i=1}^I a_i R_i^{\,b}\Big)^{1/b} < c_0 B^{d},
\)
there exists $\varepsilon>0$ such that
$F_X(x^\circ) < c_0 B^{d}$ 
for 
$x_i^\circ := R_i + \varepsilon,\; i \in [I].$
Choose $\varepsilon$ small enough that $R_i < x_i^\circ < 1$ for all $i$.
Define
\[
c_{\min} := \left(\frac{F_X(x^\circ)}{c_0}\right)^{1/d} \geq 0.
\]
Then $F_X(x^\circ) < c_0 B^d$ implies $0 \leq c_{\min} < B$. Pick any
$c^\circ$ with
$c_{\min} < c^\circ < B.$
We then have
\(
F_X(x^\circ)
= c_0 c_{\min}^{\,d}
< c_0 (c^\circ)^d,
\)
which is equivalent to
\(
\Big(\sum_{i=1}^I a_i (x_i^\circ)^{b}\Big)^{1/b}
- c_0 (c^\circ)^d < 0.
\)

Thus, at $(x^\circ,c^\circ)$,
\(
R_i - x_i^\circ < 0,\quad x_i^\circ - 1 < 0,\quad
-c^\circ < 0,\quad c^\circ - B < 0,
\)
and the frontier constraint holds strictly. Hence all inequality constraints
of Problem~\eqref{eq:stylized-problem} are strictly satisfied at
$(x^\circ,c^\circ)$, so Slater’s condition holds. \qedsymbol
\end{proof}

\bigskip

\begin{lemma}[Frontier Constraint is Binding]\label{lem:frontier-active}
Under Assumptions~\ref{assump:utility}--\ref{assump:frontier}, the frontier constraint is binding at any optimal solution $(x^*,c^*)$ to Problem~\eqref{eq:stylized-problem}:
\(
\sum_{i=1}^I a_i (x_i^*)^{b}
\;=\; c_0^{\,b} (c^*)^{bd}.
\)
\end{lemma}

\begin{proof}{Proof of Lemma~\ref{lem:frontier-active}.}
Suppose, for contradiction, that the frontier constraint is not binding at an optimal point
$(x^*,c^*)$:
\(
\sum_{i=1}^I a_i (x_i^*)^{b} \;<\; c_0^{\,b}(c^*)^{bd}.
\)

If $x^*$ does not saturate its upper bounds, we may increase at least one
coordinate $x_i$ slightly while remaining feasible.  
Because $\beta_i>0$, this strictly increases the objective
$\sum_i \beta_i x_i - \lambda c$, contradicting optimality.

If $x^*$ saturates all upper bounds, then the slack inequality implies that
$c^*$ is strictly larger than the minimum cost compatible with
$x^*$.  
Reducing $c^*$ while holding $x^*$ fixed preserves feasibility and
strictly increases the objective because $\lambda>0$, again contradicting
optimality.

Therefore, equality must hold and the frontier must be binding at an optimum. \qedsymbol
\end{proof}

\bigskip

\begin{proof}{Proof of Theorem~\ref{thm:optimal-solution}.}
Let $(\mathcal{P}_\le)$ denote Problem~\eqref{eq:stylized-problem} and let Assumptions~\ref{assump:utility}--\ref{assump:nondeg} hold.

\bigskip

\noindent \textbf{KKT for $(\mathcal{P}_\le)$.}
By Lemma~\ref{lem:convexity}, $(\mathcal{P}_\le)$ is a convex optimization problem and, by Lemma~\ref{lem:slater},  Slater's condition holds, so the KKT conditions are necessary and sufficient for optimality of $(\mathcal{P}_\le)$.

\medskip

\noindent \textbf{Equality-frontier reformulation.}
By Lemma~\ref{lem:frontier-active}, any optimal solution $(x^*,c^*)$ of $(\mathcal{P}_\le)$ satisfies the frontier constraint with equality. Let $(\mathcal{P}_=)$ denote the problem obtained by replacing the frontier
inequality by this equality constraint:
\begin{equation}\label{prob:eq-frontier} \tag{$\mathcal{P}_=$}
\begin{aligned}
\max_{x,c}\quad & \sum_{i=1}^I \beta_i x_i - \lambda c\\
\text{s.t.}\quad
& R_i \le x_i \le 1,\quad i=1,\dots,I,\\
& 0 \le c \le B,\\
& \sum_{i=1}^I a_i x_i^b = c_0^b c^{bd}.
\end{aligned}
\end{equation}
Every optimal solution of $(\mathcal{P}_\le)$ is feasible for $(\mathcal{P}_=)$ and attains the same objective value, and conversely any optimal solution of $(\mathcal{P}_=)$ is clearly feasible for $(\mathcal{P}_\le)$. Thus $(\mathcal{P}_\le)$ and $(\mathcal{P}_=)$ have the same optimal value and the same set of optimal solutions. In what follows, we work with $(\mathcal{P}_=)$ purely as an equivalent representation of the optimum of $(\mathcal{P}_\le)$.

\medskip

\noindent \textbf{KKT for $(\mathcal{P}_=)$.}
We now derive the KKT conditions for $(\mathcal{P}_=)$. 
For each $i\in[I]$, let $\mu_i^-\ge0$ and $\mu_i^+\ge0$ denote the multipliers for the lower and upper bounds $R_i \le x_i$ and $x_i \le 1$, respectively. 
Let $\nu^-\ge0$ and $\nu^+\ge0$ be the multipliers for $0\le c$ and $c\le B$, respectively.
Let $\mu_0\in\mathbb{R}$ be the multiplier for the equality constraint $\sum_{i=1}^I a_i x_i^b = c_0^b c^{bd}$.
The Lagrangian of $(\mathcal{P}_=)$ is
\[
\mathcal{L}(x,c;\mu,\nu,\mu_0)
= \sum_{i=1}^I \beta_i x_i - \lambda c
  - \sum_{i=1}^I \mu_i^- (R_i - x_i)
  - \sum_{i=1}^I \mu_i^+ (x_i - 1)
\]
\[
  - \nu^-(-c) - \nu^+(c-B)
  - \mu_0\Big(\sum_{i=1}^I a_i x_i^b - c_0^b c^{bd}\Big).
\]
The KKT conditions are:
\begin{itemize}
\item \emph{Primal feasibility:}
\(
R_i \le x_i \le 1,\ i \in [I],\qquad
0 \le c \le B,\qquad
\sum_{i=1}^I a_i x_i^b = c_0^b c^{bd}.
\)

\item \emph{Dual feasibility:}
\(
\mu_i^-,\mu_i^+,\nu^-,\nu^+ \ge 0, \mu_0\in\mathbb{R}.
\)

\item \emph{Complementary slackness:}
\(
\mu_i^-(R_i - x_i) = 0,\quad
\mu_i^+(x_i - 1) = 0,\quad
\nu^-(-c) = 0,\quad
\nu^+(c-B) = 0.
\)

\item \emph{Stationarity:} for each $i\in[I]$,
\begin{equation}\label{eq:kkt-xi}
\frac{\partial\mathcal{L}}{\partial x_i}
= \beta_i + \mu_i^- - \mu_i^+ - \mu_0 a_i b x_i^{b-1} = 0,
\end{equation}
and
\begin{equation}\label{eq:kkt-c}
\frac{\partial\mathcal{L}}{\partial c}
= -\lambda + \nu^- - \nu^+ + \mu_0 c_0^b b d c^{bd-1} = 0.
\end{equation}
\end{itemize}
Let $(x^*,c^*)$ be an optimal solution of $(\mathcal{P}_=)$ (and hence of $(\mathcal{P}_\le)$) with associated multipliers $(\mu^{-*},\mu^{+*},\nu^{-*},\nu^{+*},\mu_0^*)$.
For notational simplicity, we write $(x,c)$ instead of $(x^*,c^*)$ in the characterization of the optimal solutions below.

\medskip

\noindent \textbf{Characterization of $x^*$.}
Fix $i\in[I]$. We distinguish three cases:

\smallskip \noindent
\emph{(i) Lower bound active: $x_i = R_i$.}  
Then $R_i - x_i = 0$ so $\mu_i^-$ may be positive, while $x_i-1<0$ implies $\mu_i^+=0$. 
Hence \eqref{eq:kkt-xi} becomes
\(
\mu_i^- = \mu_0 a_i b R_i^{b-1} - \beta_i.
\)
Dual feasibility $\mu_i^-\ge0$ is equivalent to
\(
\beta_i \le \mu_0 a_i b R_i^{b-1},
\)
and in this case $x_i=R_i$ satisfies the KKT system.

\smallskip \noindent
\emph{(ii) Upper bound active: $x_i = 1$.}  
Here $x_i-1=0$ so $\mu_i^+\ge0$, while $R_i-x_i<0$ implies $\mu_i^-=0$.
Hence \eqref{eq:kkt-xi} becomes
\(
\mu_i^+ = \beta_i - \mu_0 a_i b.
\)
Dual feasibility $\mu_i^+\ge0$ is equivalent to
\(
\beta_i \ge \mu_0 a_i b,
\)
and in this case $x_i=1$ satisfies the KKT system.

\smallskip \noindent
\emph{(iii) Interior coordinate: $R_i < x_i < 1$.}  
If $x_i$ lies strictly between its bounds, then both box constraints are slack, so $\mu_i^-=\mu_i^+=0$. 
Hence \eqref{eq:kkt-xi} becomes
\(
\beta_i = \mu_0 a_i b x_i^{b-1}
\ \Rightarrow\ 
x_i = \left(\frac{\beta_i}{\mu_0 a_i b}\right)^{\tfrac{1}{b-1}}.
\)
For this value to be interior, we must have
\(
R_i < \left(\frac{\beta_i}{\mu_0 a_i b}\right)^{\tfrac{1}{b-1}} < 1,
\)
which is equivalent to
\(
\mu_0 a_i b R_i^{b-1} < \beta_i < \mu_0 a_i b.
\)

\smallskip \noindent
Combining the three cases yields the three-tier characterization stated in the theorem.

\medskip

\noindent \textbf{Characterization of $c^*$.}
We now analyze~\eqref{eq:kkt-c}. In the nondegenerate case where at least one $R_i>0$, the frontier constraint implies $c>0$ for any feasible solution, so the lower bound $c\ge0$ can never be active at an optimum. We therefore only need to distinguish the interior case $0<c<B$ and the budget-binding case $c=B$.

\smallskip \noindent
\emph{(i) Interior spend: $0 < c < B$.}  
If $c$ lies strictly inside its bounds, then both $0\le c$ and $c\le B$ are slack, so $\nu^-=\nu^+=0$ by complementary slackness. Then, assuming $bd\neq1$, \eqref{eq:kkt-c} reduces to
\(
\lambda = \mu_0 c_0^b b d c^{bd-1}
\ \Rightarrow\ 
c = \left(\frac{\lambda}{\mu_0 c_0^b b d}\right)^{\tfrac{1}{bd-1}}.
\)

If $bd=1$, the stationarity condition reduces to
\(
\lambda = \mu_0 c_0^b b d,
\)
so any interior optimal solution must satisfy $\mu_0 = \lambda/(c_0^b b d)$, and the corresponding $c^*$ is then determined implicitly from the frontier equality $\sum_{i=1}^I a_i (x_i^*)^b = c_0^b (c^*)^{bd}$ and the expression for $x^*$. (For expositional simplicity, we present the closed-form expression for $c^*$ in the generic case $bd\neq 1$.)

\smallskip \noindent
\emph{(ii) Budget-binding spend: $c = B$.}  
If $c=B>0$, then the upper bound $c\le B$ is active and $\nu^+\ge0$, while the
lower bound $0\le c$ is slack and $\nu^-=0$. Substituting into
\eqref{eq:kkt-c} with $c=B$ yields
\(
\nu^+ = \mu_0 c_0^b b d B^{bd-1} - \lambda.
\)
Dual feasibility $\nu^+\ge0$ is equivalent to
\(
\mu_0 c_0^b b d B^{bd-1} \;\ge\; \lambda,
\)
and in this case $c^*=B$ satisfies the KKT system.

\smallskip \noindent
Combining the two cases yield the expression in the theorem.

\medskip

\noindent \textbf{Sign of $\mu_0$ and uniqueness.}
In the original inequality formulation $(\mathcal{P}_\le)$, the frontier
constraint is binding at optimum (Lemma~\ref{lem:frontier-active}) and further
relaxing it strictly increases the objective (since all $\beta_i>0$ and
$\lambda>0$). Hence the associated inequality multiplier is strictly positive,
and the corresponding $\mu_0$ in $(\mathcal{P}_=)$ can be chosen strictly
positive. Finally, under $b>1$, the map $x\mapsto\sum_i a_i x_i^b$ is strictly
convex, and combined with the linear objective and frontier equality this
implies that the optimal solution $(x^*,c^*)$ is unique. This completes
the proof. \qedsymbol
\end{proof}

\newpage

\begin{proof}{Proof of Proposition~\ref{thm:budget-sensitivity}.}
Let Assumptions~\ref{assump:utility}-\ref{assump:nondeg} hold.
Fix $B$ and let $(x^*,c^*)$ be an optimal solution to Problem \ref{eq:stylized-problem}  with $c^*=B>0$. 
We define
$S:=\{\,i : R_i < x_i^* < 1\,\}$,
$Y^* := \sum_{j\in S} a_j (x_j^*)^b$, and
$W^* := \sum_{j=1}^I a_j (x_j^*)^b.$
To simplify exposition, we further assume $S \not=\emptyset \Rightarrow Y^*>0$.
We take comparative statics under a standard fixed-active-set assumption: the active set of constraints remains unchanged as $B$ varies locally. 
We are interested in
\(
{\partial x_i^*}/{\partial B}
\text{ and }
\varepsilon_i
:= {\partial \ln x_i^*}/{\partial \ln B} = 0
\)
for each $i \in [I]$.

\medskip
\noindent\textbf{Boundary coordinates.}
If $i\notin S$, then $x_i^*$ is at a bound ($x_i^* = R_i$ or $x_i^* = 1$) and, by the fixed-active-set assumption, this bound remains binding in
a neighborhood of $B$. Thus,
\[
\frac{\partial x_i^*}{\partial B} = 0,
\qquad
\varepsilon_i
= \frac{\partial \ln x_i^*}{\partial \ln B} = 0,
\quad i\notin S.
\]

\medskip
\noindent\textbf{Average elasticity.}
Since $c^*=B>0$ and the frontier binds (Lemma~\ref{lem:frontier-active}), we have
\begin{equation}
\big(W^*\big)^{1/b}
= \Big(\sum_{i=1}^I a_i (x_i^*)^b\Big)^{1/b}
= c_0 B^d
\quad \Leftrightarrow \quad
\frac{1}{b}\ln W^* = \ln c_0 + d\ln B.
\label{eq:prop-1-frontier}
\end{equation}
We differentiate \eqref{eq:prop-1-frontier} with respect to $\ln B$, yielding
\begin{equation}
\frac{1}{b}\cdot\frac{1}{W^*}\frac{\partial W^*}{\partial \ln B} = d.
\label{eq:prop-1-frontier-derivative-1}
\end{equation}
Next, using the fact that
\[
\varepsilon_i = \frac{\partial \ln x_i^*}{\partial \ln B}
= \frac{1}{x_i^*}\frac{\partial x_i^*}{\partial \ln B}
\quad \Rightarrow \quad 
\partial x_i^*/\partial \ln B = x_i^*\varepsilon_i,
\]
we explicitly compute the derivative ${\partial W^*}/{\partial \ln B}$ as
\begin{equation}
\frac{\partial W^*}{\partial \ln B}
= \sum_{i=1}^I a_i b (x_i^*)^{b-1}\frac{\partial x_i^*}{\partial \ln B}
= \sum_{i=1}^I a_i b (x_i^*)^b \varepsilon_i.
\label{eq:prop-1-derivative-w}
\end{equation}
Substituting \eqref{eq:prop-1-derivative-w} into the frontier derivative \eqref{eq:prop-1-frontier-derivative-1} and cancelling $b$,
\begin{equation}
\frac{1}{W^*}\sum_{i=1}^I a_i (x_i^*)^b \varepsilon_i = d.
\label{eq:avg-epsilon}
\end{equation}

\medskip
\noindent\textbf{Stationarity for interior coordinates.}
We now focus on coordinates $i \in S$.
Let $\mu_0$ denote the Lagrange multiplier on the frontier constraint in its
norm form,
\(
\Big(\sum_{j=1}^I a_j x_j^b\Big)^{1/b} \le c_0 c^d,
\)
and let $\mu_i^-, \mu_i^+$ denote the multipliers on the lower and upper bound constraints on $x_i$.
For $i\in S$ the bounds are slack, so complementary slackness gives
$\mu_i^-=\mu_i^+=0$ and stationarity for $x_i$ reduces to
\[
\beta_i
= \mu_0 a_i x_i^{\,b-1} (W^*)^{1/b-1}
\quad \Leftrightarrow \quad
\ln \beta_i
= \ln \mu_0 + \ln a_i + (b-1)\ln x_i^*
  + \Big(\tfrac{1}{b}-1\Big)\ln W^*.
\]
Differentiating with respect to $\ln B$ (noting that $\beta_i,a_i$ are
constants),
\begin{equation}
0
= \frac{\partial \ln \mu_0}{\partial \ln B}
  + (b-1)\varepsilon_i
  + \Big(\tfrac{1}{b}-1\Big)\frac{\partial \ln W^*}{\partial \ln B},
\qquad i\in S.
\label{eq:prop-1-stationarity}
\end{equation}
From the frontier equation \eqref{eq:prop-1-frontier}, we obtain
$\partial \ln W^*/\partial \ln B = bd$. Plugging into \eqref{eq:prop-1-stationarity},
\begin{equation}
0
= \frac{\partial \ln \mu_0}{\partial \ln B}
  + (b-1)\varepsilon_i
  + \Big(\tfrac{1}{b}-1\Big)bd
\quad \Rightarrow \quad
(b-1)\varepsilon_i
= -\,\frac{\partial \ln \mu_0}{\partial \ln B}
  + (b-1)d,
\qquad i\in S.
\label{eq:epsilon-common}
\end{equation}
The right-hand side in \eqref{eq:epsilon-common} does not depend on $i$, so all interior measures share a
common elasticity: there exists $\varepsilon$ such that $\varepsilon_i=\varepsilon$
for all $i\in S$.

\medskip
\noindent\textbf{Solving for the common elasticity.}
For $i\notin S$, we have $\varepsilon_i=0$, so in
\eqref{eq:avg-epsilon} only $i \in S$ contribute:
\[
\frac{1}{W^*}\sum_{i\in S} a_i (x_i^*)^b \varepsilon = d
\quad \Rightarrow \quad
\varepsilon\,Y^* = d\,W^*
\quad\Rightarrow\quad
\varepsilon = d\,\frac{W^*}{Y^*}.
\]

\medskip
\noindent\textbf{Level derivatives for interior coordinates.}
Therefore, for each $i \in S$,
\[
\varepsilon_i
= \frac{\partial \ln x_i^*}{\partial \ln B} = \varepsilon = d W^*/Y^*,
\qquad
\frac{\partial x_i^*}{\partial B} = \frac{x_i^*}{B}\frac{\partial \ln x_i^*}{\partial \ln B}
= \frac{W^* x_i^* d}{Y^* B}.
\]
Finally, in the special case where all measures are interior, $S=[I]$
and hence $Y^* = W^*$, we obtain $\varepsilon = d$. This completes the
proof. \qedsymbol
\end{proof}

\bigskip

\begin{proof}{Proof of Proposition~\ref{prop:regulatory-impact}.}
Let Assumptions~\ref{assump:utility}-\ref{assump:nondeg} hold.
Fix $R_k$ and let $(x^*,c^*)$ be an optimal solution to Problem \ref{eq:stylized-problem}. 
We define
$S:=\{\,i\neq k : R_i < x_i^* < 1\,\}$,
$Y^* := \sum_{j\in S} a_j (x_j^*)^b$, and
$W^* := \sum_{j=1}^I a_j (x_j^*)^b.$
To simplify exposition, we further assume $S \not=\emptyset \Rightarrow Y^*>0$ and $b>1$, $bd\neq1$.
We work with the equivalent formulation \eqref{prob:eq-frontier} in which the frontier is imposed as an equality,
\[
\sum_{i=1}^I a_i (x_i^*)^b = c_0^b (c^*)^{bd} \quad \Leftrightarrow \quad W^* = c_0^b (c^*)^{bd}.
\tag{F}
\label{eq:frontier-equality}
\]
The Lagrangian and KKT conditions for Problem \eqref{prob:eq-frontier} are given in the proof of Theorem~\ref{thm:optimal-solution}. 
We take comparative statics under a standard fixed-active-set assumption: the active set of constraints remains unchanged as $R_k$ varies locally. 


\medskip
\noindent\textbf{Part (i): Direct effect.}
If $x_k^* = R_k$, then feasibility and the fixed-active-set assumption require $x_k^*(R_k+\varepsilon)=R_k+\varepsilon$ for sufficiently small $\varepsilon$, hence $\partial x_k^*/\partial R_k = 1$; as $R_k$ increases marginally, $x_k^*$ must increase by the same amount to maintain feasibility.
If $x_k^*>R_k$, the lower bound on $x_k$ is slack and does not enter the binding KKT system; for small perturbations of $R_k$ that preserve slackness, the solution remains unchanged, so $\partial x_k^*/\partial R_k=0$.

\medskip
\noindent\textbf{Preliminaries for parts (ii) and (iii).}
Differentiating the frontier equality \eqref{eq:frontier-equality} yields
\begin{equation}
\sum_{i=1}^I a_i b (x_i^*)^{b-1}\,\partial x_i^*
= c_0^b b d (c^*)^{bd-1}\,\partial c^*.
\tag{DF}
\label{eq:frontier-diff-generic}
\end{equation}
For any $i \in S$, complementary slackness gives $\mu_i^-=\mu_i^+=0$ and stationarity reduces to
\begin{equation}
\beta_i = \mu_0 a_i b (x_i^*)^{b-1}.
\tag{K$_i$}
\label{eq:Ki}
\end{equation}
Differentiating \eqref{eq:Ki} with respect to $R_k$ 
gives
\begin{equation}
0 = (x_i^*)^{b-1}\,\partial\mu_0
    + \mu_0 (b-1)(x_i^*)^{b-2}\,\partial x_i^*
\quad \Longrightarrow \quad 
\partial x_i^*
= -\,\frac{x_i^*}{\mu_0(b-1)}\,\partial\mu_0
\quad\text{for all }i \in S.
\tag{DK$_i$}
\label{eq:dxi-mu0}
\end{equation}
When $c^*<B$, the cost stationarity condition is
\begin{equation}
\lambda = \mu_0 c_0^b b d (c^*)^{bd-1},
\tag{K$_c$}
\label{eq:Kc}
\end{equation}
and differentiating \eqref{eq:Kc} with respect to $R_k$ gives
\begin{equation}
0 = c_0^b b d\big[ (c^*)^{bd-1}\,\partial\mu_0
+ \mu_0 (bd-1)(c^*)^{bd-2}\,\partial c^*\big]
\quad \Longrightarrow \quad 
\partial\mu_0
= -\,\mu_0 (bd-1)(c^*)^{-1}\,\partial c^*.
\tag{DK$_c$}
\label{eq:dmu0-generic}
\end{equation}
We now apply these generic relations case by case.

\medskip
\noindent\textbf{Part (ii): Effect on cost and spillovers when $c^*=B$.}
Suppose $x_k^*=R_k$ and $c^*=B$. Because $c^*=B$ is fixed in this part, we have $\partial c^*=0$, and
\eqref{eq:frontier-diff-generic} reduces to
\(
\sum_{i=1}^I a_i b (x_i^*)^{b-1}\,\partial x_i^* = 0.
\)
For $i\notin S$, $x_i^*\in\{R_i,1\}$ is pinned by a binding constraint under
the fixed-active-set assumption, and hence $\partial x_i^*=0$.
For $i\in S$, using $\partial x_k^* = \partial R_k$, $\partial x_j^*=0$ for
$j\notin S\cup\{k\}$, and $\partial c^*=0$, we obtain from \eqref{eq:frontier-diff-generic} 
\[
a_k b R_k^{b-1}\,\partial R_k
+ \sum_{i\in S} a_i b (x_i^*)^{b-1}\,\partial x_i^* = 0.
\]
Substituting \eqref{eq:dxi-mu0} and dividing by $b$,
\[
a_k R_k^{b-1}\,\partial R_k
- \frac{1}{\mu_0(b-1)}
  \sum_{i\in S} a_i (x_i^*)^b\,\partial\mu_0
= 0.
\]
By definition of $Y^*$,
\[
a_k R_k^{b-1}\,\partial R_k
- \frac{Y^*}{\mu_0(b-1)}\,\partial\mu_0 = 0
\quad\Longrightarrow\quad
\partial\mu_0 = \frac{a_k R_k^{b-1}\mu_0(b-1)}{Y^*}\,\partial R_k.
\]
Substituting this into \eqref{eq:dxi-mu0} gives, for $i\in S$,
\[
\partial x_i^*
= -\,\frac{x_i^*}{\mu_0(b-1)}
   \cdot\frac{a_k R_k^{b-1}\mu_0(b-1)}{Y^*}\,\partial R_k
= -\,\frac{a_k R_k^{b-1} x_i^*}{Y^*}\,\partial R_k,
\]
that is,
\[
\frac{\partial x_i^*}{\partial R_k}
= -\,\frac{a_k R_k^{b-1} x_i^*}{Y^*},
\qquad i\in S.
\]

\medskip
\noindent\textbf{Part (iii): Effect on cost and spillovers when $c^*<B$.}
Suppose $x_k^*=R_k$ and $c^*<B$. 
For $i\notin{S}$ with $x_i^*\in\{R_i,1\}$, under the fixed-active-set assumption we again have $\partial x_i^* = 0$.
To analyze the effect on $i\in{S}$ and on $c$, we use $\partial x_k^*=\partial R_k$, 
$\partial x_j^*=0$ for $j\notin{S}\cup\{k\}$ to obtain from \eqref{eq:frontier-diff-generic}
\[
a_k b R_k^{b-1}\,\partial R_k
+ \sum_{i\in{S}} a_i b (x_i^*)^{b-1}\,\partial x_i^*
= c_0^b b d (c^*)^{bd-1}\,\partial c^*.
\]
Dividing by $b$ and substituting \eqref{eq:dxi-mu0},
\[
a_k R_k^{b-1}\,\partial R_k
- \frac{1}{\mu_0(b-1)}
   \sum_{i\in{S}} a_i (x_i^*)^b\,\partial\mu_0
= c_0^b d (c^*)^{bd-1}\,\partial c^*,
\]
that is,
\begin{equation}
a_k R_k^{b-1}\,\partial R_k
- \frac{Y^*}{\mu_0(b-1)}\,\partial\mu_0
= c_0^b d (c^*)^{bd-1}\,\partial c^*.
\label{eq:F-diff-Rk}
\end{equation}

\smallskip
\textit{Effect on cost.}
Substituting \eqref{eq:dmu0-generic}, i.e., differentiated cost stationarity \eqref{eq:Kc}, into \eqref{eq:F-diff-Rk} yields
\[
a_k R_k^{b-1}\,\partial R_k
+ \frac{Y^*(bd-1)}{b-1}(c^*)^{-1}\,\partial c^*
= c_0^b d (c^*)^{bd-1}\,\partial c^*.
\]
Rearranging terms and solving for $\partial c^*$,
\begin{equation}
\frac{\partial c^*}{\partial R_k}
= \frac{a_k R_k^{b-1}}{
c_0^b d (c^*)^{bd-1}
- \dfrac{Y^*(bd-1)}{b-1}(c^*)^{-1}
}.
\label{eq:dc-dRk-raw}
\end{equation}
Using $W^* = \sum_{j=1}^I a_j (x_j^*)^b = c_0^b (c^*)^{bd}$, we have
$c_0^b (c^*)^{bd-1} = W^*/c^*$, so the denominator becomes
\[
\frac{d W^*}{c^*}
- \frac{Y^*(bd-1)}{b-1}(c^*)^{-1}
= \frac{1}{c^*}\left[
d W^* - \frac{bd-1}{b-1}\,Y^*
\right].
\]
Thus \eqref{eq:dc-dRk-raw} simplifies to
\begin{equation}
\frac{\partial c^*}{\partial R_k}
= \frac{a_k R_k^{b-1} c^*}{
d W^* - \dfrac{bd-1}{b-1}\,Y^*
}.
\label{eq:dc-dRk-final}
\end{equation}

\smallskip
\textit{Spillovers.}
For each $i\in S$, we can obtain the corresponding spillover effect on
$x_i^*$ by again substituting \eqref{eq:dmu0-generic} into \eqref{eq:dxi-mu0}, yielding
\[
\partial x_i^*
= -\,\frac{x_i^*}{\mu_0(b-1)}\,\partial\mu_0
= \frac{x_i^*(bd-1)}{b-1}\,(c^*)^{-1}\,\partial c^*,
\qquad i\in S.
\]
Using \eqref{eq:dc-dRk-final}, this gives
\begin{equation}
\frac{\partial x_i^*}{\partial R_k}
= \frac{x_i^*(bd-1)}{b-1}\,(c^*)^{-1}
   \cdot \frac{a_k R_k^{b-1} c^*}{
   d W^* - \dfrac{bd-1}{b-1}\,Y^*
   }
= \frac{a_k R_k^{b-1} x_i^*\,\dfrac{bd-1}{b-1}}{
   d W^* - \dfrac{bd-1}{b-1}\,Y^*
},
\quad i\in S.
\label{eq:dxi-dRk-final-iii}
\end{equation}

\smallskip
\textit{Sign analysis.}
To analyze the sign of \eqref{eq:dc-dRk-final}, note that $a_k>0$, $c^*>0$, and $R_k^{b-1}\ge0$. By
Assumption~\ref{assump:nondeg}, $W^*>0$, and by construction
$0\le Y^*\le W^*$. Hence
\(
d W^* - \frac{bd-1}{b-1}\,Y^*
= W^*\left[
d - \frac{bd-1}{b-1}\frac{Y^*}{W^*}
\right].
\)
The term in brackets is minimized when $Y^*/W^*=1$, in which case
\(
d - \frac{bd-1}{b-1}
= \frac{1-d}{b-1} \;\ge\; 0,
\)
with equality only if $d=1$. Under the additional nondegeneracy assumptions ($0<d<1$ and
at least one interior coordinate), the denominator in \eqref{eq:dc-dRk-final} is strictly positive, so
\(
\frac{\partial c^*}{\partial R_k} > 0.
\)

For the sign of \eqref{eq:dxi-dRk-final-iii}, note that the denominator is the same as in \eqref{eq:dc-dRk-final}, so,
under $0<d<1$ and at least one interior coordinate, it is strictly positive.
Thus, the sign of $\partial x_i^*/\partial R_k$ for $i\in S$ is then determined by
the sign of $(bd-1)$: interior measures increase with $R_k$ when $bd>1$
and decrease when $bd<1$.
This completes the proof. \qedsymbol
\end{proof}

\bigskip

\begin{proof}{Proof of Proposition~\ref{thm:technology-progress}.}
Let Assumptions~\ref{assump:utility}-\ref{assump:nondeg} hold.
Fix all parameters and let $(x^*,c^*)$ be an optimal solution to Problem \ref{eq:stylized-problem}. 
We define
$S:=\{\,i: R_i < x_i^* < 1\,\}$,
$Y^* := \sum_{j\in S} a_j (x_j^*)^b$,
$W^* := \sum_{j=1}^I a_j (x_j^*)^b$,
$Z^* := \sum_{j=1}^I a_j (x_j^*)^b \ln x_j^*$,
and
$Z^{\mathrm{int}*} := \sum_{j\in S} a_j (x_j^*)^b \ln x_j^*$.
To simplify exposition, we further assume $S \not=\emptyset \Rightarrow Y^*>0$, $c^* = B>0$, and $b>1$.
We work with the equivalent formulation \eqref{prob:eq-frontier} in which the frontier is imposed as an equality. By assumption $c^*=B$,
\[
\sum_{i=1}^I a_i (x_i^*)^b = c_0^b (c^*)^{bd} \quad \Leftrightarrow \quad W^* = c_0^b B^{bd}.
\tag{F}
\label{F-eq}
\]
The Lagrangian and KKT conditions for Problem \eqref{prob:eq-frontier} are given in the proof of Theorem~\ref{thm:optimal-solution}. 
We take comparative statics under a standard fixed-active-set assumption: the active set of constraints remains unchanged as one of $(c_0,d,b)$ (as indicated in each part) varies locally. 

For each interior coordinate $i\in S$, complementary slackness implies
$\mu_i^-=\mu_i^+=0$, and the stationarity condition for $x_i$ is
\[
\beta_i = \mu_0 a_i b (x_i^*)^{b-1},
\tag{K$_i$}\label{Kx-eq}
\]
where $\mu_0>0$ is the multiplier on \eqref{F-eq}. For $i\notin S$, $x_i^*$ is fixed at a bound
($R_i$ or $1$) in a neighborhood of the parameter by the fixed--active--set
assumption, so $\partial x_i^*/\partial\theta = 0$ for any parameter
$\theta\in\{c_0,d,b\}$.

\medskip
\noindent\textbf{Part (i): Impact of $c_0$.}
Fix all parameters except $c_0$. Define
\[
\varepsilon_i^{(c)} := \frac{\partial \ln x_i^*}{\partial \ln c_0}.
\]
Differentiating \eqref{F-eq} in log form, that is,
\(
\ln W^* = b\ln c_0 + b d \ln B,
\)
with respect to $\ln c_0$ yields
\[
\frac{\partial \ln W^*}{\partial \ln c_0} = b.
\]
Using $W^* = \sum_i a_i (x_i^*)^b$ and
$\partial W^*/\partial\ln c_0 = \sum_i a_i b (x_i^*)^b \varepsilon_i^{(c)}$,
we get
\begin{equation}
\frac{1}{W^*}\sum_{i=1}^I a_i b (x_i^*)^b \varepsilon_i^{(c)} = b
\quad\Longrightarrow\quad
\frac{1}{W^*}\sum_{i=1}^I a_i (x_i^*)^b \varepsilon_i^{(c)} = 1.
\label{prop-3-avg-c0}
\end{equation}
For $i\notin S$, $x_i^*$ is fixed at a bound and
$\varepsilon_i^{(c)}=0$. For $i\in S$, take logs of
\eqref{Kx-eq}:
\[
\ln\beta_i
= \ln\mu_0 + \ln a_i + \ln b + (b-1)\ln x_i^*.
\]
Differentiating with respect to $\ln c_0$,
\[
0 = \frac{\partial\ln\mu_0}{\partial\ln c_0}
  + (b-1)\varepsilon_i^{(c)}
\quad \Longrightarrow \quad
(b-1)\varepsilon_i^{(c)}
= -\,\frac{\partial\ln\mu_0}{\partial\ln c_0},
\]
so all interior coordinates share a common elasticity
$\varepsilon_i^{(c)} = \varepsilon_c$ for $i\in S$. Substituting into
\eqref{prop-3-avg-c0} and observing that only $i\in S$ contribute,
\[
\varepsilon_c \frac{Y^*}{W^*} = 1
\quad\Longrightarrow\quad
\varepsilon_c = \frac{W^*}{Y^*}.
\]
Thus
\[
\frac{\partial \ln x_i^*}{\partial \ln c_0}
=
\begin{cases}
\displaystyle \frac{W^*}{Y^*}, & i\in S,\\[4pt]
0, & i\notin S,
\end{cases}
\]
and in level form,
\[
\frac{\partial x_i^*}{\partial c_0}
= \varepsilon_c\,\frac{x_i^*}{c_0}
=
\begin{cases}
\displaystyle \frac{W^*}{Y^*}\,\frac{x_i^*}{c_0}, & i\in S,\\[4pt]
0, & i\notin S.
\end{cases}
\]

\medskip
\noindent\textbf{Part (ii): Impact of $d$.}
Fix all parameters except $d$. Define
\[
\varepsilon_i^{(d)} := \frac{\partial \ln x_i^*}{\partial d}.
\]
Differentiating \eqref{F-eq} in log form, that is, $\ln W^* = b\ln c_0 + b d \ln B$, now with respect to $d$ gives
\[
\frac{\partial \ln W^*}{\partial d} = b\ln B.
\]
Using $\partial W^*/\partial d = \sum_i a_i b (x_i^*)^b \varepsilon_i^{(d)}$,
we obtain
\begin{equation}
\frac{1}{W^*}\sum_{i=1}^I a_i (x_i^*)^b \varepsilon_i^{(d)} = \ln B.
\label{avg-d}
\end{equation}
For $i\notin S$, $x_i^*$ is fixed and $\varepsilon_i^{(d)}=0$. For $i\in S$,
the log form of \eqref{Kx-eq} is as above and its derivative with respect to
$d$ yields
\[
0 = \frac{\partial\ln\mu_0}{\partial d}
  + (b-1)\varepsilon_i^{(d)}
\quad \Longrightarrow \quad
(b-1)\varepsilon_i^{(d)}
= -\,\frac{\partial\ln\mu_0}{\partial d},
\]
so all interior coordinates share a common elasticity
$\varepsilon_i^{(d)} = \varepsilon_d$ for $i\in S$. Substituting into
\eqref{avg-d} and summing over $S$ only,
\[
\varepsilon_d \frac{Y^*}{W^*} = \ln B
\quad\Longrightarrow\quad
\varepsilon_d = \frac{W^*}{Y^*}\,\ln B.
\]
Thus
\[
\frac{\partial \ln x_i^*}{\partial d}
=
\begin{cases}
\displaystyle \frac{W^*}{Y^*}\,\ln B, & i\in S,\\[4pt]
0, & i\notin S,
\end{cases}
\]
and in level form,
\[
\frac{\partial x_i^*}{\partial d}
= \varepsilon_d x_i^*
=
\begin{cases}
\displaystyle \frac{W^*}{Y^*}\,\ln B\,x_i^*, & i\in S,\\[4pt]
0, & i\notin S.
\end{cases}
\]

\medskip
\noindent\textbf{Part (iii): Impact of $b$.}
Fix all parameters except $b$.

\smallskip
\emph{(a) Cost.}
When the budget constraint binds and remains binding as $b$ varies locally,
we have $c^*(b) = B$ in a neighborhood of $b$, hence
$\partial c^*/\partial b = 0$.

\smallskip
\emph{(b) Redistributive effects.}
For $i\notin S$, $x_i^*$ is fixed at a bound and does not depend on $b$, so
$\partial x_i^*/\partial b = 0$. For $i\in S$, define
\[
\eta_i := \frac{\partial \ln x_i^*}{\partial b}.
\]
Differentiating the frontier equality \eqref{F-eq} with respect to $b$ yields
\[
\sum_{i=1}^I a_i \frac{\partial}{\partial b}\big((x_i^*)^b\big)
= \frac{\partial}{\partial b}\big(c_0^b B^{bd}\big)
= c_0^b B^{bd}(\ln c_0 + d\ln B)
= W^*(\ln c_0 + d\ln B).
\]
For each $i$,
\[
\frac{\partial}{\partial b}\big((x_i^*)^b\big)
= (x_i^*)^b\ln x_i^*
  + b(x_i^*)^{b-1}\frac{\partial x_i^*}{\partial b}
= (x_i^*)^b\ln x_i^* + b(x_i^*)^b\eta_i,
\]
so, noting that $\eta_i=0$ for $i\notin S$, we obtain
\begin{equation}
Z^* + b\sum_{i=1}^I a_i (x_i^*)^b\eta_i
= W^*(\ln c_0 + d\ln B)
\quad \Longrightarrow \quad
Z^* + b\sum_{i\in S} a_i (x_i^*)^b\eta_i
= W^*(\ln c_0 + d\ln B).
\label{frontier-b}
\end{equation}

Next, for $i\in S$ the interior FOC \eqref{Kx-eq} implies
\[
\ln\beta_i
= \ln\mu_0 + \ln a_i + \ln b + (b-1)\ln x_i^*.
\]
Differentiating with respect to $b$,
\[
0
= \frac{\partial\ln\mu_0}{\partial b}
  + \frac{1}{b}
  + (b-1)\eta_i
  + \ln x_i^*.
\]
Define
\[
\gamma := -\,\frac{\partial\ln\mu_0}{\partial b} - \frac{1}{b}.
\]
Then
\[
(b-1)\eta_i + \ln x_i^* = \gamma,
\qquad i\in S,
\]
or equivalently
\begin{equation}
\eta_i = \frac{\gamma - \ln x_i^*}{b-1},
\qquad i\in S.
\label{eta-i}
\end{equation}

Substituting \eqref{eta-i} into the sum in the LHS of \eqref{frontier-b},
\[
\sum_{i\in S} a_i (x_i^*)^b\eta_i
= \sum_{i\in S} a_i (x_i^*)^b\frac{\gamma - \ln x_i^*}{b-1}
= \frac{1}{b-1}\Big(\gamma Y^* - Z^{\mathrm{int}*}\Big).
\]
Thus, \eqref{frontier-b} becomes
\[
Z^* + \frac{b}{b-1}\big(\gamma Y^* - Z^{\mathrm{int}*}\big)
= W^*(\ln c_0 + d\ln B),
\]
and rearranging,
\[
\frac{b}{b-1}\gamma Y^*
= W^*(\ln c_0 + d\ln B)
  - Z^* + \frac{b}{b-1}Z^{\mathrm{int}*}.
\]
Multiplying both sides by $(b-1)/b$ gives
\[
\gamma Y^*
= \frac{b-1}{b}W^*(\ln c_0 + d\ln B)
  - \frac{b-1}{b}Z^* + Z^{\mathrm{int}*},
\]
so
\begin{equation}
\gamma
= \frac{Z^{\mathrm{int}*}}{Y^*}
  + \frac{b-1}{b}\frac{W^*}{Y^*}
    \Big(\ln c_0 + d\ln B - \frac{Z^*}{W^*}\Big).
\label{gamma-eq}
\end{equation}

Put together, we obtain, for each
$i\in S$,
\(
\frac{\partial \ln x_i^*}{\partial b}
= \eta_i
= \frac{\gamma - \ln x_i^*}{b-1},
\)
with $\gamma$ given by \eqref{gamma-eq}. For $i\notin S$ we have
$\frac{\partial x_i^*}{\partial b} = 0$.
Finally, since $b>1$, for any $i,j\in S$,
\[
\eta_i - \eta_j
= \frac{\ln x_j^* - \ln x_i^*}{b-1}.
\]
Hence
\[
x_i^*>x_j^*
\quad\Longrightarrow\quad
\ln x_i^* > \ln x_j^*
\quad\Longrightarrow\quad
\eta_i < \eta_j.
\]
Thus larger interior coordinates respond less (and may be pushed down more)
than smaller ones when $b$ increases, capturing the redistributive role of
$b$ across dimensions. This completes the proof. \qedsymbol
\end{proof}

\newpage

\section{Empirical Validation: Detailed Analyses \& Extended Results}
\label{ec:datasets}
In this section, we provide details on the two cases studies as well as extended computational results. 

\subsection{PRISM} \label{ec-ssec:prism}

\begin{figure}[!h]
   \centering
   \includegraphics[width=0.5\linewidth]{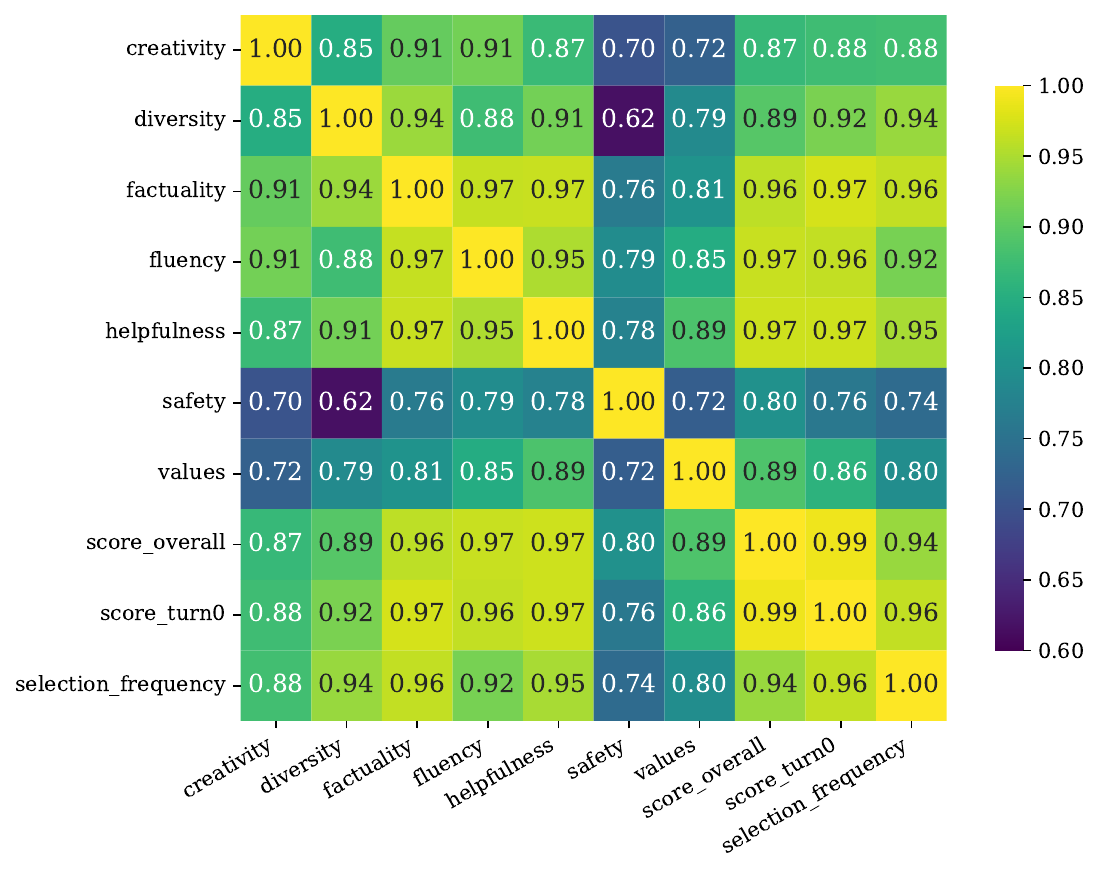}
   \caption{Correlation heatmap of model-internal metrics provided in the PRISM dataset. This set of variables was augmented \emph{a posteriori} with two additional benchmark variables,   MMLU and MTEB, obtained from Hugging Face. The high correlations between the PRISM-derived model performance measures motivates the Measure Extraction step of our pipeline.}
   \label{Fig:prism_correlations}
\end{figure}

We build our first dataset to asses the ML-Compass framework starting from PRISM dataset, which collects tracked human–LLM interactions \citep{kirk2024prism}. In this dataset, a group of users were exposed to diverse large language models and select their preferred one, evaluating the model responses along several dimensions such as helpfulness, factuality, creativity, and safety.  In addition, users were asked to provide their preference profiles over the same attributes prior to interacting with the models, allowing a first depiction of their interests. Using these data, we constructed a model-level dataset by aggregating user evaluations to obtain performance scores for each model across attributes, which we further enriched with external benchmark metrics. The results of this first aggregation step at the model level lead to the PRISM model-level dataset, whose correlations are shown in Figure~\ref{Fig:prism_correlations}.

However, so as to guide our ML-Compass formulation across a user-preference framework, we also developed a user and task classification across the dataset, thanks to the data available, and we further enrich the dataset provided by \cite{kirk2024prism}. 

\paragraph{User type creation:}

To address the user-preference objective of MLC, we first classified PRISM users into distinct types. The goal was to group all users in the dataset into recurrent user profiles, illustrating how different organizations could segment their users or clients based on their interaction patterns with LLMs. To this end, we relied on the previously computed latent space defined by two factors: a conversational experience dimension (C1) and a risk-aware competence dimension (C2).

\begin{figure}[!h]
   \begin{minipage}{0.5\textwidth}
     \centering
     \includegraphics[width=\linewidth]{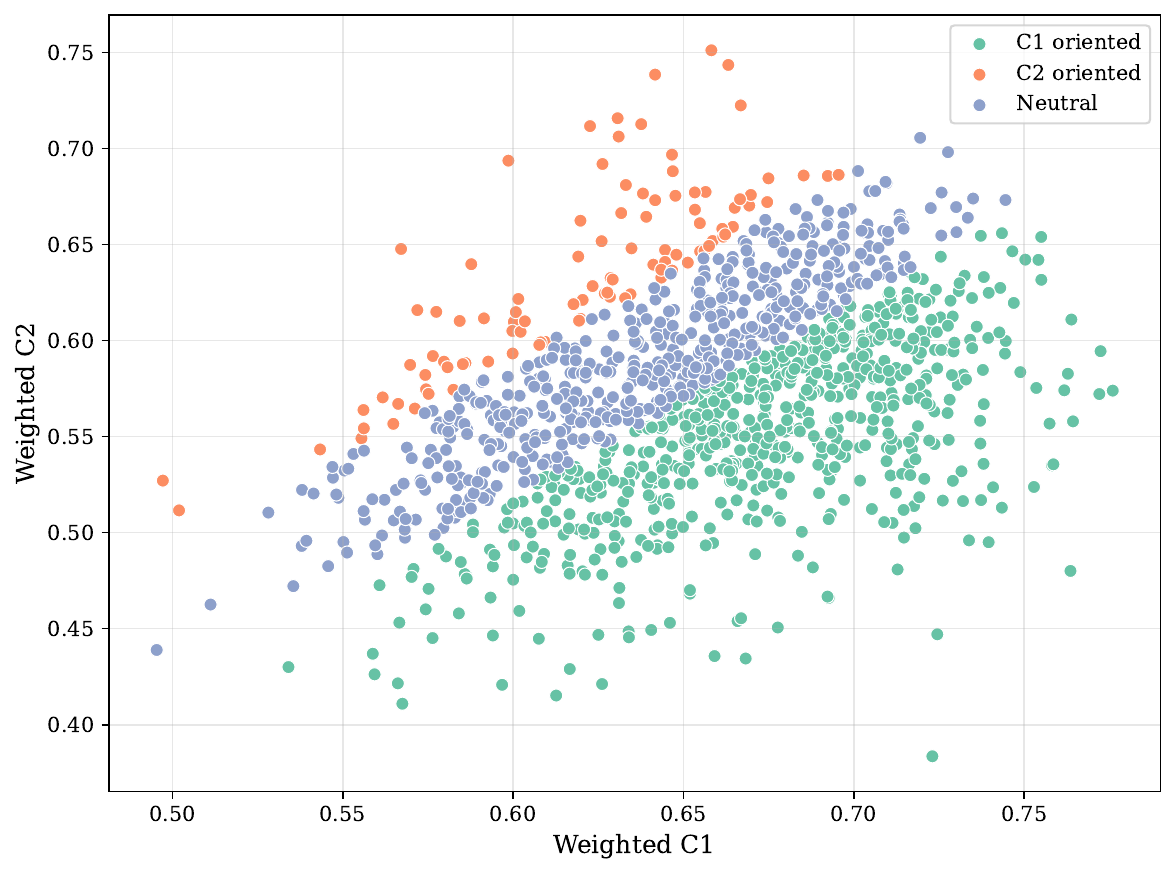}
   \end{minipage}\hspace{0.4cm}
   \begin{minipage}{0.5\textwidth}
     \centering
     \includegraphics[width=\linewidth]{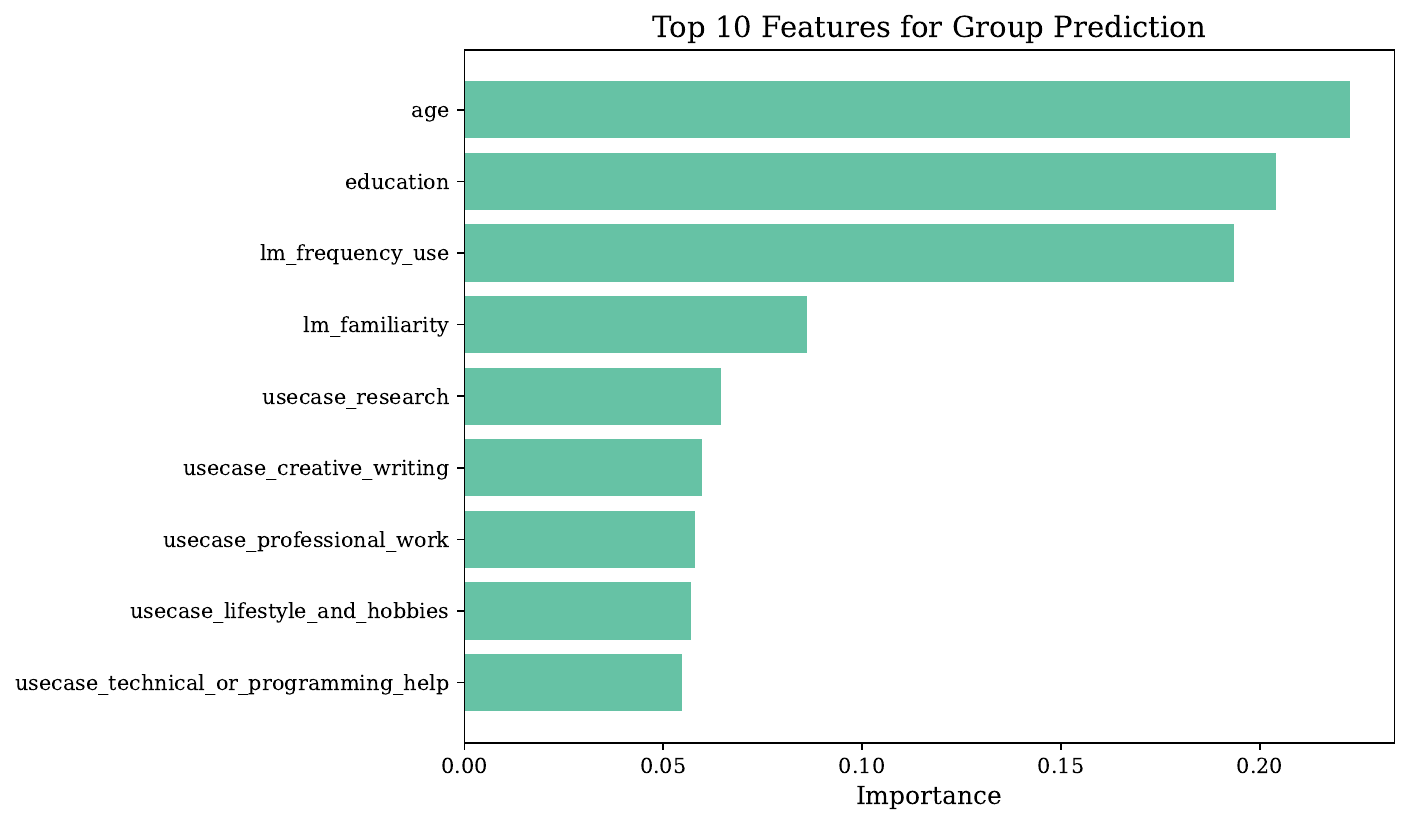}
   \end{minipage}
   \caption{Visualizations for user type creation. The left panel shows the projection of each user in the model space, as well as the segmentation. The right panel illustrates, once the predictor of types had been trained, the most important features to predict the user type clusters.}\label{Fig:prism_users}
\end{figure}

For each user, we computed a weighted average of the C1 and C2 coordinates of the models they interacted with, where weights reflected whether a model’s output was evaluated above each user quality threshold. In this way, user-model interactions of PRISM were mapped to a point in the same C1–C2 space as the models, capturing the types of model capabilities that the users tend to prefer. Then the users were segmented into three types based on the difference between their inferred C1 and C2 preferences. Specifically, we identified C1 oriented users, who favor models with stronger conversational experience, and C2 oriented users, who prefer models performing better along the second dimension, characterized by stronger safety and benchmark-aligned properties. In addition, we defined a neutral user group, comprising users whose preferences lie between these two extremes. The model space projection over the set of users is shown at figure~\ref{Fig:prism_users}, illustrating also the differences in group sizes: over 600 C1 oriented users, over 500 neutral users, and around 100 C2 oriented users.

To control group sizes and avoid imbalance in the final classification, asymmetric thresholds were applied when assigning users to types. This approach ensured stable and interpretable groups across the dataset. Eventually, this procedure yielded a robust user segmentation, which we label as follows. The \textit{Ethics-focused User} group corresponds to users primarily oriented toward the first latent dimension, indicating a stronger concern for human-aligned properties in AI, such as values, diversity, fluency, and overall conversational quality. The \textit{Safety-focused User} group includes users oriented toward the second dimension, emphasizing safety values and benchmark-related performance, suggesting a higher level of familiarity with LLM capabilities and a preference for more technical properties. Finally, the\textit{ General User} group captures users whose preferences fall within the neutral region of the latent space.

After assigning users to types based on revealed preferences, we train a supervised classifier to predict groups from observable user attributes. This step, although not needed for the classification procedure, served us with three purposes: (i) validating the internal consistency of the types, (ii) identifying the features that most strongly characterize each group, and (iii) enabling the prediction of types for new users with limited interaction data. Results can also be found in figure~\ref{Fig:prism_users}.

\begin{figure}[!h]
   \centering
   \includegraphics[width=\linewidth]{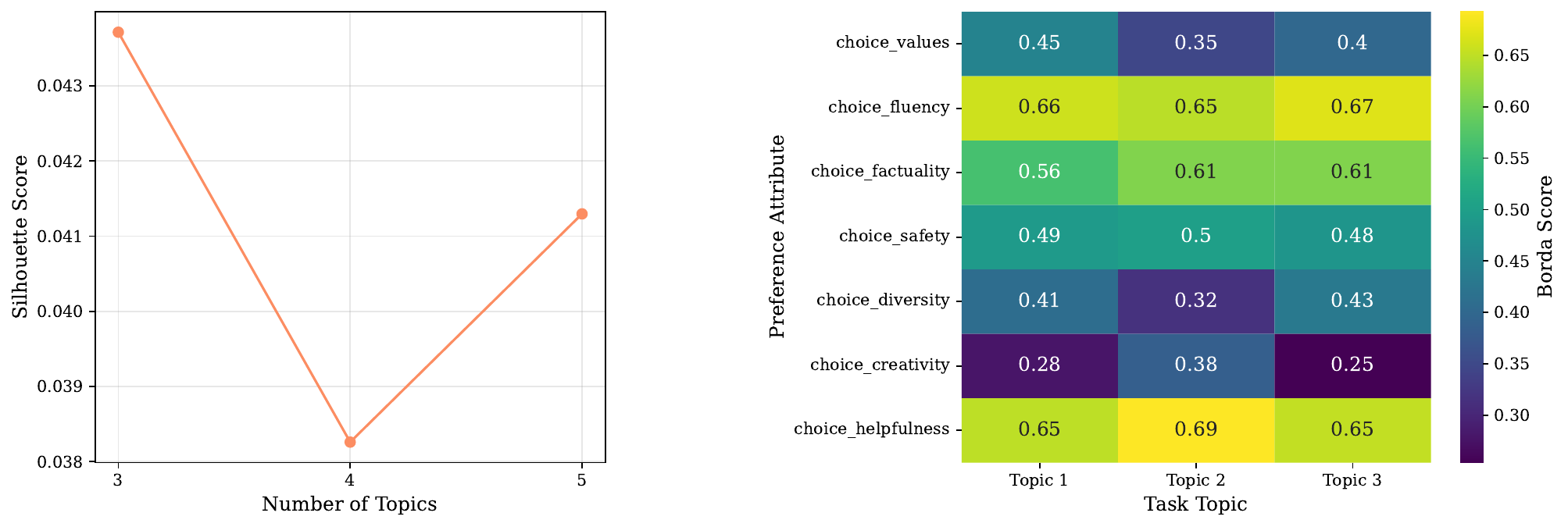}
   \caption{Task topic creation in the PRISM dataset. The left panel shows the evolution of the silhouette index as a function of the number of clusters. The right panel illustrates, for the three-cluster solution, the distribution of Borda preference scores across the resulting task topics.}
   \label{Fig:prism_task_creation}
\end{figure}

\begin{figure}[!h]
   \centering
   \includegraphics[width=\linewidth]{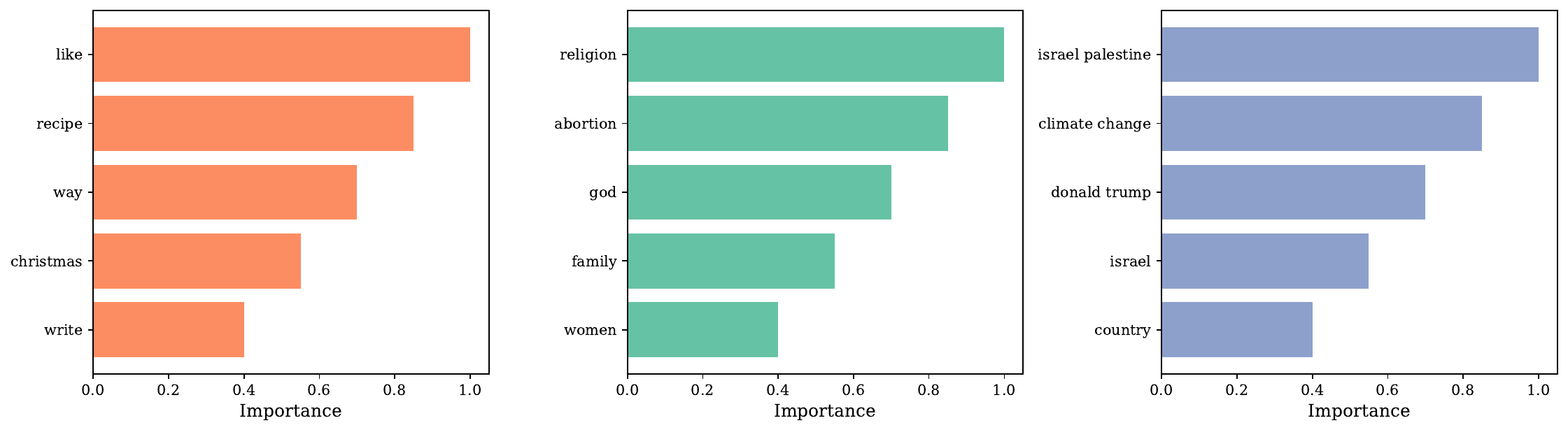}
   \caption{Top keywords in the PRISM dataset for each task topic. Each panel corresponds to one task topic (Topics 1, 2, and 3, respectively) and displays the five most representative keywords identified using TF--IDF analysis.}
   \label{Fig:prism_tasks}
\end{figure}

\paragraph{Task topic creation:} 
We also performed a task-level classification of the conversations in the PRISM dataset in order to identify distinct categories that could be used to account for user preferences. The results, shown in Figure~\ref{Fig:prism_task_creation}, indicate that a three-cluster solution yields the most robust grouping according to the silhouette index. However, an inspection based exclusively on the preference attributes associated with these three task topics does not provide a clear interpretation of the resulting clusters. To gain deeper insight, we conducted an additional analysis of the most representative keywords within each cluster using a TF--IDF approach. The results of this analysis are shown in Figure \ref{Fig:prism_tasks}, and help clarify the semantic meaning underlying each task topic.

This analysis allows us to better characterize the nature of each task topic. The first topic can be interpreted as a general, day-to-day conversational use of LLMs, which is consistent with the relatively high weights assigned to helpfulness and fluency, as shown in Figure~\ref{Fig:prism_task_creation}. The second topic corresponds to a more introspective or reflective use of LLMs, including discussions of spiritual themes, personal concerns, or moral values, again emphasizing fluency and helpfulness. The third topic captures more political or controversial forms of discourse, where fluency, factuality, and helpfulness emerge as particularly relevant evaluation dimensions. From this point onward, we denote the three conversation types as \textit{Everyday Use}, \textit{Reflective Use}, and \textit{Controversial Discourse}.

Despite providing useful qualitative insights, this task-topic characterization remains less rigorous and less effective at capturing user heterogeneity than the user user type construction introduced earlier. In this sense, the lack of standardized datasets that explicitly distinguish different types of tasks in LLM usage represents a clear gap in the Operations Management and decision-making literature, and constitutes a promising direction for future work.

\subsection{HealthBench} \label{ec-ssec:hb}

As we briefly described in Section \ref{ssec:datasets}, HealthBench is a state-of-the-art benchmark developed by OpenAI in collaboration with 262 physicians from 60 countries, designed to assess LLMs in a realistic healthcare context. Its primary objective is to evaluate model behaviour across a wide and clinically representative set of scenarios. To achieve this, HealthBench builds on three main components: the conversations, the rubrics, and the internal classification system. Below, we summarize the most relevant features of each component for our proposal.

\paragraph{Conversations:}
The majority of conversations were synthetically generated after physicians were asked to enumerate the types of situations that should be included. The final dataset—comprising 5,000 conversation prompts—was constructed to ensure diversity across languages, multi-turn interactions, countries, and demographic groups. The resulting set includes a broad range of user personas: general users seeking medical information, healthcare professionals, and highly specialised users, familiar with APIs and large language models. These conversations are intended to reflect realistic use cases, such as summarizing clinical information, identifying diagnosis codes, drafting treatment plans, or assisting with patient communication. The overarching goal is to evaluate model performance in settings that represent authentic medical contexts.

\paragraph{Rubrics:} 

One drawback of using a non-test approach or non-multiple-choice questions is that there is no straightforward method for evaluating model responses. To address this and to design a fully robust, trustworthy, and autonomous benchmark, \cite{arora2025healthbench} leverage the combined strengths of LLM-based assessment and physician expertise by introducing a rubric-based evaluation framework. 

Under this approach, each model response is graded according to a rubric that is tailored to the specific conversation.
For each conversation, physicians were asked to write rubric criteria that could be directly applied to evaluate the responses of the models under assessment. Specifically, each example includes a list of rubric items describing desirable (or undesirable) properties of a correct response. Each rubric criterion is assigned a non-zero score, with negative points associated with undesirable behaviors. To compute the final grade, a model-based grader (specifically GPT-4.1) evaluates each criterion and determines whether the model response satisfies it. If so, the corresponding points—positive or negative—are awarded; otherwise, zero points are assigned. The total score is then divided by the maximum possible score to produce the final normalized grade (with negative totals clipped to zero).

\cite{arora2025healthbench} demonstrate that GPT-4.1 can operate as a reliable model-based grader, achieving performance comparable to that of physicians. This underlies the restriction of using the OpenAI API for assessing the conversations in our evaluation framework.

\paragraph{Classification:} 

A central component of HealthBench is its conversation and evaluation classification system. HealthBench assesses LLMs across seven themes (further divided into seventeen categories) and five evaluation axes. Themes and categories are associated with the conversations themselves, whereas the evaluation axes correspond to the rubric criteria used to grade responses.

Each of the 5,000 original conversations in HealthBench was mapped to one of the seven themes under physician supervision. As a second step, physicians were asked by the OpenAI researchers to assign each example to one of the two or three categories associated with that theme (e.g., emergent, conditionally emergent, or not emergent for the emergency referral theme). When a majority agreement among physicians existed, that categorization was adopted. Only 3,671 conversations achieved this level of annotation, forming HealthBench Consensus—a smaller but more richly annotated dataset with additional metadata and higher-quality labels.

Next, each rubric criterion in a conversation was mapped to one of the five evaluation axes, indicating which aspect of model behavior the criterion assesses. Importantly, not all axes need to be evaluated for every conversation: some examples involve only one dimension, while others span multiple axes. For every conversation in HealthBench Consensus, physicians also crafted consensus rubric criteria—a minimal set of key criteria that are objectively gradable and broadly applicable to all examples within that category.

As a result, each HealthBench evaluation example consists of: (i) a conversation between a user and an LLM (with one or more messages, ending with a user query); (ii) a set of rubric criteria assessing the attributes of an ideal response, each linked to an evaluation axis; (iii) a theme describing the main task; and, optionally, (iv) a category within that theme.

From the full set of 5,000 conversations, we used HealthBench Consensus as the seed for our evaluation dataset. Although this subset is smaller, containing only those conversations with a clear category and consensus criteria, it provides richer metadata and higher-quality annotations—crucial for assessing user preferences in MLC. As \cite{arora2025healthbench} note, HealthBench Consensus can be thought of as a version of HealthBench with a higher level of physician validation, which has greater precision but lower recall in identifying model behavior failures.

Having described the HealthBench dataset, we now explain how we constructed our evaluation dataset from a benchmark—unlike in the PRISM case—and how we developed the task-topic and user-type classifications for this dataset.

As explained in the main paper, a central objective when working with HealthBench was to construct a user-preference dataset starting from a benchmark-only evaluation framework. In other words, we needed to transform a standard benchmark into a dataset that fulfills the requirements outlined in Figure~\ref{fig:compass-flowchart}. Our goal was to derive, from the publicly available benchmark, a set of outcome variables that could be interpreted as utilities perceived by users or organizations when deploying large language models in healthcare contexts. Therefore, we treated the benchmark assessments as a source of outcome data suitable for training a utility function.

In this respect, the richness of HealthBench was crucial. Its detailed classification of conversations and the associated metadata for each evaluation instance enabled us to capture not only performance outcomes, but also contextual dependencies such as user preferences, task topics, and usage scenarios.

Our next objective was to link the HealthBench scores obtained for each model with publicly available cost and capability metrics of the corresponding large language models. To this end, we relied on the information provided by \cite{LLMLeaderboard}, which aggregates a wide range of model-level statistics, including token generation metrics, pricing information, benchmark scores, and technical specifications such as context length and (active) parameter counts. By associating HealthBench outcomes with these model-level metrics, we were able to construct a dataset that jointly captures costs, capabilities, and outcomes aligned with the operational interests of healthcare institutions. Taken together, the HealthBench scores and the statistics from \cite{LLMLeaderboard} provide the necessary components to build a suitable dataset for the MLC framework.

However, \cite{arora2025healthbench} does not report results for many models outside the OpenAI ecosystem, with only a few exceptions (e.g., Llama-based variants). As a result, we needed to execute the HealthBench benchmark ourselves. This motivated our focus on open-source models, including the DeepSeek, Qwen, Gemma, and LLaMA families. Beyond coverage considerations, open-source models also represent a more realistic deployment option for healthcare institutions, where concerns about privacy, accountability, and control are often paramount.

To make this large-scale evaluation feasible, we relied on the ORCHARD cluster, 
a cloud-based research computing infrastructure developed in collaboration with Google, comprising 296 NVIDIA H100 GPUs across 37 compute nodes, each equipped with 6~TB of local SSD storage and multiple high-speed network interfaces. In our experiments, we had access to four nodes, corresponding to up to four NVIDIA H100 GPUs. From our perspective, the execution of HealthBench on a diverse set of open-source, state-of-the-art models already constitutes a valuable contribution to the LLM literature, with potential implications for organizational decision-making and healthcare AI deployment strategies.
All inference was carried out by modifying the publicly available HealthBench codebase, with two major changes. First, we implemented a custom sampling pipeline to generate model responses locally for the selected open-source models. Second, we restructured the evaluation process from an on-the-fly, sequential assessment—where each conversation is evaluated individually via API calls, as in \cite{arora2025healthbench}—to a batch-based approach. In our implementation, we first generated all model responses and subsequently evaluated them in batches using the HealthBench rubrics. This design choice was driven by cost considerations, as batch evaluation via the OpenAI API for GPT-4.1 was approximately half as expensive as the sequential approach. Given a limited evaluation budget, this allowed us to include a larger and more diverse set of models.
Consequently, the final model set was determined by three constraints: (i) the computational limits imposed by the available resources, which allowed us to run models with up to hundreds of billions of parameters; (ii) the availability of corresponding model statistics in \cite{LLMLeaderboard}; and (iii) the budget allocated to API-based evaluation.

Finally, it is important to note that large language model inference involves a wide range of configurable generation parameters. To ensure consistency between our evaluation pipeline and the statistics reported in \cite{LLMLeaderboard}, we carefully aligned our inference settings with those described in \cite{AAII}. This alignment was essential to guarantee comparability between outcome data, cost metrics, and capability measures.

For the second part of the dataset creation, we had to work on differentiating tasks and types within the original set of conversations, as our main objective was enriching as much as possible to facilitate the user consideration of preferences that MLC encourages. For the first classification, we leveraged the opportunity given by the categorical division of HealthBench Consensus. Although we considered the option of just work with themes, we opt for this lower level granularity due to the narrower and specific range of tasks depicted and the possibility of assessing the LLM under a much richer scenario. Therefore, we were able to work with the categories depicted at Table~\ref{tab:hb_task_translation}, that were renamed for us so as to offer a more interpretable idea of the category.

\begin{table}[h]
\centering
\caption{The table shows the original nomenclature used by OpenAI in their analysis (theme and categories) alongside the corresponding terminology we adopted when developing the MLC framework. We also include the number of conversations of each type in the original HealthBench Consensus dataset.}
\label{tab:hb_task_translation}
\begin{tabular}{lll}
\hline
Theme                                                                                        & Category                                  & MLC nomenclature  \\ \hline
\multirow{3}{*}{Emergency referrals}                                                         & Emergent                                  & Immediate Emergency      \\
                                                                                             & Conditionally emergent                    & Conditional Emergency    \\
                                                                                             & Non-emergent                              & No Emergency             \\ \hline
\multirow{2}{*}{\begin{tabular}[c]{@{}l@{}}Expertise tailored \\ communication\end{tabular}} & Health professional user                  & Health Professional       \\
                                                                                             & Non-health professional user              & Non-Health Professional  \\ \hline
\multirow{3}{*}{\begin{tabular}[c]{@{}l@{}}Responding under \\ uncertainty\end{tabular}}     & Any reducible uncertainty                 & Reduce Uncertainty       \\
                                                                                             & Only irreducible uncertainty              & Deal Uncertainty         \\
                                                                                             & No uncertainty                            & No Uncertainty           \\ \hline
\multirow{2}{*}{Response depth}                                                              & Query requiring a simple response         & Simple Answer   \\
                                                                                             & An ideal response is detailed             & Detailed Answer  \\ \hline
\multirow{2}{*}{Health data tasks}                                                           & Enough information to complete task       & Task Can Be Done         \\
                                                                                             & Not enough information to complete task   & Task Needs More Info     \\ \hline
\multirow{3}{*}{Global health}                                                               & Healthcare context matters and is clear   & Context Needed Known     \\
                                                                                             & Healthcare context matters and is unclear & Context Needed Unknown   \\
                                                                                             & Healthcare context does not matter        & Context Not Needed       \\ \hline
\multirow{2}{*}{Context-seeking}                                                             & Enough context                            & Context Sufficient       \\
                                                                                             & Not enough context                        & Context Missing          \\ \hline
\end{tabular}
\end{table}
When it comes to user type creation, the task proved considerably more complex than in the PRISM case, where types could be defined more straightforwardly (or at least more sensibly) thanks to the explicit user preferences available in the original dataset. Although HealthBench provides a richer semantic structure at the conversation level—particularly with respect to task topics—it lacks an extensive characterization of user types, beyond the developers’ intention to include diverse profiles and languages.

Therefore, our analysis relies on two main pillars. First, we build upon the limited information provided by \citet{arora2025healthbench} regarding user types in HealthBench: \emph{“we also aimed to generate conversations across a range of user personas, including individual users seeking information for themselves or a loved one, healthcare professionals, and users of large language models through APIs who may use the models to build applications to, for example, summarize notes, determine diagnosis codes, and message patients.”} Second, we rely on a large language model to perform the classification of each conversation into a user type, given an appropriate set of instructions.

Based on these considerations, we defined four user types: \emph{General Users}, \emph{General Knowledgeable Users}\footnote{We distinguish between general users and general knowledgeable users to obtain a richer and more granular classification.}, \emph{Healthcare Professionals}, and \emph{LLM API Users}. By combining a carefully designed instruction template with a few-shot prompting strategy \citep{brown2020languagemodelsfewshotlearners}, we adapted the LLM to perform the required text classification task. The procedure was as follows:

\begin{enumerate}
    \item We first constructed four idealized conversations, each representative of one of the four user types, to serve as demonstrations for the few-shot prompting setup.
    
    \item We then designed a chat template with the following instructions: (i) based on the four provided examples, (ii) the LLM was required to assign exactly one user type label to each conversation from the predefined set (\emph{General}, \emph{General Knowledgeable}, \emph{Healthcare Professional}, and \emph{LLM API}), and (iii) it was asked to briefly explain the features of the conversation that motivated its classification, in order to provide a degree of transparency.
    
    \item Each conversation, together with the instruction template, was sent to the OpenAI API and evaluated using the GPT-4.1 model, a well-established and widely recognized large language model.
    
    \item Finally, we collected the outputs and obtained the resulting user type classification for each conversation.
\end{enumerate}

Once we had obtained the results for each conversation, we obtained a four classes description for the HealthBench conversation dataset, with the following quantities for each one of the types: 2,154 \emph{General} users, 304 \emph{General Knowledgeable} users, 1,205 \emph{Healthcare Professional} users, and 8 \emph{LLM API} users. Given the shown numbers, we decided to merge both the  \emph{General Knowledgeable} and \emph{LLM API} types into a single one, adding the eight cases of   \emph{LLM API}  to the  \emph{General Knowledgeable} user type.

\subsection{Measure Extraction}
\label{ec:measure-extractor}

The purpose of the factor extraction step is to create new latent variables from the original observed variables using classical Factor Analysis. This transformation enables dimensionality reduction with minimal information loss, to improve the efficiency of the Pareto multi-objective optimization procedure \citep{Emmerich2018}. Classical factor analysis is framed as an errors-in-variables model \citep{Joreskog1983}, in which the covariance structure of the observed variables is explained through a reduced set of common factors.
To conduct the analysis, we rely on the standard formulation of the common factor model and enrich it through two additional well-established procedures: a rotation step and a sparsity-inducing post-processing stage.

The rotation step aims to improve the interpretability of the factor loading matrix without altering the underlying factor solution. We employ promax rotation criteria \citep{HendricksonWhite1964}, which redistribute the loadings to obtain a simpler, more meaningful structure.

The sparsity post-processing further enhances interpretability by encouraging near-zero loadings. Essentially, it imposes a threshold on the loading of each variable and restricts the maximum number of factor contributions per variable. This procedure promotes a clearer association between factors and subsets of variables. It acts as a refinement of the rotated solution, yielding a more interpretable and structurally transparent representation of the latent space.

An important point to examine was how the previously described methods impacted our concrete LLM dataset. To this end, we analyzed how the cumulative variance explained by the factors changed as we modified the configuration parameters of the factor analysis, specifically when enabling rotation and sparsity post-processing. The results for both datasets can be found in Table~\ref{tab:variance_explained}.

\begin{table}[h]
\centering
\caption{The cumulative variance explained by the selected factors varies according to the Factor Analysis approach employed.}
\label{tab:variance_explained}
\begin{tabular}{llcccc}
\hline
Dataset                      & Number of factors & Classical & Rotated & Sparsified & Rotated+Sparsified \\ \hline
\multirow{2}{*}{PRISM}       & 2               & 0.9074    & 0.8800  & 0.8896     & 0.8607             \\
                             & 3               & 0.9425    & 0.8732  & 0.8699     & 0.8429             \\ \hline
\multirow{3}{*}{HealthBench} & 2               & 0.7306    & 0.7420  & 0.6993     & 0.7140             \\
                             & 3               & 0.8048    & 0.7804  & 0.7758     & 0.7504             \\
                             & 4               & 0.8472    & 0.7818  & 0.7805     & 0.7280             \\ \hline
\end{tabular}
\end{table}

As Table \ref{tab:variance_explained} shows, the amount of variance explained varies depending on the specific approach applied to the classical formulation of the factor analysis problem. Interestingly, while the classical formulation of Factor Analysis generally results in higher variance explained as the number of factors increases (as expected), applying complementary methods such as sparsity post-processing or rotation can actually reduce the explained variance at higher dimensions. We attribute this effect to (i) a loss of information caused by sparsity post-processing (due to the deletion of small loadings), and (ii) numerical issues encountered when implementing the rotation method in the FactorAnalyzer Python library \citep{factoranalyzer}. Therefore, although these methods provide valuable insights for interpreting the outcome data, they can also decrease the variance explained as the number of factors grows.

\subsection{Frontier Estimation} \label{ec:frontier}
A central point in the frontier estimation was the sensitivity analysis conducted, performed with the variation across the following variables:

\paragraph{Input variables:} 
For the HealthBench dataset, we explored different configurations of inputs for technical data and cost variables. Specifically, we evaluated with a full set of inputs (considering general benchmarks and technical data) and with a set of reduced inputs (considering just the general bechmarks), and we used either the number of parameters or the first answer token time as the cost variable. For the PRISM dataset, no variation was applied to the input variables.

\paragraph{Number of factors:} 
We varied the number of factors extracted from the internal metric dataset. For HealthBench, we considered 2, 3, and 4 factors, whereas for PRISM we used 2 and 3.

\paragraph{Number of tiers:} 
We varied the number of tiers used for the clustering step in frontier construction. The number of tiers ranged from 2 to 5 for both datasets, although larger numbers of tiers were also considered for HealthBench due to its larger model population.

\paragraph{Tolerance:} 
This parameter controls the error allowed when determining whether a model is considered Pareto dominant. As the tolerance increases, the acceptance criterion for the optimal frontier becomes more relaxed. We considered the following range of values: $\{0, 0.01, 0.025, 0.05, 0.075, 0.10\}$.

\paragraph{Cost aggregation method:} 
This parameter determines the aggregation method used within each tier for the estimation of tier-level costs. We considered the arithmetic mean, geometric mean, and median.

The following analysis is based on the variation of the above stated variables for both HealthBench and PRISM. We conducted the analysis and extract values for the metrics that define the quality and rigor of our approach, as well as the robustness of our findings. We compared the value of the $b$ and $d$ parameters stated at assumption \ref{assump:frontier}, the frontier normalized fitting error (the error the frontier with respect to the models that shape the frontier, normalised with the number of models that shape that frontier) and the Pearson and Spearman correlation coefficient to compare the proposed cost for each frontier with the real costs of each model. Results can be found in Tables \ref{tab:sensitivity_analysis_prism} and \ref{tab:sensitivity_analysis_hb}.

\begin{table}[h]
\caption{Sensitivity analysis for PRISM dataset. Outcome-frontier values are expressed as mean ± standard deviation. \textit{G. mean} and \textit{A. mean }cells represent geometric mean and arithmetic mean respectively.}
\hspace{-0.75cm}
\begin{tabular}{llccccc}
\hline
Variable                                                                             & Value      & $b$ parameter & $d$ parameter & Fitting error     & Pearson       & Spearman      \\ \hline
\multirow{2}{*}{\begin{tabular}[c]{@{}l@{}}Number\\ of factors\end{tabular}}         & 2          & 1.508 ± 0.913 & 0.755 ± 1.076 & 0.00254 ± 0.00062 & 0.262 ± 0.101 & 0.224 ± 0.088 \\
                                                                                     & 3          & 1.625 ± 1.179 & 0.710 ± 1.240 & 0.00212 ± 0.00064 & 0.368 ± 0.031 & 0.218 ± 0.044 \\ \hline
\multirow{4}{*}{Tiers}                                                               & 2          & 1.661 ± 1.654 & 1.332 ± 1.411 & 0.00293 ± 0.00043 & 0.229 ± 0.134 & 0.147 ± 0.038 \\
                                                                                     & 3          & 1.371 ± 0.990 & 0.994 ± 1.608 & 0.00268 ± 0.00064 & 0.324 ± 0.036 & 0.223 ± 0.087 \\
                                                                                     & 4          & 1.909 ± 0.290 & 0.199 ± 0.094 & 0.00181 ± 0.00047 & 0.359 ± 0.034 & 0.270 ± 0.028 \\
                                                                                     & 5          & 1.327 ± 0.703 & 0.407 ± 0.205 & 0.00191 ± 0.00013 & 0.348 ± 0.055 & 0.245 ± 0.038 \\ \hline
\multirow{3}{*}{\begin{tabular}[c]{@{}l@{}}Cost\\ aggregation\\ method\end{tabular}} & G. mean & 1.567 ± 1.059 & 0.344 ± 0.212 & 0.00233 ± 0.00067 & 0.314 ± 0.088 & 0.221 ± 0.070 \\
                                                                                     & A. mean       & 1.567 ± 1.059 & 0.334 ± 0.306 & 0.00233 ± 0.00067 & 0.328 ± 0.097 & 0.221 ± 0.070 \\
                                                                                     & Median     & 1.567 ± 1.059 & 1.521 ± 1.728 & 0.00233 ± 0.00067 & 0.302 ± 0.089 & 0.221 ± 0.070 \\ \hline
\multirow{6}{*}{Tolerance}                                                           & 0.000      & 1.599 ± 1.060 & 0.688 ± 1.190 & 0.00246 ± 0.00073 & 0.321 ± 0.095 & 0.247 ± 0.078 \\
                                                                                     & 0.010      & 1.599 ± 1.060 & 0.688 ± 1.190 & 0.00246 ± 0.00073 & 0.321 ± 0.095 & 0.247 ± 0.078 \\
                                                                                     & 0.025      & 1.599 ± 1.060 & 0.688 ± 1.190 & 0.00246 ± 0.00073 & 0.321 ± 0.095 & 0.247 ± 0.078 \\
                                                                                     & 0.050      & 1.298 ± 0.886 & 0.774 ± 1.155 & 0.00218 ± 0.00055 & 0.318 ± 0.094 & 0.204 ± 0.053 \\
                                                                                     & 0.075      & 1.298 ± 0.886 & 0.774 ± 1.155 & 0.00218 ± 0.00055 & 0.318 ± 0.094 & 0.204 ± 0.053 \\
                                                                                     & 0.100      & 2.006 ± 1.257 & 0.783 ± 1.184 & 0.00226 ± 0.00063 & 0.290 ± 0.078 & 0.180 ± 0.041 \\ \hline
\end{tabular}
\label{tab:sensitivity_analysis_prism}
\end{table}

\begin{table}[h]
\caption{Sensitivity analysis for the HealthBench dataset. Outcome-frontier values are expressed as mean ± standard deviation. As in PRISM table, \textit{G. mean} and \textit{A. mean} cells represent geometric mean and arithmetic mean respectively. Input-set legend: Set 1 = parameter count as cost; benchmark scores + speed metrics as capability. Set 2 = parameter count as cost; benchmark scores as capability. Set 3 = latency as cost; benchmark scores + speed metrics as capability. Set 4 = latency as cost; benchmark scores as capability.}
\hspace{-0.7cm}
\begin{tabular}{llccccc}
\hline
Variable                                                                             & Value   & $b$ parameter & $d$ parameter & Fitting error     & Pearson       & Spearman      \\ \hline
\multirow{3}{*}{\begin{tabular}[c]{@{}l@{}}Number\\ of factors\end{tabular}}         & 2       & 0.594 ± 0.358 & 0.953 ± 1.465 & 0.00217 ± 0.00134 & 0.326 ± 0.212 & 0.411 ± 0.136 \\
                                                                                     & 3       & 0.552 ± 0.322 & 0.797 ± 0.856 & 0.00299 ± 0.00126 & 0.300 ± 0.174 & 0.399 ± 0.120 \\
                                                                                     & 4       & 1.127 ± 0.309 & 0.971 ± 1.716 & 0.00215 ± 0.00155 & 0.348 ± 0.253 & 0.433 ± 0.150 \\ \hline
\multirow{5}{*}{\begin{tabular}[c]{@{}l@{}}Number\\ of tiers\end{tabular}}           & 2       & 0.493 ± 0.482 & 1.396 ± 2.211 & 0.00405 ± 0.00193 & 0.253 ± 0.223 & 0.368 ± 0.144 \\
                                                                                     & 3       & 0.820 ± 0.424 & 1.144 ± 1.967 & 0.00271 ± 0.00108 & 0.305 ± 0.209 & 0.410 ± 0.135 \\
                                                                                     & 4       & 0.722 ± 0.345 & 0.674 ± 0.450 & 0.00219 ± 0.00080 & 0.348 ± 0.218 & 0.426 ± 0.126 \\
                                                                                     & 5       & 0.855 ± 0.377 & 0.612 ± 0.404 & 0.00177 ± 0.00068 & 0.350 ± 0.210 & 0.428 ± 0.134 \\
                                                                                     & Higher  & 0.900 ± 0.335 & 0.708 ± 0.394 & 0.00146 ± 0.00054 & 0.369 ± 0.203 & 0.440 ± 0.133 \\ \hline
\multirow{3}{*}{\begin{tabular}[c]{@{}l@{}}Cost\\ aggregation\\ method\end{tabular}} & G. mean & 0.758 ± 0.422 & 0.477 ± 0.274 & 0.00244 ± 0.00144 & 0.309 ± 0.224 & 0.414 ± 0.137 \\
                                                                                     & A. mean & 0.758 ± 0.422 & 1.650 ± 2.196 & 0.00244 ± 0.00144 & 0.355 ± 0.194 & 0.414 ± 0.137 \\
                                                                                     & Median  & 0.758 ± 0.422 & 0.593 ± 0.331 & 0.00244 ± 0.00144 & 0.311 ± 0.227 & 0.414 ± 0.137 \\ \hline
\multirow{6}{*}{Tolerance}                                                           & 0.000   & 0.709 ± 0.442 & 0.881 ± 1.314 & 0.00244 ± 0.00144 & 0.331 ± 0.224 & 0.415 ± 0.143 \\
                                                                                     & 0.010   & 0.719 ± 0.446 & 0.928 ± 1.479 & 0.00245 ± 0.00143 & 0.331 ± 0.226 & 0.417 ± 0.143 \\
                                                                                     & 0.025   & 0.763 ± 0.403 & 0.927 ± 1.487 & 0.00243 ± 0.00140 & 0.321 ± 0.217 & 0.409 ± 0.137 \\
                                                                                     & 0.050   & 0.756 ± 0.401 & 0.947 ± 1.507 & 0.00240 ± 0.00139 & 0.321 ± 0.218 & 0.409 ± 0.137 \\
                                                                                     & 0.075   & 0.791 ± 0.419 & 0.855 ± 1.200 & 0.00244 ± 0.00162 & 0.327 ± 0.209 & 0.415 ± 0.132 \\
                                                                                     & 0.100   & 0.810 ± 0.411 & 0.904 ± 1.369 & 0.00245 ± 0.00138 & 0.319 ± 0.205 & 0.422 ± 0.129 \\ \hline
\multirow{4}{*}{Input set}                                                           & 1       & 0.993 ± 0.366 & 0.929 ± 0.440 & 0.00166 ± 0.00110 & 0.454 ± 0.081 & 0.381 ± 0.060 \\
                                                                                     & 2       & 0.993 ± 0.366 & 1.473 ± 2.548 & 0.00166 ± 0.00110 & 0.045 ± 0.122 & 0.257 ± 0.058 \\
                                                                                     & 3       & 0.522 ± 0.332 & 0.616 ± 0.281 & 0.00321 ± 0.00133 & 0.546 ± 0.090 & 0.577 ± 0.067 \\
                                                                                     & 4       & 0.522 ± 0.332 & 0.610 ± 0.734 & 0.00321 ± 0.00133 & 0.253 ± 0.092 & 0.443 ± 0.101 \\ \hline
\end{tabular}
\label{tab:sensitivity_analysis_hb}
\end{table}

The sensitivity analysis yielded several important insights regarding the factors that influence frontier estimation. Given tables \ref{tab:sensitivity_analysis_prism} and \ref{tab:sensitivity_analysis_hb} the main observations, for both datasets PRISM and HealtBench, are as follows:

\begin{itemize}
    \item In both case studies, to a greater or lesser extent, increasing the number of factors or the number of tiers leads to an increase in $b$. This is because either there are more dimensions available to substitute models, or more similar models are grouped together, resulting in higher substitutability.
    
    \item Similarly, for both case studies, increasing the number of factors—or, more notably, the number of tiers—leads to an decrease in $d$. In other words, the diminishing returns of additional investments become less pronounced as these parameters grow.
    
    \item Variations in the remaining parameters generally do not have a significant impact on frontier estimation, and no clear trends could be identified. The fitting error, the pearson correlation coefficient and the spearman correlation coefficient remain stable between all the configurations, although there is a subtle decrease in the first one and a slight overall increase the last two coefficients. 
    
    \item Overall, the purpose of this analysis is to identify appropriate configurations. It is important to emphasize that the optimal settings are data and application dependent. For instance, they may depend on the number of input variables considered (balancing interpretability and factor analysis fidelity), the desired price granularity, and other context-specific considerations. The characteristics of the sample dataset are also crucial when selecting parameters such as the number of tiers or factors, since small sample sizes can lead to problematic situations if these parameters are increased.
\end{itemize}

\begin{figure}[!htb]
   \begin{minipage}{0.49\textwidth}
     \centering
     \includegraphics[width=\linewidth]{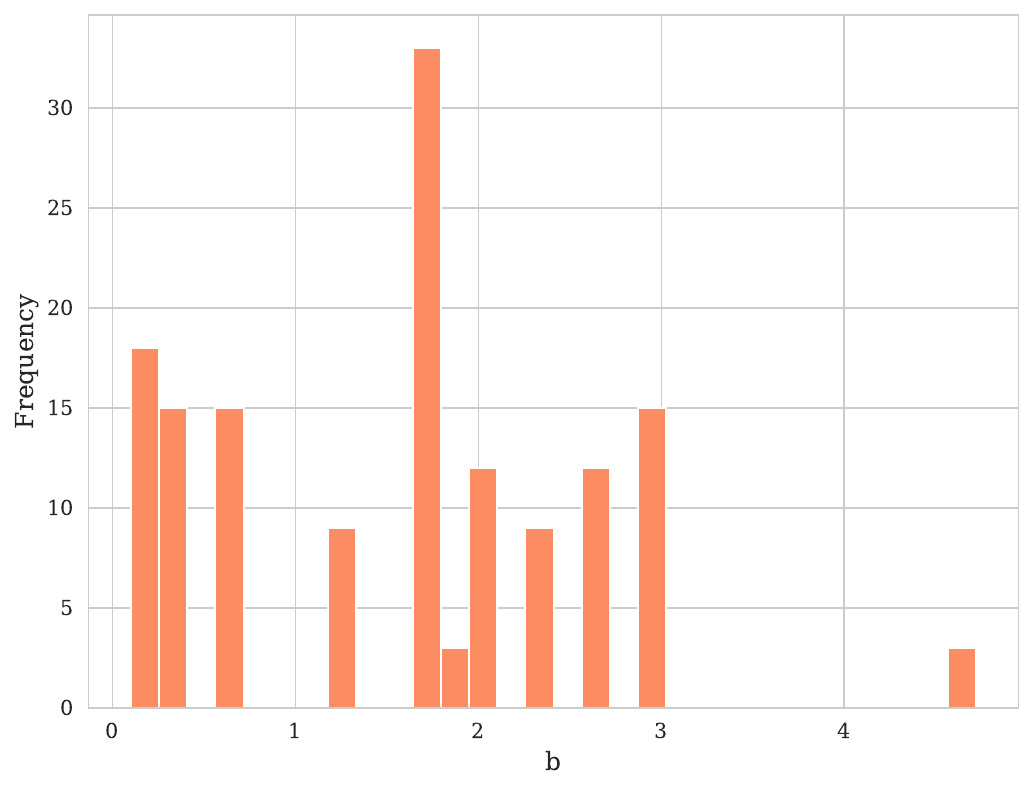}
   \end{minipage}\hfill
   \begin{minipage}{0.49\textwidth}
     \centering
     \includegraphics[width=\linewidth]{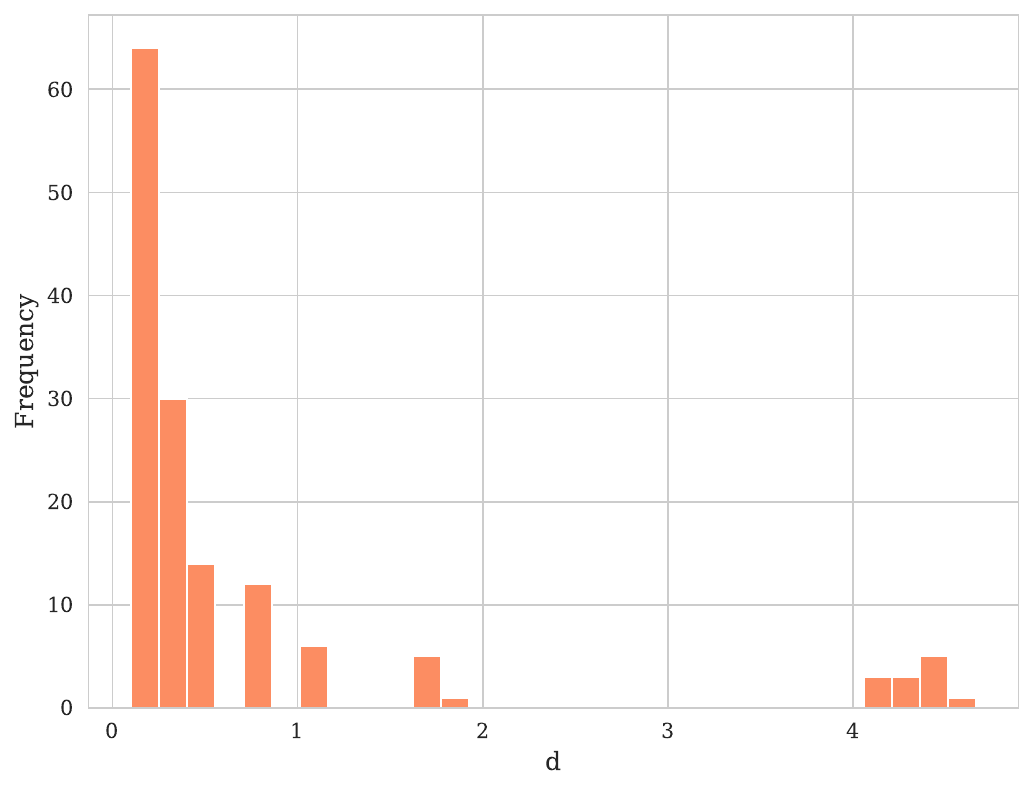}
   \end{minipage}
   \caption{Histogram showing the empirical distribution of parameter $b$ and $d$ across all configurations used in the sensitivity analysis for PRISM.}\label{Fig:histPRISM}
\end{figure}

Another important aspect we examined during the sensitivity analysis was the scope and applicability of our assumptions \ref{assump:frontier} related to frontier attainability. Specifically, the results are shown in Figures~\ref{Fig:histPRISM} and~\ref{Fig:histdHB}, and the main purpose was to assess the extent to which the $b$ and $d$ values comply with the previously stated constraints.

For the PRISM case, as Figure~\ref{Fig:histPRISM} shows, both assumptions remain sensible for the vast majority of cases. Furthermore, Table~\ref{tab:sensitivity_analysis_prism} indicates that unfeasible values of $b$—with respect to the MLC formulation—rarely appear as outliers across almost all parameter configurations, as can be observed from the variance shifts. We attribute this to poorly behaving specific configurations due to the small sample size of PRISM.

Conversely, for the $d$ parameter, some configurations can indeed challenge the reliability of our assumption. Specifically, using the median aggregation method for costs and employing a low number of tiers tends to produce uninformative values of $d$ within our problem formulation. Regarding the first case, our explanation is that the median can distort the estimated $d$ value in PRISM because, in this approximation, the produced tier costs tend to cluster around extreme low or high values. Concerning the second case, our reasoning is that with a very low number of tiers, clustering can sometimes average models with widely different costs and performances, making small resource increases appear disproportionately effective, which can push the estimated $d$ above the value of 1.

\begin{figure}[!htb]
   \begin{minipage}{0.49\textwidth}
     \centering
     \includegraphics[width=\linewidth]{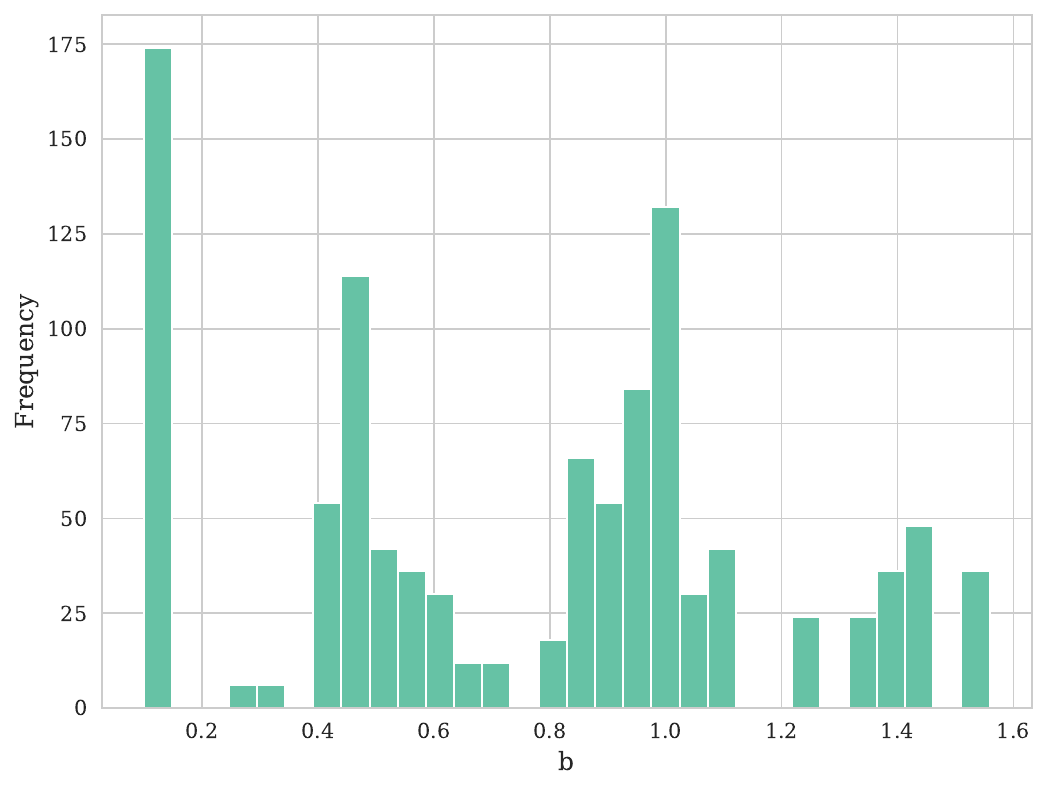}
   \end{minipage}\hfill
   \begin{minipage}{0.49\textwidth}
     \centering
     \includegraphics[width=\linewidth]{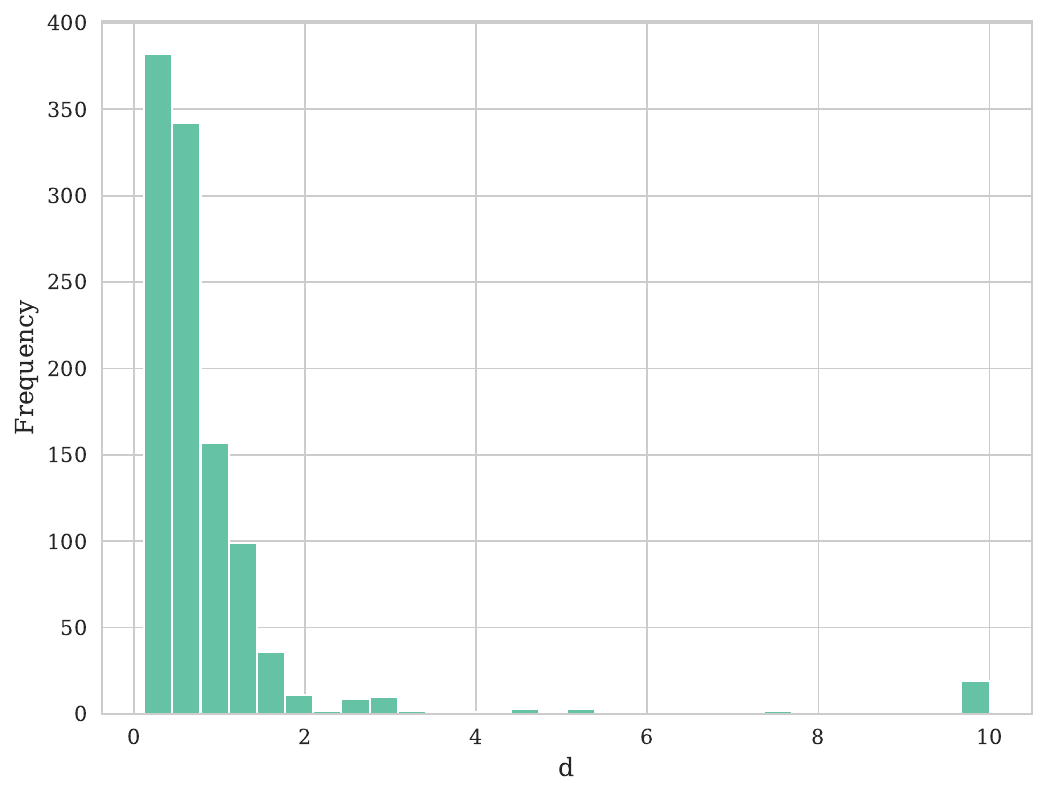}
   \end{minipage}
   \caption{Histogram showing the empirical distribution of parameter $b$ and $d$ across all configurations used in the sensitivity analysis for HealthBench.}\label{Fig:histdHB}
\end{figure}

For the HealthBench dataset, the situation is more complex. The assumption made for the $d$ parameter remains feasible, whereas for $b$ it becomes less realistic compared to PRISM in many cases. Regarding the $d$ parameter, the assumption appears sensible, and Table~\ref{tab:sensitivity_analysis_hb} highlights lower numbers of tiers and the use of the arithmetic mean in the cost aggregation method as factors contributing to this situation. In addition, the specific input set 2 (using first answer token time as the cost and the full set of capability data, including benchmarks and technical specifications) also plays a role. The behavior of $d$ with respect to the number of tiers is consistent with the explanation observed for PRISM, which reinforces our reasoning. Concerning the arithmetic mean, the likely explanation is that extreme values within clusters lead to less sensible cost estimates when using the arithmetic mean. The effect of input set 2 remains plausible, as higher first answer token times (indicating more reasoning time) could also be correlated with higher processing speed, since the answer may have been partially preprocessed.

Finally, the assumption regarding the $b$ parameter is more challenging for HealthBench. In general, input sets 2 and 3, which rely on capability data without technical data, consistently yield uninformative $b$ values. This can be attributed to the decrease in substitutability when technical data is omitted: relying solely on benchmark capabilities results in models that are less substitutable and more homogeneously distributed (excelling in one benchmark often implies performing well across others).

To conclude this section, we complement the HealthBench visualizations presented in Section~\ref{ssec:case-frontier} with two additional frontier views focusing on the third component used in the analysis. The corresponding results are shown in Figure \ref{Fig:histdHB}.

\begin{figure}[!htb]
   \begin{minipage}{0.49\textwidth}
     \centering
     \includegraphics[width=\linewidth]{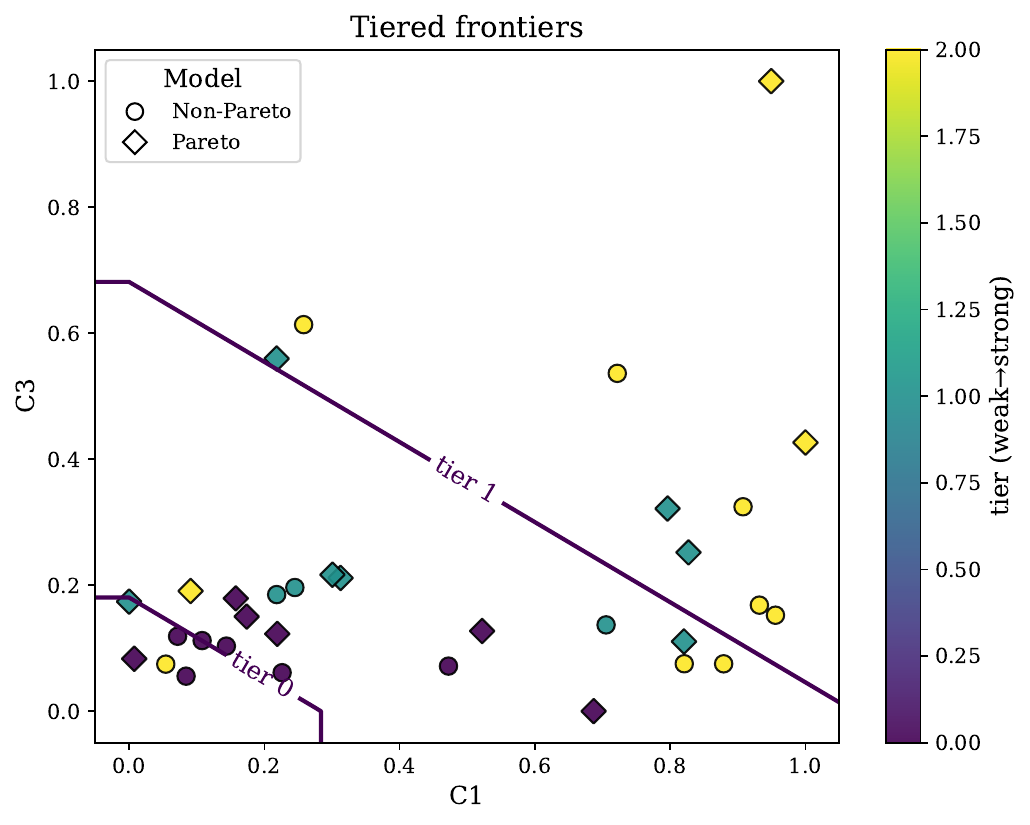}
   \end{minipage}\hfill
   \begin{minipage}{0.49\textwidth}
     \centering
     \includegraphics[width=\linewidth]{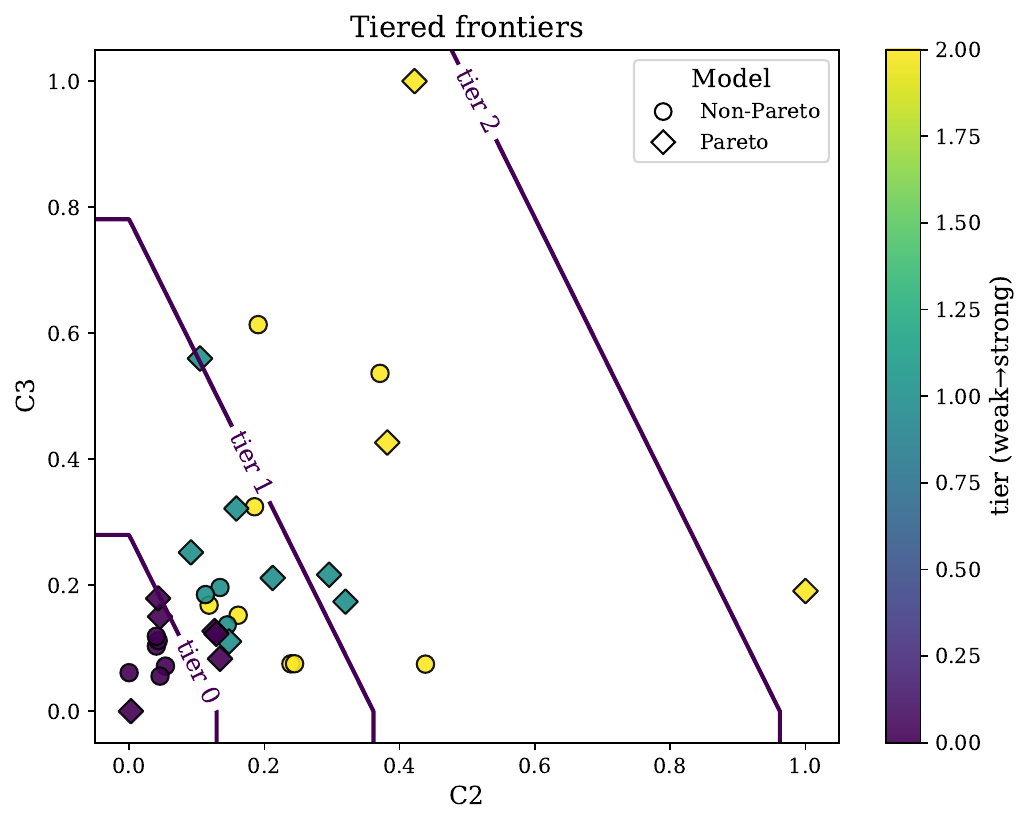}
   \end{minipage}
   \caption{Two additional frontier visualization perspectives for HealthBench case given the frontier show in section \ref{ssec:case-frontier}.}\label{Fig:histdHB}
\end{figure}

\subsection{Utility Estimation} \label{ec:utility}
In this section, we present the full utility estimation results for both PRISM and HealthBench (in the main paper, we only showed a subset of HB task types), as well as additional utility estimation approaches. We analyze how the utility estimation changes as the contextual information used for estimation is progressively enriched. To this end, we incorporate additional features beyond the internal model metrics, capturing aspects related to the specific conversation and its context (e.g., user characteristics and task-related information). Table \ref{tab:utility_auc_no_z_type} shows the performance of utility estimators without $z$ type considerations for PRISM and HealthBench.

\begin{table}[h]
\centering
\caption{Out-of-sample performance of utility estimators without $z$ type considerations for PRISM and HealthBench. The columns report the number of training and test interactions, as well as the AUC on the training and test sets. }
\label{tab:utility_auc_no_z_type}
\begin{tabular}{lcccc}
\hline
Dataset     & $|\mathcal{I}_{\text{train}}|$ & $|\mathcal{I}_{\text{test}}|$ & Train AUC                         & \multicolumn{1}{c}{Test AUC}     \\ \hline
PRISM       & 23,342                         & 5,835                         & \multicolumn{1}{r}{0.639 (0.002)} & \multicolumn{1}{r}{0.639 (0.008)} \\ \hline
HealthBench & 96,914                         & 24,228                        & 0.597 (0.000)                     & 0.596 (0.000)                     \\ \hline
\end{tabular}
\end{table}

For the PRISM dataset, we include user-level contextual features, namely the user’s age, level of education, familiarity with large language models, and frequency of LLM usage. In addition, we consider a set of indicators capturing the primary user use case of LLM, including research, professional work, creative writing, technical or programming assistance, and lifestyle or hobby-related tasks. Finally, we incorporate the user preference-related variables developed during the user type creation, namely \textit{weighted\_c1}, \textit{weighted\_c2}, and their difference, \textit{c1\_c2\_diff}.

For HealthBench, due to the limited availability of contextual variables related to users or tasks, we construct a set of alternative exogenous features to support the estimation process. Specifically, we consider the length of the initial conversation, which may indirectly influence the LLM response by providing additional contextual information about the task. In addition, we incorporate the first 128 principal components of a text embedding computed from the conversation. The embedding is obtained using the OpenAI API with the \texttt{text-embedding-3-large} model. We then apply principal component analysis (PCA) to retain the components explaining the largest share of variance, which are intended to capture the dominant semantic characteristics of the conversation and therefore, the underlying task intent.

Table \ref{tab:utility_auc_no_z_type_extra_features} shows the performance of utility estimators without $z$ type considerations for PRISM and HealthBench but with extra features.

\begin{table}[h]
\centering
\caption{Out-of-sample performance of utility estimators without $z$ type considerations for PRISM and HealthBench, but including extra-features about users, and tasks. The columns report the number of training and test interactions, as well as the AUC on the training and test sets.}
\label{tab:utility_auc_no_z_type_extra_features}
\begin{tabular}{lccrr}
\hline
Dataset     & $|\mathcal{I}_{\text{train}}|$ & $|\mathcal{I}_{\text{test}}|$ & \multicolumn{1}{c}{Train AUC} & \multicolumn{1}{c}{Test AUC} \\ \hline
PRISM       & 23,342                         & 5,835                         & 0.714 (0.002)                 & 0.714 (0.008)                 \\ \hline
HealthBench & 96,914                         & 24,228                        & 0.914 (0.001)                 & 0.678 (0.007)                 \\ \hline
\end{tabular}
\end{table}

Table~\ref{tab:utility_auc_archetype} presents the utility estimation results for both datasets, grouped according to the user type classification introduced earlier.

\begin{table}[h]
\centering
\caption{Out-of-sample performance of utility estimators by user type for PRISM and HealthBench, respectively. The columns report the user type, the number of training and test interactions, as well as the AUC on the training and test sets.}
\label{tab:utility_auc_archetype}
\begin{tabular}{clcccc}
\hline
\multicolumn{1}{l}{Dataset}  & User type          & \multicolumn{1}{l}{$|\mathcal{I}_{\text{train}}|$} & \multicolumn{1}{l}{$|\mathcal{I}_{\text{test}}|$} & \multicolumn{1}{l}{Train AUC} & \multicolumn{1}{l}{Test AUC} \\ \hline
\multirow{4}{*}{PRISM}       & Ethics-focused          & 11,688                                             & 2,922                                             & 0.668 (0.003)                 & 0.668 (0.014)                \\
                             & Safety-focused          & 1,764                                              & 441                                               & 0.687 (0.006)                 & 0.687 (0.027)                \\
                             & General                 & 9,889                                              & 2,472                                             & 0.649 (0.003)                 & 0.649 (0.013)                \\
                             & MEAN                    & 7,780                                              & 1,945                                             & 0.668 (0.016)                 & 0.668 (0.016)                \\ \hline
\multirow{4}{*}{HealthBench} & General                 & 56,865                                             & 14,216                                            & 0.589 (0.001)                 & 0.588 (0.004)                \\
                             & General knowledgeable   & 8,236                                              & 2,059                                             & 0.577 (0.003)                 & 0.572 (0.011)                \\
                             & Healthcare professional & 31,812                                             & 7,953                                             & 0.628 (0.001)                 & 0.627 (0.006)                \\
                             & MEAN                    & 32,304                                             & 8,076                                             & 0.598 (0.022)                 & 0.596 (0.023)                \\ \hline
\end{tabular}
\end{table}

Table~\ref{tab:utility_auc_tasks} presents the utility estimation results for both datasets, grouped according to the task topic classification introduced earlier.

\begin{table}[h]
\centering
\caption{Out-of-sample performance of utility estimators by task type for PRISM and HealthBench. The columns report the task topic, the number of training and test interactions, as well as the AUC on the training and test sets. }
\label{tab:utility_auc_tasks}
\begin{tabular}{llcccc}
\hline
Dataset                      & Task Topic & $|\mathcal{I}_{\text{train}}|$ & $|\mathcal{I}_{\text{test}}|$ & Train AUC     & Test AUC      \\ \hline
\multirow{4}{*}{PRISM}       & Everyday Use     & 7,512                          & 1,878                         & 0.642 (0.002) & 0.643 (0.009) \\
                             & Reflective Use     & 10,640                         & 2,660                         & 0.637 (0.003) & 0.636 (0.012) \\
                             & Controversial Discourse            & 5,188                          & 1,297                         & 0.645 (0.003) & 0.645 (0.011) \\
                             & MEAN                    & 7,780                          & 1,945                         & 0.641 (0.003) & 0.641 (0.004) \\ \hline
\multirow{9}{*}{HealthBench} & Immediate Emergency     & 3,669                          & 917                           & 0.697 (0.005) & 0.693 (0.022) \\
                             & Conditional Emergency   & 4,752                          & 1,188                         & 0.749 (0.002) & 0.742 (0.008) \\
                             & No Emergency            & 3,537                          & 884                           & 0.735 (0.003) & 0.709 (0.016) \\
                             & Health Professional     & 8,500                          & 2,125                         & 0.805 (0.006) & 0.801 (0.022) \\
                             & Non-Health Professional & 11,193                         & 2,798                         & 0.687 (0.002) & 0.682 (0.010) \\
                             & Reduce Uncertainty      & 7,524                          & 1,881                         & 0.804 (0.003) & 0.797 (0.016) \\
                             & Deal Uncertainty        & 5,227                          & 1,306                         & 0.668 (0.009) & 0.656 (0.038) \\
                             & No Uncertainty          & 6,019                          & 1,504                         & 0.702 (0.007) & 0.690 (0.026) \\
                             & Simple Answer           & 4,303                          & 1,075                         & 0.648 (0.003) & 0.640 (0.014) \\
                             & Detailed Answer         & 4,250                          & 1,062                         & 0.719 (0.002) & 0.713 (0.008) \\
                             & Task Can Be Done        & 5,676                          & 1,419                         & 0.669 (0.005) & 0.661 (0.021) \\
                             & Task Needs More Info    & 4,752                          & 1,188                         & 0.717 (0.006) & 0.701 (0.020) \\
                             & Context Needed Known    & 5,148                          & 1,287                         & 0.713 (0.006) & 0.695 (0.026) \\
                             & Context Needed Unknown  & 5,781                          & 1,445                         & 0.656 (0.003) & 0.649 (0.011) \\
                             & Context Not Needed      & 5,808                          & 1,452                         & 0.705 (0.007) & 0.695 (0.032) \\
                             & Context Sufficient      & 5,992                          & 1,498                         & 0.784 (0.009) & 0.773 (0.027) \\
                             & Context Missing         & 4,778                          & 1,194                         & 0.789 (0.011) & 0.782 (0.042) \\
                             & MEAN                    & 5,700                          & 1,424                         & 0.720 (0.049) & 0.711 (0.050) \\ \hline
\end{tabular}
\end{table}

Table \ref{tab:utiltiy_auc_totaltable} presents the results for the two study cases selected for PRISM and HealthBench, corresponding to user types and task topics, respectively. We report AUC values for both the linear and non-linear estimators. 

\begin{table}[]
\centering
\caption{Out-of-sample performance of utility estimators by user type and task type for PRISM and HealthBench, respectively. The columns report the $z$ type (i.e. user or task), the number of training and test interactions, as well as the AUC on the training and test sets. The last three columns correspond to the weights of the two (or three) internal measures in the utility function classifier. For PRISM, the values are directly extracted from the logistic regressor coefficients and normalized, whereas for HealthBench, the values are computed as the feature importance gains of each factor, normalized by the total sum of importance gains between the three factors in the classifier.}
\label{tab:utiltiy_auc_totaltable}
\resizebox{0.8\textwidth}{!}{
\begin{tabular}{llccllll}
\hline
\multirow{2}{*}{Dataset}      & \multirow{2}{*}{$z$ type} & \multirow{2}{*}{$|\mathcal{I}_{\text{train}}|$} & \multirow{2}{*}{$|\mathcal{I}_{\text{test}}|$} & \multicolumn{2}{c}{Train AUC}                                         & \multicolumn{2}{c}{Test AUC}                                          \\
                              &                           &                                                 &                                                & \multicolumn{1}{c}{Linear}        & \multicolumn{1}{c}{Non-linear}    & \multicolumn{1}{c}{Linear}        & \multicolumn{1}{c}{Non-linear}    \\ \hline
\multirow{4}{*}{PRISM}        & Ethics-focused User       & 11,688                                          & 2,922                                          & \multicolumn{1}{c}{0.668 (0.003)} & \multicolumn{1}{c}{0.682 (0.003)} & \multicolumn{1}{c}{0.668 (0.014)} & \multicolumn{1}{c}{0.679 (0.011)} \\
                              & Safety-focused User       & 1,764                                           & 441                                            & \multicolumn{1}{c}{0.687 (0.006)} & \multicolumn{1}{c}{0.708 (0.005)} & \multicolumn{1}{c}{0.687 (0.027)} & \multicolumn{1}{c}{0.696 (0.019)} \\
                              & General User              & 9,889                                           & 2,472                                          & \multicolumn{1}{c}{0.649 (0.003)} & \multicolumn{1}{c}{0.675 (0.002)} & \multicolumn{1}{c}{0.649 (0.013)} & \multicolumn{1}{c}{0.672 (0.009)} \\
                              & MEAN                      & 7,780                                           & 1,945                                          & \multicolumn{1}{c}{0.668 (0.016)} & \multicolumn{1}{c}{0.688 (0.014)} & \multicolumn{1}{c}{0.668 (0.016)} & \multicolumn{1}{c}{0.682 (0.010)} \\ \hline
\multirow{18}{*}{HealthBench} & Immediate Emergency       & 3,669                                           & 917                                            & 0.641 (0.005)                     & 0.697 (0.005)                     & 0.643 (0.020)                     & 0.693 (0.022)                     \\
                              & Conditional Emergency     & 4,752                                           & 1,188                                          & 0.670 (0.003)                     & 0.749 (0.002)                     & 0.670 (0.012)                     & 0.742 (0.008)                     \\
                              & No Emergency              & 3,537                                           & 884                                            & 0.622 (0.006)                     & 0.735 (0.003)                     & 0.621 (0.030)                     & 0.709 (0.016)                     \\
                              & Health Professional       & 8,500                                           & 2,125                                          & 0.754 (0.006)                     & 0.805 (0.006)                     & 0.754 (0.025)                     & 0.801 (0.022)                     \\
                              & Non-Health Professional   & 11,193                                          & 2,798                                          & 0.556 (0.003)                     & 0.687 (0.002)                     & 0.556 (0.012)                     & 0.682 (0.010)                     \\
                              & Reduce Uncertainty        & 7,524                                           & 1,881                                          & 0.664 (0.009)                     & 0.804 (0.003)                     & 0.664 (0.017)                     & 0.797 (0.016)                     \\
                              & Deal Uncertainty          & 5,227                                           & 1,306                                          & 0.563 (0.004)                     & 0.668 (0.009)                     & 0.563 (0.018)                     & 0.656 (0.038)                     \\
                              & No Uncertainty            & 6,019                                           & 1,504                                          & 0.630 (0.005)                     & 0.702 (0.007)                     & 0.628 (0.013)                     & 0.690 (0.026)                     \\
                              & Simple Answer             & 4,303                                           & 1,075                                          & 0.559 (0.003)                     & 0.648 (0.003)                     & 0.556 (0.011)                     & 0.640 (0.014)                     \\
                              & Detailed Answer           & 4,250                                           & 1,062                                          & 0.664 (0.004)                     & 0.719 (0.002)                     & 0.664 (0.018)                     & 0.713 (0.008)                     \\
                              & Task Can Be Done          & 5,676                                           & 1,419                                          & 0.635 (0.004)                     & 0.669 (0.005)                     & 0.637 (0.016)                     & 0.661 (0.021)                     \\
                              & Task Needs More Info      & 4,752                                           & 1,188                                          & 0.582 (0.010)                     & 0.717 (0.006)                     & 0.585 (0.039)                     & 0.701 (0.020)                     \\
                              & Context Needed Known      & 5,148                                           & 1,287                                          & 0.642 (0.008)                     & 0.713 (0.006)                     & 0.632 (0.043)                     & 0.695 (0.026)                     \\
                              & Context Needed Unknown    & 5,781                                           & 1,445                                          & 0.611 (0.005)                     & 0.656 (0.003)                     & 0.611 (0.018)                     & 0.649 (0.011)                     \\
                              & Context Not Needed        & 5,808                                           & 1,452                                          & 0.639 (0.004)                     & 0.705 (0.007)                     & 0.642 (0.020)                     & 0.695 (0.032)                     \\
                              & Context Sufficient        & 5,992                                           & 1,498                                          & 0.693 (0.007)                     & 0.784 (0.009)                     & 0.688 (0.015)                     & 0.773 (0.027)                     \\
                              & Context Missing           & 4,778                                           & 1,194                                          & 0.649 (0.011)                     & 0.789 (0.011)                     & 0.649 (0.022)                     & 0.782 (0.042)                     \\
                              & MEAN                      & 5,700                                           & 1,424                                          & 0.634 (0.049)                     & 0.720 (0.049)                     & 0.633 (0.049)                     & 0.711 (0.050)                     \\ \hline
\end{tabular}}
\end{table}

Finally, Table \ref{tab:utility_auc} presents the full results for the main line of analysis we did in the paper, using the linear estimator for PRISM and the non-linear estimator for HealthBench.

\begin{table}[t]
\centering
\caption{Out-of-sample performance of utility estimators by user type and task type for PRISM and HealthBench, respectively. The columns report the $z$ type (i.e. user or task), the number of training and test interactions, as well as the AUC on the training and test sets. The last three columns correspond to the weights of the two (or three) internal measures in the utility function classifier. For PRISM, the values are directly extracted from the logistic regressor coefficients and normalized, whereas for HealthBench, the values are computed as the feature importance gains of each factor, normalized by the total sum of importance gains between the three factors in the classifier.}
\label{tab:utility_auc}
\resizebox{0.8\textwidth}{!}{
\begin{tabular}{llccccccc}
\toprule
Dataset & $z$ type & $|\mathcal{I}_{\text{train}}|$ & $|\mathcal{I}_{\text{test}}|$ & Train AUC & Test AUC & $\hat\beta_1$ & $\hat\beta_2$ & $\hat\beta_3$\\
\midrule
\multirow{5}{*}{PRISM} 
  & Ethics-focused User & 11,688&	2,922&	0.668 (0.003) &	0.668 (0.014)	&	1.000&	0.000 & \textemdash\\
  & Safety-focused User & 1,764&	441&	0.687 (0.006)	&	0.687 (0.027)	&	0.017&	0.983 & \textemdash\\
  & General User & 9,889&	2,472&	0.649 (0.003)&	0.649 (0.013)	&	0.579 	&	0.421 & \textemdash\\
  & MEAN        & 7,780&	1,945&	0.668 (0.016)	&	0.668 (0.016)	& \textemdash & \textemdash & \textemdash\\
\midrule
\multirow{19}{*}{HealthBench}

  & Immediate Emergency               & 3,669&	917&	0.697 (0.005)&	0.693 (0.022)&	0.810&	0.034&	0.156 \\
  & Conditional Emergency             & 4,752&	1,188&	0.749 (0.002)&	0.742 (0.008)&	0.603&	0.195&	0.202\\
  & No Emergency                      & 3,537&	884&	0.735 (0.003)&	0.709 (0.016)&	0.340&	0.576&	0.085\\
  & Health Professional               & 8,500&	2,125&	0.805 (0.006)&	0.801 (0.022)&	0.848&	0.091&	0.061\\
  & Non-Health Professional           & 11,193&	2,798&	0.687 (0.002)&	0.682 (0.010)&	0.295&	0.380&	0.325\\
  & Reduce Uncertainty          & 7,524&	1,881&	0.804 (0.003)&	0.797 (0.016)&	0.635&	0.108&	0.257\\
  & Deal Uncertainty      & 5,227&	1,306&	0.668 (0.009)&	0.656 (0.038)&	0.092&	0.101&	0.806\\
  & No Uncertainty                    & 6,019&	1,504&	0.702 (0.007)&	0.690 (0.026)&	0.798&	0.156&	0.046 \\
  & Simple Answer                            &	4,303&	1,075&	0.648 (0.003)&	0.640 (0.014)&	0.155&	0.677&	0.168  \\
  & Detailed Answer                          & 4,250&	1,062&	0.719 (0.002)&	0.713 (0.008)&	0.755&	0.189&	0.056\\
  & Task Can Be Done      & 5,676&	1,419&	0.669 (0.005)&	0.661 (0.021)&	0.683&	0.229&	0.088\\
  & Task Needs More Info  & 4,752&	1,188&	0.717 (0.006)&	0.701 (0.020)&	0.346&	0.420&	0.234 \\
  & Context Needed Known                    & 5,148&	1,287&	0.713 (0.006)&	0.695 (0.026)&	0.499&	0.286&	0.215
\\
  & Context Needed Unknown                & 5,781&	1,445&	0.656 (0.003)&	0.649 (0.011)&	0.754&	0.245&	0.000 \\
  & Context Not Needed           &	5,808&	1,452&	0.705 (0.007)&	0.695 (0.032)&	0.674&	0.174&	0.152\\
  & Context Sufficient          & 5,992&	1,498&	0.784 (0.009)&	0.773 (0.027)&	0.592&	0.345&	0.062\\
  & Context Missing       & 4,778&	1,194&	0.789 (0.011)&	0.782 (0.042)&	0.511&	0.422&	0.067 \\
  & MEAN &	5,700&	1,424&	0.720 (0.049)&	0.711 (0.050)& \textemdash & \textemdash & \textemdash\\		
  
\bottomrule
\end{tabular}}
\end{table}

\clearpage
\subsection{Optimization} \label{ec:optimization}

To provide insight into the versatility and case-fit of ML-Compass, we evaluate five deployment settings that reflect common organizational decision regimes:

\begin{enumerate}
    \item \textbf{Pure Capability}, where model selection relies solely on estimated utility, without any cost penalty. In this setting, our recommendation problem considers only the first term of the maximization objective shown in Equation~\ref{eq:empirical_problem}.
    
    \item \textbf{Cost-Aware (low)}, which introduces the second term of the objective function by applying a mild cost penalty that effectively constrains the normalized cost variable to remain above a lower threshold. The strength of this penalty is controlled by the parameter $\lambda$, set differently for each dataset: $\lambda=0.05$ for PRISM and $\lambda=0.50$ for HealthBench, in order to achieve comparable cost pressure.
    
    \item \textbf{Cost-Aware (high)}, which applies a stronger cost penalty, tightening the lower bound of the normalized cost variable according to the cost-sensitivity parameter $\lambda$. Here, we set $\lambda=0.50$ for PRISM and $\lambda=5.00$ for HealthBench.
    
    \item \textbf{Constrained (single)}, combining the mild cost penalty described above with a binding compliance screen on a single key internal measure ($C_2$), enforced by requiring its normalized value to exceed 0.5.
    
    \item \textbf{Constrained (all)}, which also uses the mild cost penalty, while imposing multi-dimensional compliance screens that require each normalized internal measure to exceed 0.33 of its observed range.
\end{enumerate}

Importantly, these compliance screens act as \emph{range constraints} on normalized variables, rather than eliminating a fixed fraction of models. By normalizing each internal measure to the $[0,1]$ interval, the penalties ensure that only models meeting the minimum threshold are considered, effectively restricting the feasible range in a controlled and interpretable way.

We will start by providing PRISM results for all these five scenarios, and then HealthBench results.

\begin{table*}[t]
\centering
\scriptsize
\setlength{\tabcolsep}{3pt}
\renewcommand{\arraystretch}{1.12}
\caption{PRISM — Pure Capability (no cost, no constraints). Oracle targets $(x^*,c^*)$ are shown when applicable; the model column is reported only when a single model is selected (otherwise \textemdash).}
\label{tab:prism_pure_capability_landscape_modelrule}
\resizebox{\linewidth}{!}{%
\begin{tabular}{llcclcccccccc}
\toprule
Method & $z$-type & $z$-type coefs & $x^*$ & $c^*$ & $u^*$ & obj$^*$ & Tier & Model & $x_{\text{sel}}$ & $c_{\text{sel}}$ & $u_{\text{sel}}$ & obj$_{\text{sel}}$ \\
\midrule
\texttt{mlc\_differentiated}&	Ethics-focused user&	(1.00, 0.00)&	(1.00, 0.00)&	10.10&	1.00&	1.00&	2&	\texttt{cohere-command}&	(1.00, 0.59)&	1.62&	1.00&	1\\
\texttt{actual}&	Ethics-focused user&	\textemdash&	\textemdash&	\textemdash&	\textemdash&	\textemdash&	2&	\texttt{cohere-command}&	(1.00, 0.59)&	1.62&	1.00&	1\\
\texttt{mlc\_differentiated}&	Safety-focused user&	(0.02, 0.98)&	(0.46, 1.00)&	10.10&	0.99&	0.99&	2&	\texttt{gpt-4-1106-preview}&	(0.66, 1.00)&	15.00&	0.99&	0.99\\
\texttt{actual}&	Safety-focused user&	\textemdash&	\textemdash&	\textemdash&	\textemdash&	\textemdash&	2&	\texttt{gpt-4}&	(0.62, 0.97)&	37.50&	0.97&	0.97\\
\texttt{mlc\_differentiated}&	General user&	(0.58, 0.42)&	(0.83, 0.74)&	10.10&	0.79&	0.79&	2&	\texttt{cohere-command}&	(1.00, 0.59)&	1.62&	0.83&	0.83\\
\texttt{actual}&	General user&	\textemdash&	\textemdash&	\textemdash&	\textemdash&	\textemdash&	2&	\texttt{cohere-command}&	(1.00, 0.59)&	1.62&	0.83&	0.83\\
\texttt{mlc\_differentiated\_avg}&	\textemdash&	\textemdash&	(0.76, 0.58)&	10.10&	0.93&	0.93&	2&	\textemdash &	(0.89, 0.73)&	6.08&	0.94&	0.94\\
\bottomrule
\end{tabular}%
}
\end{table*}

\begin{table*}[t] 
\centering 
\scriptsize 
\setlength{\tabcolsep}{3pt} 
\renewcommand{\arraystretch}{1.12} 
\caption{PRISM — Cost-Aware (Low $\lambda$) results. Oracle targets $(x^*,c^*)$ are shown when applicable; the model column is reported only when a single model is selected (otherwise \textemdash).} 
\label{tab:prism_cost_aware_lowlambda} 
\resizebox{\linewidth}{!}{%
\begin{tabular}{llcclcccccccc} 
\toprule 
Method & $z$-type & $z$-type coefs & $x^*$ & $c^*$ & $u^*$ & obj$^*$ & Tier & Model & $x_{\text{sel}}$ & $c_{\text{sel}}$ & $u_{\text{sel}}$ & obj$_{\text{sel}}$ \\ 
\midrule
\texttt{mlc\_differentiated} & Ethics-focused user & (1.00, 0.00) & (0.80, 0.00) & 3.30 & 0.80 & 0.63 & 0 & \texttt{oasst-pythia-12b} & (0.61, 0.00) & 0.30 & 0.61 & 0.59 \\
\texttt{actual} & Ethics-focused user & \textemdash & \textemdash & \textemdash & \textemdash & \textemdash & 2 & \texttt{cohere-command} & (1.00, 0.59) & 1.62 & 0.92 & 0.75 \\
\texttt{mlc\_differentiated} & Safety-focused user & (0.02, 0.98) & (0.07, 0.84) & 3.39 & 0.83 & 0.66 & 2 & \texttt{gpt-4-1106-preview} & (0.66, 1.00) & 15.00 & 0.99 & 0.24 \\
\texttt{actual} & Safety-focused user & \textemdash & \textemdash & \textemdash & \textemdash & \textemdash & 2 & \texttt{gpt-4} & (0.62, 0.97) & 37.50 & 0.97 & -0.91 \\
\texttt{mlc\_differentiated} & General user & (0.58, 0.42) & (0.62, 0.55) & 2.44 & 0.59 & 0.47 & 0 & \texttt{mistral-7b-instruct-v0.1} & (0.58, 0.50) & 0.16 & 0.55 & 0.54 \\
\texttt{actual} & General user & \textemdash & \textemdash & \textemdash & \textemdash & \textemdash & 2 & \texttt{cohere-command} & (1.00, 0.59) & 1.62 & 0.83 & 0.75 \\
\texttt{mlc\_differentiated\_avg} & \textemdash & \textemdash & (0.50, 0.46) & 3.04 & 0.74 & 0.59 & 0 & \textemdash & (0.62, 0.50) & 5.15 & 0.72 & 0.46 \\
\bottomrule 
\end{tabular}%
}
\end{table*}

\begin{table*}[t]
\centering
\scriptsize
\setlength{\tabcolsep}{3pt}
\renewcommand{\arraystretch}{1.12}
\caption{PRISM — Cost-Aware (High $\lambda$) results. Oracle targets $(x^*,c^*)$ are shown when applicable; the model column is reported only when a single model is selected (otherwise \textemdash).}
\label{tab:prism_cost_aware_high_lambda}
\resizebox{\linewidth}{!}{%
\begin{tabular}{llcclcccccccc}
\toprule
Method & $z$-type & $z$-type coefs & $x^*$ & $c^*$ & $u^*$ & obj$^*$ & Tier & Model & $x_{\text{sel}}$ & $c_{\text{sel}}$ & $u_{\text{sel}}$ & obj$_{\text{sel}}$ \\
\midrule
\texttt{mlc\_differentiated} & Ethics-focused user & (1.0, 0.0) & (0.69, 0.0) & 1.59 & 0.69 & -0.11 & 0 & \texttt{oasst-pythia-12b} & (0.61, 0.0) & 0.30 & 0.61 & 0.46 \\
\texttt{actual} & Ethics-focused user& \textemdash & \textemdash & \textemdash & \textemdash & \textemdash & 2 & \texttt{cohere-command} & (1.0, 0.59) & 1.63 & 1.0 & 0.19 \\
\texttt{mlc\_differentiated} & Safety-focused user & (0.02, 0.98) & (0.07, 0.72) & 1.59 & 0.71 & -0.09 & 0 & \texttt{mistral-7b-instruct-v0.1} & (0.58, 0.50) & 0.16 & 0.51 & 0.42 \\
\texttt{actual} & Safety-focused user & \textemdash & \textemdash & \textemdash & \textemdash & \textemdash & 2 & \texttt{gpt-4} & (0.62, 0.97) & 37.50 & 0.97 & -17.78 \\
\texttt{mlc\_differentiated} & General user & (0.58, 0.42) & (0.57, 0.51) & 1.59 & 0.54 & -0.25 & 0 & \texttt{mistral-7b-instruct-v0.1} & (0.58, 0.50) & 0.16 & 0.55 & 0.47 \\
\texttt{actual} & General user & \textemdash & \textemdash & \textemdash & \textemdash & \textemdash & 2 & \texttt{cohere-command} & (1.0, 0.59) & 1.63 & 0.83 & 0.02 \\
\texttt{mlc\_differentiated\_avg} & \textemdash & \textemdash & (0.44, 0.41) & 1.59 & 0.65 & -0.15 & 0 & \textemdash & (0.59, 0.34) & 0.21 & 0.55 & 0.45 \\
\bottomrule
\end{tabular}%
}
\end{table*}

\begin{table*}[t]
\centering
\scriptsize
\setlength{\tabcolsep}{3pt}
\renewcommand{\arraystretch}{1.12}
\caption{PRISM — Constrained (single measure) results. Oracle targets $(x^*,c^*)$ are shown when applicable; the model column is reported only when a single model is selected (otherwise \textemdash).}
\label{tab:prism_constrained_single_measure}
\resizebox{\linewidth}{!}{%
\begin{tabular}{llcclcccccccc}
\toprule
Method & $z$-type & $z$-type coefs & $x^*$ & $c^*$ & $u^*$ & obj$^*$ & Tier & Model & $x_{\text{sel}}$ & $c_{\text{sel}}$ & $u_{\text{sel}}$ & obj$_{\text{sel}}$ \\
\midrule
\texttt{mlc\_differentiated} & Ethics-focused user & (1.0, 0.0) & (0.75, 0.50) & 4.02 & 0.75 & 0.55 & 0 & \texttt{mistral-7b-instruct-v0.1} & (0.58, 0.50) & 0.16 & 0.58 & 0.57 \\
\texttt{actual} & Ethics-focused user & \textemdash & \textemdash & \textemdash & \textemdash & \textemdash & 2 & \texttt{cohere-command} & (1.0, 0.59) & 1.63 & 1.0 & 0.92 \\
\texttt{mlc\_differentiated} & Safety-focused user & (0.02, 0.98) & (0.07, 0.84) & 3.39 & 0.83 & 0.66 & 2 & \texttt{gpt-4-1106-preview} & (0.66, 1.0) & 15.0 & 0.99 & 0.24 \\
\texttt{actual} & Safety-focused user & \textemdash & \textemdash & \textemdash & \textemdash & \textemdash & 2 & \texttt{gpt-4} & (0.62, 0.97) & 37.5 & 0.97 & -0.91 \\
\texttt{mlc\_differentiated} & General user & (0.58, 0.42) & (0.62, 0.55) & 2.44 & 0.59 & 0.47 & 0 & \texttt{mistral-7b-instruct-v0.1} & (0.58, 0.50) & 0.16 & 0.55 & 0.54 \\
\texttt{actual} & General user & \textemdash & \textemdash & \textemdash & \textemdash & \textemdash & 2 & \texttt{cohere-command} & (1.0, 0.59) & 1.63 & 0.83 & 0.75 \\
\texttt{mlc\_differentiated\_avg} & \textemdash & \textemdash & (0.48, 0.63) & 3.29 & 0.72 & 0.56 & 0 & \textemdash & (0.61, 0.67) & 5.11 & 0.71 & 0.45 \\
\bottomrule
\end{tabular}%
}
\end{table*}

\begin{table*}[t] 
\centering 
\scriptsize 
\setlength{\tabcolsep}{3pt} 
\renewcommand{\arraystretch}{1.12} 
\caption{PRISM — Constrained (all) results. Oracle targets $(x^*,c^*)$ are shown when applicable; the model column is reported only when a single model is selected (otherwise \textemdash).} 
\label{tab:prism_constrained_all} 
\resizebox{\linewidth}{!}{%
\begin{tabular}{llcclcccccccc} 
\toprule 
Method & $z$-type & $z$-type coefs & $x^*$ & $c^*$ & $u^*$ & obj$^*$ & Tier & Model & $x_{\text{sel}}$ & $c_{\text{sel}}$ & $u_{\text{sel}}$ & obj$_{\text{sel}}$ \\ 
\midrule
\texttt{mlc\_differentiated} & Ethics-focused user & (1.00, 0.00) & (0.78, 0.33) & 3.50 & 0.78 & 0.61 & 0 & \texttt{llama-2-7b-chat} & (0.60, 0.39) & 0.10 & 0.60 & 0.60 \\
\texttt{actual} & Ethics-focused user & \textemdash & \textemdash & \textemdash & \textemdash & \textemdash & 2 & \texttt{cohere-command} & (1.00, 0.59) & 1.63 & 1.00 & 0.92 \\
\texttt{mlc\_differentiated} & Safety-focused user & (0.02, 0.98) & (0.33, 0.82) & 3.65 & 0.81 & 0.63 & 2 & \texttt{gpt-4-1106-preview} & (0.66, 1.00) & 15.00 & 0.99 & 0.24 \\
\texttt{actual} & Safety-focused user & \textemdash & \textemdash & \textemdash & \textemdash & \textemdash & 2 & \texttt{gpt-4} & (0.62, 0.97) & 37.50 & 0.97 & -0.91 \\
\texttt{mlc\_differentiated} & General user & (0.58, 0.42) & (0.62, 0.55) & 2.44 & 0.59 & 0.47 & 0 & \texttt{mistral-7b-instruct-v0.1} & (0.58, 0.50) & 0.16 & 0.55 & 0.54 \\
\texttt{actual} & General user & \textemdash & \textemdash & \textemdash & \textemdash & \textemdash & 2 & \texttt{cohere-command} & (1.00, 0.59) & 1.63 & 0.83 & 0.75 \\
\texttt{mlc\_differentiated\_avg} & \textemdash & \textemdash & (0.58, 0.57) & 3.20 & 0.73 & 0.57 & 0 & \textemdash & (0.62, 0.63) & 5.09 & 0.72 & 0.46 \\
\bottomrule 
\end{tabular}%
}
\end{table*}

\clearpage

\begin{table*}[t]
\centering
\scriptsize
\setlength{\tabcolsep}{3pt}
\renewcommand{\arraystretch}{1.12}
\caption{HealthBench — Pure Capability (no cost, no constraints). Oracle targets $(x^*,c^*)$ are shown when applicable; the model column is reported only when a single model is selected (otherwise \textemdash).}
\label{tab:hb_pure_capability_landscape_modelrule}
\resizebox{\linewidth}{!}{%
\begin{tabular}{llcclcccccccc}
\toprule
Method & $z$-type & $z$-type coefs & $x^*$ & $c^*$ & $u^*$ & obj$^*$ & Tier & Model & $x_{\text{sel}}$ & $c_{\text{sel}}$ & $u_{\text{sel}}$ & obj$_{\text{sel}}$ \\

\midrule
\texttt{mlc\_differentiated} &	Reduce Uncertainty&	(0.64, 0.11, 0.26)&	(0.74, 0.52, 0.69)&	0.30&	0.71&	0.71&	0 &	\texttt{gemma-3-4b}&	(0.11, 0.04, 0.11)&	0.00&	0.94&	0.94\\
\texttt{actual}&	Reduce Uncertainty&	\textemdash&	\textemdash&	\textemdash&	\textemdash&	\textemdash&	0 &	\texttt{gemma-3-4b}&	(0.11, 0.04, 0.11)&	0.00&	0.94&	0.94\\
\texttt{mlc\_differentiated}&	Conditionally emergent&	(0.60, 0.20, 0.20)&	(0.74, 0.52, 0.69)&	0.30&	0.98&	0.98&	2 &	\texttt{gpt-oss-120b}&	(0.95, 0.42, 1.00)&	1.00&	1.00&	1.00\\
\texttt{actual}&	Conditionally emergent&	\textemdash&	\textemdash&	\textemdash&	\textemdash&	\textemdash&	2 &	\texttt{gpt-oss-120b}&	(0.95, 0.42, 1.00)&	1.00&	1.00&	1.00\\
\texttt{mlc\_differentiated}&	Context Not Needed&	(0.67, 0.17, 0.15)&	(0.74, 0.52, 0.69)&	0.30&	0.90&	0.90&	1 &	\texttt{qwen3-next-80b-a3b-nr}&	(0.83, 0.09, 0.25)&	0.67&	0.96&	0.96\\
\texttt{actual}&	Context Not Needed&	\textemdash&	\textemdash&	\textemdash&	\textemdash&	\textemdash&	1 &	\texttt{qwen3-next-80b-a3b-nr}&	(0.83, 0.09, 0.25)&	0.67&	0.96&	0.96\\
\texttt{mlc\_differentiated}&	Context Needed Unknown&	(0.75, 0.25, 0.00)&	(0.74, 0.52, 0.69)&	0.30&	0.80&	0.80&	2 &	\texttt{gpt-oss-120b}&	(0.95, 0.42, 1.00)&	1.00&	0.80&	0.80\\
\bottomrule
\end{tabular}%
}
\end{table*}

\begin{table*}[t] 
\centering 
\scriptsize 
\setlength{\tabcolsep}{3pt} 
\renewcommand{\arraystretch}{1.12} 
\caption{HealthBench — Cost-Aware (Low $\lambda$) results. Oracle targets $(x^*,c^*)$ are shown when applicable; the model column is reported only when a single model is selected (otherwise \textemdash).} 
\label{tab:hb_cost_aware_lowlambda} 
\resizebox{\linewidth}{!}{%
\begin{tabular}{llcclcccccccc} 
\toprule 
Method & $z$-type & $z$-type coefs & $x^*$ & $c^*$ & $u^*$ & obj$^*$ & Tier & Model & $x_{\text{sel}}$ & $c_{\text{sel}}$ & $u_{\text{sel}}$ & obj$_{\text{sel}}$ \\ 
\midrule
\texttt{mlc\_differentiated} & Reduce Uncertainty & (0.64, 0.11, 0.26) & (0.13, 0.00, 0.00) & 0.02 & 0.92 & 0.91 & 0 & \texttt{gemma-3-4b} & (0.11, 0.04, 0.11) & 0.00 & 0.94 & 0.94 \\
\texttt{actual} & Reduce Uncertainty & \textemdash & \textemdash & \textemdash & \textemdash & \textemdash & 0 & \texttt{gemma-3-4b} & (0.11, 0.04, 0.11) & 0.00 & 0.94 & 0.94 \\
\texttt{mlc\_differentiated} & Conditional Emergency & (0.60, 0.20, 0.20) & (0.39, 0.07, 0.30) & 0.02 & 0.87 & 0.86 & 2 & \texttt{gpt-oss-120b} & (0.95, 0.42, 1.00) & 1.00 & 1.00 & 0.50 \\
\texttt{actual} & Conditional Emergency & \textemdash & \textemdash & \textemdash & \textemdash & \textemdash & 2 & \texttt{gpt-oss-120b} & (0.95, 0.42, 1.00) & 1.00 & 1.00 & 0.50 \\
\texttt{mlc\_differentiated} & Context Not Needed & (0.67, 0.17, 0.15) & (0.31, 0.00, 0.20) & 0.02 & 0.94 & 0.93 & 1 & \texttt{qwen3-next-80b-a3b-nr} & (0.83, 0.09, 0.25) & 0.67 & 0.96 & 0.62 \\
\texttt{actual} & Context Not Needed & \textemdash & \textemdash & \textemdash & \textemdash & \textemdash & 1 & \texttt{qwen3-next-80b-a3b-nr} & (0.83, 0.09, 0.25) & 0.67 & 0.96 & 0.62 \\
\texttt{mlc\_differentiated} & Context Needed Unknown & (0.75, 0.25, 0.00) & (0.20, 0.40, 0.00) & 0.04 & 0.68 & 0.66 & 2 & \texttt{gpt-oss-120b} & (0.95, 0.42, 1.00) & 1.00 & 0.80 & 0.30 \\
\bottomrule
\end{tabular}%
}
\end{table*}

\begin{table*}[t] 
\centering 
\scriptsize 
\setlength{\tabcolsep}{3pt} 
\renewcommand{\arraystretch}{1.12} 
\caption{HealthBench — Cost-Aware (High $\lambda$) results. Oracle targets $(x^*,c^*)$ are shown when applicable; the model column is reported only when a single model is selected (otherwise \textemdash).} 
\label{tab:hb_cost_aware_highlambda} 
\resizebox{\linewidth}{!}{%
\begin{tabular}{llcclcccccccc} 
\toprule 
Method & $z$-type & $z$-type coefs & $x^*$ & $c^*$ & $u^*$ & obj$^*$ & Tier & Model & $x_{\text{sel}}$ & $c_{\text{sel}}$ & $u_{\text{sel}}$ & obj$_{\text{sel}}$ \\ 
\midrule
\texttt{mlc\_differentiated} & Reduce Uncertainty & (0.64, 0.11, 0.26) & (0.54, 0.00, 0.37) & 0.02 & 0.69 & 0.59 & 0 & \texttt{gemma-3-4b} & (0.11, 0.04, 0.11) & 0.00 & 0.94 & 0.94 \\
\texttt{actual} & Reduce Uncertainty & \textemdash & \textemdash & \textemdash & \textemdash & \textemdash & 0 & \texttt{gemma-3-4b} & (0.11, 0.04, 0.11) & 0.00 & 0.94 & 0.94 \\
\texttt{mlc\_differentiated} & Conditional Emergency & (0.60, 0.20, 0.20) & (0.39, 0.07, 0.30) & 0.02 & 0.87 & 0.77 & 0 & \texttt{reka-flash-3} & (0.47, 0.05, 0.07) & 0.15 & 0.74 & 0.00 \\
\texttt{actual} & Conditional Emergency & \textemdash & \textemdash & \textemdash & \textemdash & \textemdash & 2 & \texttt{gpt-oss-120b} & (0.95, 0.42, 1.00) & 1.00 & 1.00 & -4.00 \\
\texttt{mlc\_differentiated} & Context Not Needed & (0.67, 0.17, 0.15) & (0.48, 0.02, 0.37) & 0.02 & 0.94 & 0.85 & 0 & \texttt{aya-expanse-32b} & (0.07, 0.04, 0.12) & 0.25 & 0.67 & -0.56 \\
\texttt{actual} & Context Not Needed & \textemdash & \textemdash & \textemdash & \textemdash & \textemdash & 1 & \texttt{qwen3-next-80b-a3b-nr} & (0.83, 0.09, 0.25) & 0.67 & 0.96 & -2.40 \\
\texttt{mlc\_differentiated} & Context Needed Unknown & (0.75, 0.25, 0.00) & (0.21, 0.20, 0.14) & 0.02 & 0.68 & 0.58 & 0 & \texttt{gemma-3-27b} & (0.17, 0.05, 0.15) & 0.20 & 0.64 & -0.38 \\
\bottomrule
\end{tabular}%
}
\end{table*}

\begin{table*}[t] 
\centering 
\scriptsize 
\setlength{\tabcolsep}{3pt} 
\renewcommand{\arraystretch}{1.12} 
\caption{HealthBench — Constrained (single measure) results. Oracle targets $(x^*,c^*)$ are shown when applicable; the model column is reported only when a single model is selected (otherwise \textemdash).} 
\label{tab:prism_constrained_single} 
\resizebox{\linewidth}{!}{%
\begin{tabular}{llcclcccccccc} 
\toprule 
Method & $z$-type & $z$-type coefs & $x^*$ & $c^*$ & $u^*$ & obj$^*$ & Tier & Model & $x_{\text{sel}}$ & $c_{\text{sel}}$ & $u_{\text{sel}}$ & obj$_{\text{sel}}$ \\ 
\midrule
\texttt{mlc\_differentiated} & Reduce Uncertainty & (0.64, 0.11, 0.26) & (0.50, 0.12, 0.13) & 0.02 & 0.70 & 0.69 & 0 & \texttt{exaone-4.0-32b-nr} & (0.52, 0.13, 0.13) & 0.25 & 0.70 & 0.57 \\
\texttt{actual} & Reduce Uncertainty & \textemdash & \textemdash & \textemdash & \textemdash & \textemdash & 2 & \texttt{gpt-oss-120b} & (0.95, 0.42, 1.00) & 1.00 & 0.80 & 0.30 \\
\texttt{mlc\_differentiated} & Conditional Emergency & (0.60, 0.20, 0.20) & (0.50, 0.05, 0.29) & 0.02 & 0.83 & 0.82 & 2 & \texttt{gpt-oss-120b} & (0.95, 0.42, 1.00) & 1.00 & 1.00 & 0.50 \\
\texttt{actual} & Conditional Emergency & \textemdash & \textemdash & \textemdash & \textemdash & \textemdash & 2 & \texttt{gpt-oss-120b} & (0.95, 0.42, 1.00) & 1.00 & 1.00 & 0.50 \\
\texttt{mlc\_differentiated} & Context Not Needed & (0.67, 0.17, 0.15) & (0.50, 0.00, 0.27) & 0.02 & 0.94 & 0.93 & 1 & \texttt{qwen3-next-80b-a3b-nr} & (0.83, 0.09, 0.25) & 0.67 & 0.96 & 0.62 \\
\texttt{actual} & Context Not Needed & \textemdash & \textemdash & \textemdash & \textemdash & \textemdash & 1 & \texttt{qwen3-next-80b-a3b-nr} & (0.83, 0.09, 0.25) & 0.67 & 0.96 & 0.62 \\
\texttt{mlc\_differentiated} & Context Needed Unknown & (0.75, 0.25, 0.00) & (0.70, 0.40, 0.00) & 0.08 & 0.80 & 0.76 & 2 & \texttt{gpt-oss-120b} & (0.95, 0.42, 1.00) & 1.00 & 0.80 & 0.30 \\
\bottomrule
\end{tabular}%
}
\end{table*}

\begin{table*}[t] 
\centering 
\scriptsize 
\setlength{\tabcolsep}{3pt} 
\renewcommand{\arraystretch}{1.12} 
\caption{HealthBench — Constrained (all) results. Oracle targets $(x^*,c^*)$ are shown when applicable; the model column is reported only when a single model is selected (otherwise \textemdash).} 
\label{tab:HealthBench_constrained_all} 
\resizebox{\linewidth}{!}{%
\begin{tabular}{llcclcccccccc} 
\toprule 
Method & $z$-type & $z$-type coefs & $x^*$ & $c^*$ & $u^*$ & obj$^*$ & Tier & Model & $x_{\text{sel}}$ & $c_{\text{sel}}$ & $u_{\text{sel}}$ & obj$_{\text{sel}}$ \\ 
\midrule
\texttt{mlc\_differentiated} & Reduce Uncertainty & (0.64, 0.11, 0.26) & (0.97, 0.33, 0.48) & 0.18 & 0.80 & 0.72 & 2 & \texttt{gpt-oss-120b} & (0.95, 0.42, 1.00) & 1.00 & 0.80 & 0.30 \\
\texttt{actual} & Reduce Uncertainty & \textemdash & \textemdash & \textemdash & \textemdash & \textemdash & 2 & \texttt{gpt-oss-120b} & (0.95, 0.42, 1.00) & 1.00 & 0.80 & 0.30 \\
\texttt{mlc\_differentiated} & Conditional Emergency & (0.60, 0.20, 0.20) & (0.40, 0.40, 0.33) & 0.10 & 0.88 & 0.82 & 2 & \texttt{gpt-oss-120b} & (0.95, 0.42, 1.00) & 1.00 & 1.00 & 0.50 \\
\texttt{actual} & Conditional Emergency & \textemdash & \textemdash & \textemdash & \textemdash & \textemdash & 2 & \texttt{gpt-oss-120b} & (0.95, 0.42, 1.00) & 1.00 & 1.00 & 0.50 \\
\texttt{mlc\_differentiated} & Context Not Needed & (0.67, 0.17, 0.15) & (0.33, 0.40, 0.33) & 0.10 & 0.90 & 0.85 & 2 & \texttt{gpt-oss-120b} & (0.95, 0.42, 1.00) & 1.00 & 0.91 & 0.41 \\
\texttt{actual} & Context Not Needed & \textemdash & \textemdash & \textemdash & \textemdash & \textemdash & 2 & \texttt{gpt-oss-120b} & (0.95, 0.42, 1.00) & 1.00 & 0.91 & 0.41 \\
\texttt{mlc\_differentiated} & Context Needed Unknown & (0.75, 0.25, 0.00) & (0.70, 0.33, 0.33) & 0.11 & 0.57 & 0.51 & 2 & \texttt{gpt-oss-120b} & (0.95, 0.42, 1.00) & 1.00 & 0.80 & 0.30 \\
\bottomrule
\end{tabular}%
}
\end{table*}

\end{document}